\documentclass[twoside]{article}
\usepackage[accepted]{aistats2025}
\usepackage[round]{natbib}

\usepackage{hyperref}
\usepackage{url}

\usepackage{graphicx, wrapfig} 
\usepackage{amsmath}
\usepackage{amsthm}
\usepackage{amssymb}
\usepackage{caption}
\usepackage{subcaption}
\usepackage{rotating}

\usepackage{color}

\usepackage{algorithm}
\usepackage{algorithmic}

\newtheorem{theorem}{Theorem}
\newtheorem{lemma}[theorem]{Lemma}
\newtheorem{corollary}[theorem]{Corollary}

\newtheorem{definition}{Definition}
\newtheorem{example}{Example}

\usepackage{apptools} 
\AtAppendix{\counterwithin{theorem}{section}}

\newtheorem*{example_for_appendix}{Example}

\def\E{\mathbb{E}}
\def\Var{\mathrm{Var}}
\def\Cov{\mathrm{Cov}}
\def\Corr{\mathrm{Corr}}

\DeclareMathOperator*{\argmin}{arg\,min}

\newcommand{\ie}{\textit{i.e., }}

\newcommand{\categZorC}{L}
\newcommand{\categZorCsmall}{l}
\newcommand{\cutpoint}{\tau}

\definecolor{mydarkblue}{rgb}{0,0.08,0.45}
\hypersetup{ %
pdftitle={},
pdfkeywords={},
pdfborder=0 0 0,
pdfpagemode=UseNone,
colorlinks=true,
linkcolor=mydarkblue,
citecolor=mydarkblue,
filecolor=mydarkblue,
urlcolor=mydarkblue,
}

\begin{document}

\runningtitle{A Generalized Theory of Mixup for Structure-Preserving Synthetic Data}

\twocolumn[

\aistatstitle{
A Generalized Theory of Mixup \\
for Structure-Preserving Synthetic Data
}

\aistatsauthor{ Chungpa Lee \And Jongho Im \And Joseph H.T. Kim }

\aistatsaddress{ Yonsei University \And  Yonsei University \And Yonsei University } ]

\begin{abstract}
Mixup is a widely adopted data augmentation technique known for enhancing the generalization of machine learning models by interpolating between data points. Despite its success and popularity, limited attention has been given to understanding the statistical properties of the synthetic data it generates. In this paper, we delve into the theoretical underpinnings of mixup, specifically its effects on the statistical structure of synthesized data. We demonstrate that while mixup improves model performance, it can distort key statistical properties such as variance, potentially leading to unintended consequences in data synthesis. To address this, we propose a novel mixup method that incorporates a generalized and flexible weighting scheme, better preserving the original data’s structure. Through theoretical developments, we provide conditions under which our proposed method maintains the (co)variance and distributional properties of the original dataset. Numerical experiments confirm that the new approach not only preserves the statistical characteristics of the original data but also sustains model performance across repeated synthesis, alleviating concerns of model collapse identified in previous research.
\end{abstract}

\section{INTRODUCTION}\label{sec:intro}

Mixup is a prominent data augmentation method \citep{zhang2017mixup} that generates new instances by linearly combining observed instances, applicable across both structured and unstructured datasets. By training models on these interpolated samples, mixup enhances the generalization performance of state-of-the-art neural network architectures \citep{verma2019manifold, yun2019cutmix, guo2020nonlinear, sohn2022genlabel, kim2021lada, baena2022preventing, chen2020local, zhang2022m, sun2024patch}. A similar approach, SMOTE (Synthetic Minority Over-sampling Technique) \citep{chawla2002smote, he2008adasyn, bunkhumpornpat2012dbsmote, douzas2018improving}, also leverages interpolated synthetic instances to enhance model performance particularly for imbalanced or long-tail distributions, showcasing the effectiveness of mixup methods.

In this paper we place special focus on data synthesis, an important constituent of data augmentation. While there is extensive research on how synthetic data generated by mixup can enhance model performance \citep{carratino2020mixup, zhang2020does}, less attention has been given to understanding the fundamental properties of the synthesized data itself; see Sec.~\ref{sec:related:works:mixup}. In fact most mixup methods generate linearly interpolated instances by taking a weighted average where the weights are randomly drawn from distributions within the range of $[0,1]$, such as the beta or the uniform distribution. However, this interpolation process reduces the variance, which inevitably distorts the statistical structure of the original dataset both marginally and jointly. The net effect is a less dispersed dataset with more emphasis on representative instances and suppressing the others. In this regard, mixup-based synthetic datasets achieve better performance in training machine learning models from sacrificing non-representative instances, such as the tail instances, in the dataset. Naturally, understanding the impact of mixup warrants further research.

In a similar line of thought, a recent work in Nature \citep{shumailov2024ai} raises concerns regarding the risks associated with training models using data that has been repeatedly synthesized. This phenomenon, known as \textit{model collapse}, describes a situation where the tails of the original distribution are lost after repeatedly synthesizing the original dataset, demonstrating that over-reliance on synthetic data can lead to catastrophic defects in model training. As the prevalence of synthetic data from generative models increases, therefore, it is essential to carefully consider the quality and structure of this synthetic data to maintain the benefits of training on such datasets.

To this end, in this paper we formally investigate the theoretical properties of mixup and its impact on the resulting synthetic data. Our findings provide insight into how mixup alters the statistical structure of synthetic data in comparison to the original, thereby explaining why standard mixup, while beneficial for improving machine learning model performance, can lead to unintended effects in data synthesis. To address this, we also propose a new mixup method featuring a more generalized and flexible weighting scheme, allowing the synthetic dataset to better preserve the underlying structure of the original data. Our key contributions are as follows:

\begin{itemize}
    \item We theoretically derive a set of conditions for a mixup weight distribution that preserves the (co)variance for any pair of continuous variables in a general setting.
    \item We prove that the mean and variance of any numerical variable in a dataset, conditioned on a categorical variable in the same dataset, can be maintained within a specified error bound, which can be controlled by a function of mixup weights.
    \item As an additional contribution, we propose a new class of mixup weight distributions that   satisfy these theoretical conditions, thereby preserving the original data structure with respect to  both mean and variance. 
    \item
    Our numerical experiments show that the proposed mixup method generates synthetic data that preserves fundamental distributional properties, leading to more accurate statistical inferences. Also, regarding model performance with synthetic data, the proposed mixup method yields results comparable to existing synthesis techniques. Notably, it significantly maintains performance under repeated synthesis, addressing concerns raised by \citet{shumailov2024ai}.
\end{itemize}

\begin{figure*}[ht]
     \centering
     \includegraphics[width=\textwidth]{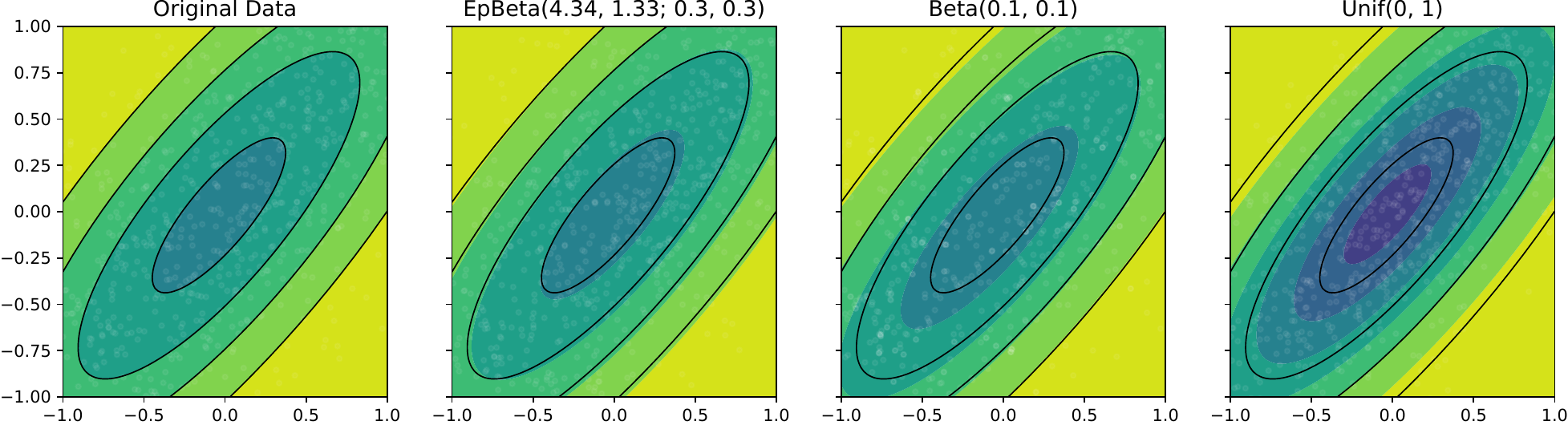}
     \caption{
     A contour plot of a sample comprising $1000$ data points from  $N\big(\left(\begin{smallmatrix}0\\0\end{smallmatrix}\right), \left(\begin{smallmatrix}1.1 &0.9\\0.9&1.1\end{smallmatrix}\right)\big)$ is shown in the first plot. This original data is then synthesized with three different mixup weight distributions: the proposed $\operatorname{EpBeta}(4.34,1.33;0.3,0.3)$, as well as two standard distributions, $\operatorname{Beta}(0.1,0.1)$ and $\operatorname{Unif}(0, 1)$.  While the synthetic data generated by $\operatorname{Beta}$ or $\operatorname{Unif}$ is shrunken, the proposed $\operatorname{EpBeta}$ preserves the original data structure.}
     \label{fig:one}
\end{figure*}

\section{RELATED WORK}
\label{sec:related:works}

\subsection{Analysis on Mixup}
\label{sec:related:works:mixup}

Several theoretical studies have explored the effects of mixup \citep{carratino2020mixup, zhang2020does, zhang2022and, park2022unified}. In particular, \cite{zhang2020does} and \cite{carratino2020mixup} demonstrate how training with mixup-generated data enhances model regularization and generalization from the perspective of empirical risk minimization. Furthermore, \cite{zhang2020does} showed that the coefficients of linear least-squares regression are preserved for any synthetic data generated by mixup methods. This preservation property follows directly from the fact that the key statistic in linear regression is the correlation, as shown in Appendix~\ref{appendix:proofs:regcoef}.

A distinctive aspect of our work is the explicit formulation of conditions that preserve the structure of synthetic data. Unlike prior approaches that require post-generation transformations to maintain statistical properties, our method ensures structure preservation during the data generation process, eliminating the need for additional computational steps.

\subsection{Synthetic Tabular Data}

In the statistical community, various methods for synthesizing tabular data have been extensively studied \citep{raghunathan2003multiple, nowok2016synthpop, kim2014multiple, si2013nonparametric, murray2016multiple}.
Advances in deep neural network-based generative models have further led to the development of techniques such as variational autoencoders \citep{xu2019modeling, ma2020vaem}, generative adversarial networks \citep{park2018data, choi2017generating, xu2019modeling, zhao2021ctab, zhao2024ctab, baowaly2019synthesizing}, diffusion models \citep{kotelnikov2023tabddpm, kim2022sos, lee2023codi, kim2022stasy, zhang2023mixed}, and large language models \citep{borisov2022language, solatorio2023realtabformer, zhang2023generative, gulati2024tabmt} for synthesizing tabular data. 

Although some of these generative models demonstrate performance comparable to traditional statistical methods, they require significant time and resources for model training. In contrast, our structure-preserving mixup method can generate high-utility synthetic data without the need for training a model. Additionally, it offers the advantage of allowing explicit control over the degree of preservation of the original data structure.

\section{PROPERTIES OF SYNTHETIC DATA GENERATED BY MIXUP \label{sec:properties}}

In this section, we analysis the statistical properties of synthetic data generated by the mixup method. Especially, we compare the mean and variance of the synthetic data $\tilde{D}$ against those of the original $D$. Matching these key statistics is crucial as many models rely on assumptions and analyses related to the mean and variance structure; see Sec.~\ref{sec:experiment}.

Data typically consists of two types of variables; continuous and categorical. To fix the notation, let us define the original instances as $D_i = (X_i,Y_i,\categZorC_i)$ for $i\in [n]$ where $X,Y\in \mathbb{R}$ are continuous variables, and $\categZorC \in[c]$ is a categorical variable with $c$ categories. We assume that $D_i$'s are independent and identically distributed according to the population distribution $\mathfrak{D}$. While $D_i$ contains only three variables for our theoretical developments, the results are general and can be easily extended to the cases of multiple continuous and categorical variables.

We denote the synthetic instances generated using the mixup by $\tilde{D}_k = (\tilde{X}_k,\tilde{Y}_k,\tilde{\categZorC}_k)$ for $k\in [m]$. 
To make the source clear, we use indices `$i,j\in[n]$' for each original instance and `$k\in[m]$' for each synthetic instance.

\subsection{Synthetic Data Generated by Mixup}
\label{sec:problemsetup}

Synthetic data using the mixup can be created as follows. First, we write the mixup weights as $W^X_k, W^Y_k, W^\categZorC_k \in \mathbb{R}$, each of which is associated to variables $X$, $Y$, and $\categZorC$, respectively. These weights are random variables, independent and identically generated from some given distributions $\mathfrak{W}^X, \mathfrak{W}^Y$, and $\mathfrak{W}^\categZorC$, respectively.  Taking values in $[0,1]$ is typical for the weights but not required. By construction, these mixup weights are independent of the original instances $D_i$ for all $i\in[n]$. A synthetic instance  $\tilde{D}_k = (\tilde{X}_k,\tilde{Y}_k,\tilde{\categZorC}_k)$ is obtained by randomly selecting two original instances $D_{i_k}$ and $D_{j_k}$ and applying the following transform:
\begin{align}
    \label{eq:mixup:x}
    \tilde{X}_k &= W^X_k X_{i_{k}} + (1-W^X_k) X_{j_{k}},
    \\
    \label{eq:mixup:y}
    \tilde{Y}_k &= W^Y_k Y_{i_{k}} + (1-W^Y_k) Y_{j_{k}},
\end{align}
\begin{align}
    \label{eq:mixup:z}
    \tilde{\categZorC}_k &= 
    \begin{cases}
    \categZorC_{i_{k}} &\text{ if } W^\categZorC_{k} \geq \cutpoint\\
    \categZorC_{j_{k}} &\text{ if } W^\categZorC_{k} < \cutpoint,
    \end{cases}
\end{align}
where $\cutpoint\in\mathbb{R}$ 
is the pre-defined cut point for the categorical variable $\categZorC$, with $\cutpoint=0.5$ being a default choice. Unlike the continuous variables, the mixup for a categorical variable requires a random selection due to its nature. 
It is noted that common mixup methods impose the same weight, following the original proposal by \cite{zhang2017mixup}. That is, all weights in \eqref{eq:mixup:x}--\eqref{eq:mixup:z} are set equal so that
\begin{equation}
    \label{eq:same:instance:weight}
    W=W^X_k=W^Y_k=W^\categZorC_k \text{ for all }k\in[m].
\end{equation}
We call \eqref{eq:mixup:x}--\eqref{eq:same:instance:weight} the equal-weight or standard mixup scheme whereas \eqref{eq:mixup:x}--\eqref{eq:mixup:z} the general-weight mixup scheme. The standard scheme is a special case of the general-weight scheme, and easier to work with; in fact most theoretical developments  in the literature have been based on the standard scheme. We differentiate these two schemes because several theoretical findings in this paper hold under the general-weight scheme, and thus applicable to a more general setting.

In what follows the indices $i, j$ and $k$ are omitted for simple notation provided that there is no confusion.

\subsection{Continuous Variable \label{sec:properties:conti}}

Under the common mixup technique explained above, it is trivial to prove that the mean of a continuous variable $X$ is always preserved, \ie $\E\big[\tilde{X}\big] = \E\big[X\big]$, regardless of the choice of mixup weight distribution: 
\begin{equation}
    \nonumber
    \E\big[\tilde{X}\big] = \E\big[W^X\big]\E\big[X\big] + (1-\E\big[W^X\big])\E\big[X\big] = \E\big[X\big].
\end{equation}
For the variance however things are more complicated. In fact we can show that (see Appendix~\ref{appendix:proof:sec2})
\begin{align}
    \Var\big[\tilde{X}\big]
     &= 
    \nonumber
    \Var\big[X\big]+
    2 \E\big[\big(W^X\big)^2 - W^X \big] \Var\big[X\big],
\end{align}
which is generally different from $\Var\big[X\big]$. The exact relationship between $\Var\big[\tilde{X}\big]$ and  $\Var\big[X\big]$ is presented below.

\begin{lemma}[Variance]
\label{thm:var}
\leavevmode For any synthetic $\tilde X$ generated 
from $X$ in (\ref{eq:mixup:x}):
\begin{enumerate}
    \item  $\Var\big[\tilde{X}\big] = \Var\big[X\big]$, if and only if (iff) the first and second moments of mixup weight are equal. That is
\begin{equation}
    \label{eq:necessary:var}
    \E\big[\big(W^X\big)^2\big] = \E\big[W^X \big].
\end{equation}

    \item $\Var\big[\tilde{X}\big] <  \Var\big[X\big]$, 
     iff $\E\big[\big(W^X\big)^2\big] < \E\big[W^X \big]$.

    \item $\Var\big[\tilde{X}\big] > \Var\big[X\big]$.
    iff $\E\big[\big(W^X\big)^2\big] > \E\big[W^X \big]$.
\end{enumerate}
\end{lemma}
 All proofs of this paper are provided in Appendix~\ref{appendix:proofs}. Another quantity of interest is the covariance because synthetic data is often required to preserve the correlation of the original data. 

\begin{theorem}[Covariance]
\label{thm:cov}
\leavevmode For any synthetic pair $(\tilde{X}, \tilde{Y})$ generated from $({X}, {Y})$ using the general-weight mixup in \eqref{eq:mixup:x} and \eqref{eq:mixup:y}:
\begin{enumerate}
    \item  $\Cov\big[\tilde{X}, \tilde{Y}\big] = \Cov\big[X, Y\big]$, iff
    \begin{flalign}
        \label{eq:necessary:cov}
        &\E\big[ W^X W^Y \big] = \frac{1}{2} \big(\E\big[ W^X  \big]+ \E\big[W^Y \big]\big). &&
    \end{flalign}

    \item $\Cov\big[\tilde{X}, \tilde{Y}\big] < \Cov\big[X, Y\big]$,
     iff 
     
     $\E\big[ W^X W^Y \big] < \frac{1}{2} \big(\E\big[ W^X  \big]+ \E\big[W^Y \big]\big)$.
    \item $\Cov\big[\tilde{X}, \tilde{Y}\big] > \Cov\big[X, Y\big]$, iff
 
    $\E\big[ W^X W^Y \big] > \frac{1}{2} \big(\E\big[ W^X  \big]+ \E\big[W^Y \big]\big)$.
\end{enumerate}
\end{theorem}

We comment that \cite{zhang2020does} showed that the coefficients of the linear regression model are preserved under the standard mixup scheme. Though they did not use in their proof, this preservation is essentially a consequence of the correlation-preserving property of the synthetic data when the mixup is conducted with equal weights as in \eqref{eq:same:instance:weight}; the following result can shorten the proof of \cite{zhang2020does}, see Appendix~\ref{appendix:proofs:regcoef}.

\begin{corollary}[Correlation]
\label{thm:corr}
     For any synthetic pair $(\tilde{X}, \tilde{Y})$ generated from $(X, Y)$ using the standard mixup scheme,
     we have
    \begin{equation}
    \Corr\big[\tilde{X}, \tilde{Y}\big] = \Corr\big[X, Y\big].
    \end{equation}
\end{corollary}

Before closing this section we emphasize that the results in this section hold generally and equally applicable for data with multiple variables with no further modifications.

\subsection{Continuous Variable Conditioned by Categorical Variable}

We now bring in the categorical variable $\categZorC$ so that we can investigate the synthetic distribution of both continuous and categorical variables jointly. This is motivated by the fact that preserving the mean and variance of the continuous variables conditional on the categorical variable is often necessary in data analyses; \ie when the height of students in a school  exhibit different distributions depending on gender, synthetic height datasets need to preserve their mean and variance for each gender.

Without loss of generality, let us consider a synthetic pair  $(\tilde{X}, \tilde{\categZorC})$ generated from $({X}, {\categZorC})$ using the general-weight mixup scheme, where $X$ is continuous and $\categZorC$  categorical. Our goal is to study the conditional mean  $\E\big[\tilde{X}|\tilde{\categZorC}\big]$. 
For this, we start with defining a special function of the mixup weights.  

\begin{definition}\label{def:u:function}
   Define the function of general mixup weights $W^X$ and $W^\categZorC$ with a cut point $\cutpoint$ as
    \begin{align}
        u\big(W^X, W^\categZorC,\cutpoint\big)
        &= \nonumber
        \E\big[ \big(1-W^X\big) \mathbf{I} \{ W^\categZorC\geq \cutpoint\} 
        \\ &\qquad\quad
        \label{eq:def:ufunction}
        + W^X  \mathbf{I} \{ W^\categZorC < \cutpoint\} \big],
    \end{align}
   where  $\mathbf{I}$ is an indicator function. Under the standard mixup scheme this  reduces to
   \begin{equation}
        \label{eq:def:ufunction:abb}
       u\big(W,\cutpoint\big)
        =
        \E\big[ \big(1-W\big) \mathbf{I} \{ W\geq \cutpoint\} + W  \mathbf{I} \{ W < \cutpoint\} \big].
   \end{equation}
\end{definition}

With this function we can show that the synthetic conditional mean $\E\big[\tilde{X} \big| \tilde{\categZorC}=\categZorCsmall \big]$ is a convex combination of the original conditional mean $\E\big[X \big| \categZorC=\categZorCsmall \big]$ and the marginal mean $\E\big[X\big]$.

\begin{theorem}[Conditional Mean]
\label{thm:con:mean} 
    For any synthetic pair  $(\tilde{X}, \tilde{\categZorC})$ generated from $({X}, {\categZorC})$ using the general-weight mixup, where $X$ is continuous and $\categZorC$ is categorical, the synthetic conditional mean $\E\big[\tilde{X} \big| \tilde{\categZorC}=\categZorCsmall \big]$ can be expressed as
    \begin{align}
        \nonumber
        \E\big[\tilde{X} \big| \tilde{\categZorC}=\categZorCsmall \big] 
        &= (1-u\big(W^X, W^\categZorC,\cutpoint\big))\cdot \E\big[X \big| \categZorC=\categZorCsmall \big]
        \\ &\qquad\; \label{eq:thm:con:mean}
        + u\big(W^X, W^\categZorC,\cutpoint\big) \cdot \E\big[X \big],
    \end{align}
or, alternatively, 
       \begin{align}
        \E\big[\tilde{X} \big| \tilde{\categZorC}=\categZorCsmall \big] 
        &= \nonumber
        \big(1-u\big(W^X, W^\categZorC,\cutpoint\big) \Pr\{ \categZorC \neq \categZorCsmall\} \big)
        \\ &\qquad \quad \nonumber
        \cdot \E\big[X \big| \categZorC=\categZorCsmall \big] 
        \\ &\qquad \nonumber
        + u\big(W^X, W^\categZorC,\cutpoint\big) \Pr\{ \categZorC \neq \categZorCsmall\} 
        \\ &\qquad \quad \label{eq:thm:con:mean:theother}
        \cdot E[X|\categZorC \neq \categZorCsmall].    
    \end{align}
\end{theorem}

Both expressions in Theorem~\ref{thm:con:mean} are weighted sums of two terms with the same first term $\E\big[X \big| {\categZorC} = \categZorCsmall \big]$, indicating that $\E\big[\tilde{X} \big| \tilde{\categZorC} = \categZorCsmall \big]$ partly uses the same information of the original data. The second terms however are different. In (\ref{eq:thm:con:mean})  it is given as $\E\big[X\big]$ which can be understood as the overall information of $X$, whereas  $\E\big[X \big| {\categZorC} \neq \categZorCsmall \big]$ in  (\ref{eq:thm:con:mean:theother}) is seen as complimentary information of the original data.  
This aligns with a general principle found in the data augmentation literature, which says that generating data leverages additional or overall information of the given data; see \cite{bowles2018gan} and \cite{mumuni2022data}. Also, focusing on   (\ref{eq:thm:con:mean:theother}), large $\Pr\{ \categZorC \neq \categZorCsmall\}$ value puts more weight on $\E\big[X \big| {\categZorC} \neq \categZorCsmall \big]$, suggesting that this probability can measure the credibility of the additional information. 

Turning to the accuracy of the conditional mean of the synthetic data relative to the original one, we present the following result. 

\begin{corollary}[Conditional Mean Gap]
\label{thm:con:mean:cor}
For any synthetic pair $(\tilde{X}, \tilde{\categZorC})$ generated from $({X}, {\categZorC})$ using the general-weight mixup, where $X$ is continuous and $\categZorC$ is categorical, the difference between the conditional mean  is given by
\begin{align}
    \nonumber
    &\big|
    \E\big[\tilde{X} \big| \tilde{\categZorC}=\categZorCsmall \big]
    - \E\big[X \big| \categZorC=\categZorCsmall \big]
    \big|
    \\&\qquad\qquad=\nonumber
    \big| u\big(W^X, W^\categZorC,\cutpoint\big) \big|
    \cdot \Pr\{ \categZorC \neq \categZorCsmall\}
    \\ &\label{eq:cond.mean.gap}
    \qquad\qquad\quad
    \cdot
    \big|
    \E\big[X \big| \categZorC=\categZorCsmall \big]
    - \E\big[X \big| \categZorC \neq \categZorCsmall \big]
    \big| .
\end{align} 
\end{corollary}
The right side of (\ref{eq:cond.mean.gap}) consists of three terms, where the first term can be controlled by the modeler and the remaining two terms are fixed for given data. Thus an important observation on this result is that the difference in the left side of (\ref{eq:cond.mean.gap}) can be made smaller by controlling the value of $u\big(W^X, W^\categZorC,\cutpoint\big)$ with suitably chosen weight random variables and the cut point. 

Assuming that the variance of the synthetic data is preserved, we can establish the following upper bound for  the conditional variance. 
\begin{theorem}[Conditional Variance Gap]
\label{thm:con:var}
    Assume that $\E\big[\big(W^X\big)^2\big] = \E\big[W^X \big]$. Then, for any synthetic pair  $(\tilde{X}, \tilde{\categZorC})$ generated from $({X}, {\categZorC})$ using the general-weight mixup, where $X$ is continuous and $\categZorC$ is categorical, the difference between the conditional variance  is bounded  as follows:
    \begin{align}
    \nonumber
    &\big|\Var\big[ \tilde{X}\big|  \tilde{\categZorC}=\categZorCsmall \big]
    -  \Var\big[ X \big| \categZorC=\categZorCsmall \big]\big|
    \\ &\qquad \nonumber
     \leq
    \big| u\big(W^X, W^\categZorC,\cutpoint\big)\big|
    \cdot \big| \Var\big[ X \big| \categZorC=\categZorCsmall \big] - \Var\big[X\big] \big|
    \\
    \nonumber
    &\qquad \quad +\big| u\big(W^X, W^\categZorC,\cutpoint\big)(1- u\big(W^X, W^\categZorC,\cutpoint\big)) \big|  
    \\
    \label{ineq.cond.var}
    &\qquad \qquad \quad
    \cdot \big(\E\big[ X \big| \categZorC=\categZorCsmall \big]-\E\big[ X \big]\big)^2.
    \end{align}
\end{theorem}

Similar to Corollary~\ref{thm:con:mean:cor}, one can make the difference of the conditional variance smaller by controlling the right side of (\ref{ineq.cond.var})  where the only quantity at the modeler's disposal is $u\big(W^X, W^\categZorC,\cutpoint\big)$. In Sec.~\ref{sec:mixup:distn} we propose a class of mixup weight distributions that has a explicit relationship to this function.

Our discussion so far shows that function $u\big(W^X, W^\categZorC,\cutpoint\big)$ in Def.~\ref{def:u:function} plays an important role in computing the conditional moments in the synthetic data. Thus we present two theoretical properties of this function before concluding this section. 

\begin{lemma}
    \label{thm:nonnegative:u}
   Under the standard mixup scheme, $u(W,\cutpoint)\in [0,1]$ holds for any $\cutpoint\in\mathbb{R}$ if $\E\big[W^2\big] = \E\big[W \big]$.
\end{lemma}

\begin{lemma}[Optimal Cut Point $\cutpoint$]
    \label{thm:condimean:cutpoint}
    Under the standard mixup scheme with $\E\big[W^2\big] = \E\big[W \big]$, the optimal cut point $\cutpoint$  is $0.5$. That is
    \begin{equation}
    0.5 = \argmin_{\cutpoint \in \mathbb{R}} \big| u\big(W,\cutpoint\big) \big| .
    \end{equation}
\end{lemma}

Lemma~\ref{thm:condimean:cutpoint} shows that $0.5$ minimizes $\big| u\big(W,\cutpoint\big) \big|$. It is natural to employ $\cutpoint=0.5$ for the general-weight mixup as well because this cut point choice is more likely to preserve the conditional mean and conditional variance as shown in Corollary~\ref{thm:con:mean:cor} and Theorem~\ref{thm:con:var}; we use $\cutpoint=0.5$ in what follows unless specified otherwise.

\begin{figure*}[ht]
     \centering
     \begin{subfigure}[b]{0.47\textwidth}
         \centering
         \includegraphics[width=\textwidth]{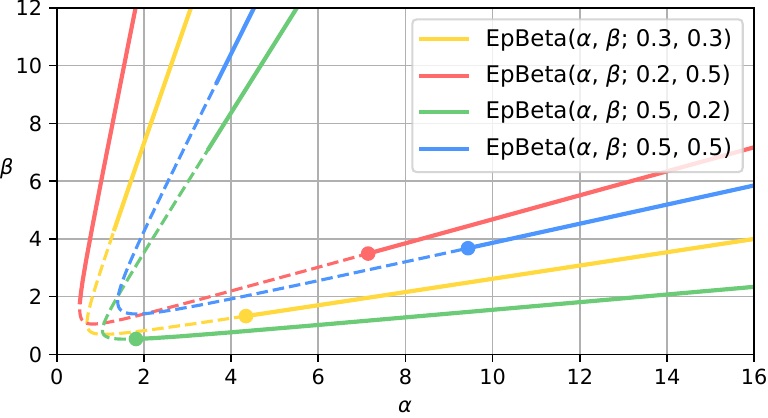}
     \end{subfigure}
     \hfill
     \begin{subfigure}[b]{0.49\textwidth}
         \centering
         \includegraphics[width=\textwidth]{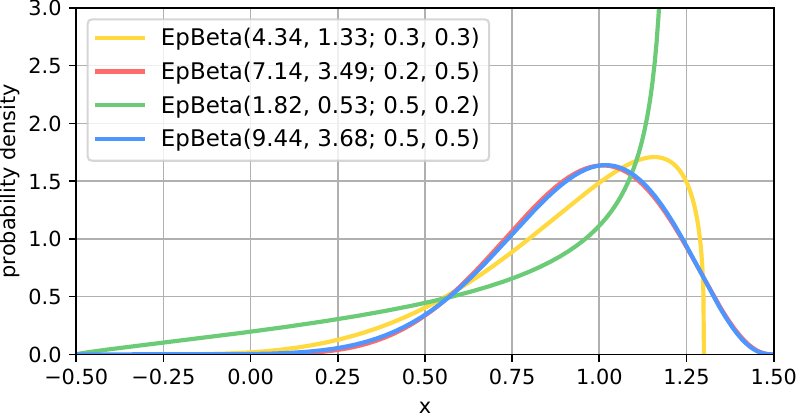}
     \end{subfigure}
     \caption{
     (Left) All $(\alpha,\beta)$ pairs that gives structure-preserving synthetic data for choices of $(\epsilon_0,\epsilon_1)$. 
 (Right) Density functions corresponding to the solid circled point of each curve on the left plot.
     }
     \label{fig:epbeta}
\end{figure*}

It is noted that categorical variable $\categZorC$ in this paper can be regarded as a multinomial variable. Therefore, the above results concerning $\categZorC$ are equally applicable for  multiple categorical variables, since combinations of multiple multinomial variables also follow the multinomial distribution only with more individual categories.

\section{STRUCTURE-PRESERVING MIXUP DISTRIBUTION\label{sec:mixup:distn}}

In this section we propose a new mixup scheme in data synthesis that can preserve key statistics such as the mean, variance and their conditional counterparts as discussed in Sec.~\ref{sec:properties}.

\subsection{Variance-Reduction Mixup}

In the synthetic data literature the most common choice for the mixup weight is to use a distribution defined on $[0,1]$ \citep{zhang2017mixup, verma2019manifold, cao2024survey}. Two prominent examples are the $\operatorname{Beta}(\alpha,\beta)$ distribution with $\alpha,\beta\in(0,\infty)$ and the $\operatorname{Unif}(0,1)$ distribution. Restricting the distribution's support to $[0,1]$ does not distort the mean of the synthetic data, but it reduces the variance inevitably as shown below, which can be found in, e.g., Proposition~1 in \citet{kim2024configuring}. 

\begin{corollary}[Variance-Reduction Mixup]
    \label{thm:bounded:support:mixup}
    For any synthetic variable $\tilde X$ generated by the mixup from a continuous $X$ in (\ref{eq:mixup:x}), let the support of mixup weight variable $W^X$ be bounded in $[0,1]$. Then 
    \begin{equation}
        \Var\big[\tilde{X}\big] \leq \Var\big[X\big],
    \end{equation}
    where the equality holds when $\Pr\big\{W^X \in \{0, 1\} \big\}=1$.
\end{corollary}
The condition $\Pr\big\{W^X \in \{0, 1\} \big\}=1$ implies that every synthetic instance is exactly same as one of the original instances, which essentially is a resampling scheme rather than data synthesis. Thus it is clear that the support of weight distribution must be expanded from $[0,1]$ in order to obtain variance-preserving synthetic data.

\subsection{Structure-Preserving Mixup}

The following simple example illustrates that the mixup weights generated from a constrained normal distribution with support $(-\infty, +\infty)$ can produce variance-preserving synthetic data. 

\begin{example}
\label{ex:weight:1}
Let the mixup weights are generated from the Gaussian distribution $\operatorname{N}(\mu, \sigma^2)$ where $\sigma = \sqrt{\mu-\mu^2}$ for some $\mu\in [0, 1]$, i.e., $W^X=W^Y \sim  \operatorname{N}(\mu, \mu-\mu^2)$. Then, under the standard mixup scheme,
 we have, for any pair $(X,Y)$,  $\Var\big[\tilde{X}\big] = \Var\big[X\big] $, $\Var\big[\tilde{Y}\big] = \Var\big[Y\big]$, and $\Cov\big[\tilde{X},\tilde{Y}\big] = \Cov\big[X,Y\big] $.
\end{example}

A problem with this example however is that it  can generate unacceptably extreme synthetic instances when the mixup weights take extreme values. One way to suppress extreme synthetic instances is to restrict the support of mixup weight distribution to a finite interval $[-\epsilon_0,1+\epsilon_1]$ for $\epsilon_0, \epsilon_1 \in [0,\infty)$, so that it always include $[0,1]$ as a sub-interval. To this extent we propose an expanded version of the standard beta distribution, also known as the four-parameter beta distribution \citep{johnson1995continuous}.

\begin{definition}[Expanded Beta Distribution]
    \label{def:epbeta}
    Let $\epsilon_0, \epsilon_1 \in [0,\infty)$ be given constants and $V\sim \operatorname{Beta}(\alpha, \beta)$. Then the random variable $W$, with support $[-\epsilon_0,1+\epsilon_1]$, is said to follow the expanded Beta distribution with parameters $(\alpha, \beta, \epsilon_0,\epsilon_1)$, or simply $W\sim \operatorname{EpBeta}(\alpha, \beta; \epsilon_0, \epsilon_1)$, if 
    \begin{equation}
        W = (1+\epsilon_0+\epsilon_1) V - \epsilon_0.
    \end{equation}
\end{definition}

The choice of $(\epsilon_0, \epsilon_1)$ in practice would reflect the meta-information or characteristics of the data. For example, if variable $X$ cannot be negative, we should set the parameter as  $\epsilon_0=0$ and $\epsilon_1> 0$, with a further constraint $X_i \leq X_j$ for selected instances in \eqref{eq:mixup:x} so that $\tilde{X}$ also remains positive. We also note that the size of $\epsilon_1$ can control  the maximum possible value of synthetic instances. To illustrate, consider an extreme case where the mixup weight is $1+\epsilon_1$ coinciding with the upper bound, and the two selected original instances are $x_{\max} = \max_{i\in[n]}\{x_i\}$ and $x_{\min} = \min_{i\in[n]}\{x_i\}$. Then the resulting synthetic instance is $ \tilde{x}_{k}=x_{\max} + \epsilon_1(x_{\max}-x_{\min})$, an extrapolation which leads to a much larger value than $x_{\max}$ by choosing a big $\epsilon_1$.

After choosing $(\epsilon_0, \epsilon_1)$ we can find some $\alpha, \beta$ that satisfy \eqref{eq:epbeta:constraint} in Theorem~\ref{thm:epbeta:var} and \eqref{eq:epbeta:conditional:constraint} in Theorem~\ref{thm:epbeta:condimean} to preserve data structure as follows.

\begin{theorem}
\label{thm:epbeta:var}
For given $\epsilon_0, \epsilon_1 \in [0,\infty)$, consider an arbitrary synthetic pair $(\tilde{X}, \tilde{Y})$ generated from $({X}, {Y})$ using the standard mixup scheme with  $W \sim \operatorname{EpBeta}(\alpha, \beta; \epsilon_0, \epsilon_1)$ 
, such that $\alpha, \beta \in (0, \infty)$ satisfy
\begin{align}
    \nonumber
    &(1+\epsilon_1 -\epsilon_0 (\beta/\alpha)) \cdot (1+\epsilon_0-\epsilon_1 (\alpha/\beta)) \cdot (1+\alpha+\beta)
    \\
    &\qquad
    = \label{eq:epbeta:constraint}
    (1+\epsilon_0+\epsilon_1)^2.
\end{align}
Then we have  $\Var\big[\tilde{X}\big] = \Var\big[X\big] $ and $\Cov\big[\tilde{X},\tilde{Y}\big] = \Cov\big[X,Y\big] $.
\end{theorem}

\begin{theorem}
    \label{thm:epbeta:condimean}
   Consider an arbitrary synthetic triple $(\tilde{X}, \tilde{Y}, \tilde{\categZorC})$ generated from $({X}, {Y}, \categZorC)$ using the standard mixup scheme with $W \sim \operatorname{EpBeta}(\alpha, \beta; \epsilon_0, \epsilon_1)$ for given $\epsilon_0, \epsilon_1 \in [0,\infty)$ and $\cutpoint=0.5$. Now suppose that, for a given  $\delta \in [0,1]$, $(\alpha, \beta)$ satisfies (\ref{eq:epbeta:constraint}) and the following 

    \begin{align}
    \nonumber
    \frac{ 1+\epsilon_0 -\epsilon_1 \alpha /\beta}{1+\alpha/\beta}
    + \frac{2(1+\epsilon_0+\epsilon_1)}{1+\beta/\alpha} \frac{B(\tilde{\epsilon};\alpha+1, \beta)}{B(1;\alpha+1, \beta)}&
    \\ \label{eq:epbeta:conditional:constraint}
    - (1+2\epsilon_0) \frac{B\left(\tilde{\epsilon};\alpha, \beta\right)}{B(1;\alpha, \beta)}
    &\leq
    \delta,
    \end{align}
    where $B(x;\alpha, \beta) = \int_{0}^{b} t^{\alpha-1} (1-t)^{\beta-1} \,dt$ is the incomplete beta function and $\tilde{\epsilon}=\frac{0.5+\epsilon_0}{1+\epsilon_0+\epsilon_1}$.
    
 Then, the gap of conditional (on categorical $\categZorC$) mean and variance are bounded as follows:
 \begin{align}
    \nonumber
    &\big|
    \E\big[\tilde{X} \big| \tilde{\categZorC}=\categZorCsmall \big]
    - \E\big[X \big| \categZorC=\categZorCsmall \big]
    \big|
    \\ \nonumber
    & \qquad
    =
    \delta
    \cdot \Pr\{ \categZorC \neq \categZorCsmall\}
    \cdot
    \big|
    \E\big[X \big| \categZorC=\categZorCsmall \big]
    - \E\big[X \big| \categZorC \neq \categZorCsmall \big]
    \big| 
 \end{align} and 
\begin{align}
    &\big|\Var\big[ \tilde{X}\big| \tilde{\categZorC}=\categZorCsmall \big]
    -  \Var\big[ X \big| \categZorC=\categZorCsmall \big]\big|
    \nonumber
    \\&\qquad \leq \nonumber
    \delta
    \cdot \big| \Var\big[ X \big| \categZorC=\categZorCsmall \big] - \Var\big[X\big] \big|
    \\
    &\qquad\quad \nonumber
    +\delta(1-\delta)
    \cdot \big(\E\big[ X \big| \categZorC=\categZorCsmall \big]-\E\big[ X \big]\big)^2.
\end{align} 
\end{theorem}

Theorem~\ref{thm:epbeta:condimean} implies that the magnitude of the gap can be controlled by $\delta$. In particular, the conditional mean and variance can be perfectly preserved as $\delta$ tends to $0$.
This result, coupled with Theorem~\ref{thm:epbeta:var}, suggests that $\delta$ can be viewed as a modulator that controls the amount of additional information to be borrowed from the original data.

On the left plot in Fig.~\ref{fig:epbeta}, each curve represents  $(\alpha,\beta)$ pairs that satisfy Theorem \ref{thm:epbeta:var} for a  selected $(\epsilon_0,\epsilon_1)$ choice; each point on the curve therefore produces structure-preserving synthetic data. The solid-line part of each curve further satisfies Theorem~\ref{thm:epbeta:condimean} with equality for $\delta=0.05$, so that the conditional mean and variance are almost preserved; for visual simplicity, we choose $(\alpha,\beta)$ pairs that make both sides of \eqref{eq:epbeta:conditional:constraint} equal  with $\alpha \geq \beta$. The right plot  in Fig.~\ref{fig:epbeta} shows the densities corresponding to each circled point shown on the left plot. For illustration Appendix~\ref{appendix:epbeta:parameters} presents  tables of $\operatorname{EpBeta}$ parameters $(\alpha, \beta)$ for additional values of $\epsilon_0$, $\epsilon_1$ and $\delta$, all of which result in structure-preserving synthetic data.

\begin{figure*}[ht]
    \centering
    \includegraphics[width=\linewidth]{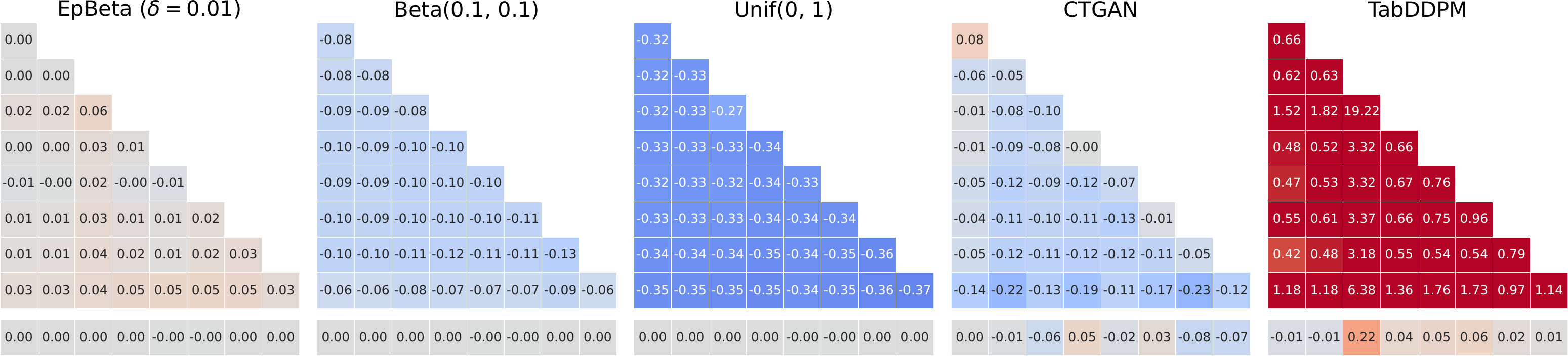}
    \caption{The relative bias of covariance (triangle) and expectation (bar at bottom) for the `Abalone' data using various synthetic generation methods. Negative bias is colored in blue and positive bias in red. Grey represents bias close to zero.}
    \label{fig:cov:matrix}
\end{figure*}

We note that structure-preserving mixup weight distributions can be defined from other distributions in a similar manner, for example, the truncated normal distribution. However, these alternative weight distributions do not enjoy all the theoretical benefits that the $\operatorname{EpBeta}$ does, as shown in this section.

\begin{algorithm}[tb]
\caption{Mixup Weight from $\operatorname{EpBeta}$}\label{alg}
\textbf{Input}: 
$\epsilon_0, \epsilon_1 \geq 0$ (Smaller values better preserve the support of synthetic instances), 
$\delta \geq 0$ (Smaller values better preserve conditional mean and variance) \\
\textbf{Output}: $w$ (Mixup weight) 

\begin{algorithmic}[1]
\STATE Identify pairs $(\alpha, \beta)$ that satisfy the following constraints:
$\alpha \geq \beta$, \eqref{eq:epbeta:constraint} in Theorem~\ref{thm:epbeta:var}, and \eqref{eq:epbeta:conditional:constraint} in Theorem~\ref{thm:epbeta:condimean} for the given $\epsilon_0$, $\epsilon_1$, and $\delta$.

\STATE Select the pair $(\alpha, \beta)$ for which $\alpha$ attains its minimum value.

\STATE Sample a Mixup weight $w$ from the $\operatorname{EpBeta}(\alpha, \beta; \epsilon_0, \epsilon_1)$ distribution.
\end{algorithmic}
\end{algorithm}

\paragraph{Guideline for Selecting Parameters.}

In practice, the user first specifies the possible ranges for $\epsilon_0$ and $\epsilon_1$, which determine the lower and upper bounds for the underlying distribution. Let $[x_l, x_u]$ be the conjectured bounds of the underlying distribution, satisfying $x_l \le x_{\min}$ and $x_u \ge x_{\max}$. Under this assumption, both $\epsilon_0$ and $\epsilon_1$ are set to $\epsilon_0 = \epsilon_1 = \frac{x_u - x_l}{x_{\max} - x_{\min}} - 1$. Next, the user specifies $\delta$ to control the tolerance on differences in the conditional mean and variance. With these parameters fixed, the values of $\alpha$ and $\beta$ are determined via Algorithm~\ref{alg}, which provides the weights used in mixup. 
An implementation of Algorithm~\ref{alg} is available at:
\url{https://github.com/leechungpa/structure-preserving-mixup}.

\section{EXPERIMENTS\label{sec:experiment}}

As in Sec~\ref{sec:mixup:distn}, we propose the $\operatorname{EpBeta}$ distribution as a mixup weight distribution that more effectively preserves the original data distribution when generating synthetic data. In this section, we demonstrate the importance of this distribution not only for ensuring consistent statistical inference but also for maintaining model performance. First, we emphasize its significance for tabular data, which is highly structured. Then, we apply the proposed mixup to image datasets, showing that the structure-preserving synthetic data sustain the model performance under repeated data synthesis.

\subsection{Tabular Data}

We synthesize $6$ different tabular datasets using three mixup methods ($\operatorname{EpBeta}$, $\operatorname{Beta}(0.1,0.1)$, and $\operatorname{Unif}(0, 1)$ and other four baseline  methods available in open-source code \citep{qian2023synthcity}; TVAE, CTGAN \citep{xu2019modeling}, TabDDPM \citep{kotelnikov2023tabddpm}, and GReaT \citep{borisov2022language}. The four number of $\operatorname{EpBeta}$ distribution parameter pairs have been selected so that it satisfies \eqref{eq:epbeta:constraint} and the equality condition \eqref{eq:epbeta:conditional:constraint} for given $\epsilon_0=\epsilon_1=0.3$ and $\delta = 0.001, 0.005, 0.01$ or $0.05$.
These $10$ synthetic datasets are evaluated and compared in terms of relative bias of key statistics, statistical inference, and the machine learning efficiency. Data descriptions and experimental details are in Appendix~\ref{appendix:exp:datasyn}.

\paragraph{Relative Bias.} We compare the relative bias of covariance and expectation from each synthetic data, calculated as $\frac{\Cov[\Tilde{X},\Tilde{Y}]-\Cov[X,Y]}{\Cov[X,Y]}$ and $\frac{\E[\Tilde{X}]-\E[X]}{\E[X]}$, respectively. As seen from Fig.~\ref{fig:cov:matrix}, the covariance gets reduced when we use the $\operatorname{Beta}$ or $\operatorname{Unif}$ as a mixup weight distribution, confirming Corollary~\ref{thm:bounded:support:mixup}. In contrast, the ML-based synthetic datasets (CTGAN and TabDDPM) substantially disturb the covariance and expectation. The proposed  $\operatorname{EpBeta}$ generates the most balanced synthetic data; the results of the other data are presented in Appendix~\ref{appendix:exp:bias}. This exercise shows that common data synthesis techniques are subject to considerable distortion in basic distributional quantities, which are often important in the early data exploration stage.

\begin{figure}[h]
    \centering
    \includegraphics[width=\linewidth]{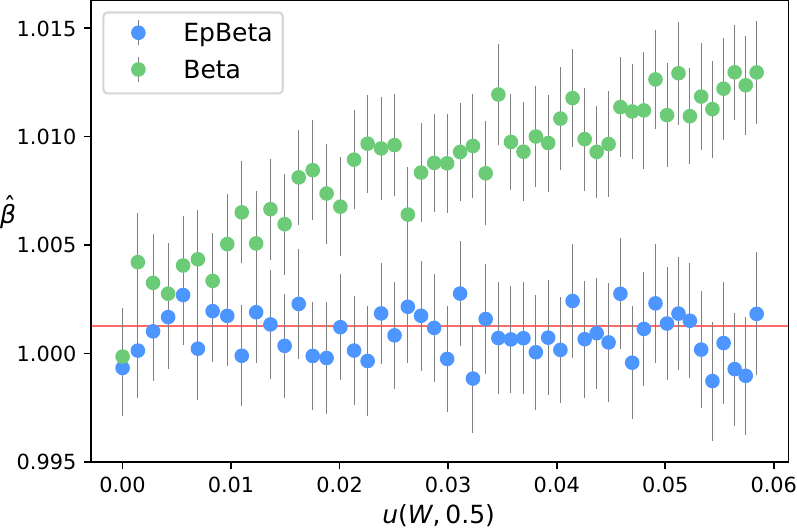}
    \caption{
    The estimated polynomial regression coefficient and its $95\%$ confidence interval for each synthetic data. The red horizontal line represents the coefficient estimate from the original data.
    }
    \label{fig:exp:beta}
\end{figure}

\paragraph{Statistical Inference.} We present a simple polynomial regression example that illustrates why preserving the variance is important from a statistical perspective. The regression analysis is a widely used tool in various studies. Specifically researchers are interested in accurate estimation of regression coefficients, which relate the target variable to the other attributes and allow an appropriate interpretation about their relationship. As previously mentioned, synthetic data generated by any mixup weight distribution preserves the coefficient of linear regression, as long as no polynomial terms are involved. For our experiment we draw $1000$ original instances with two variables $(X,Y)$ where $X\sim \operatorname{N}(5,1)$ and $Y\sim \operatorname{N}(X^2,1)$.
Then we fit a quadratic regression $Y=\beta X^2+e$ where $e\sim \operatorname{N}(0,\sigma^2)$, and obtain the coefficient estimate $\hat \beta$. From this original data, we generate synthetic data $(\tilde{X}, \tilde{Y})$ with same number of instances using, respectively, $\operatorname{EpBeta}$ and $\operatorname{Beta}$ as the weight distribution, which have various $u(W, 0.5)$ values. For each synthetic data we repeat the same regression fitting; if the synthesis is successful, the resulting coefficient estimates should be close to $\hat \beta$, the coefficient obtained from the original data. The result is presented in Fig.~\ref{fig:exp:beta} which shows that the proposed  $\operatorname{EpBeta}$  consistently produces coefficient estimates close to the true value $\hat \beta$, whereas $\operatorname{Beta}$ weight distribution yields coefficient estimates that are sometimes unacceptably distant from the true value, with  $\hat \beta$ sitting outside the confidence interval. This example elucidates that a data synthesis method that does not preserve the variance can fail even in rudimentary statistical analyses. We mention that $\operatorname{EpBeta}$ may produce biased estimators in different setting, but the bias can be reduced by using a small enough $\delta \in [0,1]$.

In Appendix~\ref{sec:statistical:classification}, we also include a classification example to highlight the significance of statistical inference.

\paragraph{Machine Learning Efficiency.}  To compare the machine learning efficiency we use synthetic datasets to train various models, such as CatBoost \citep{prokhorenkova2018catboost} and MLP, following the experiment protocol of \cite{gorishniy2021revisiting, zhao2021ctab, kotelnikov2023tabddpm}. In particular, each model has been trained using one of the synthetic datasets and tested against the original data. The focus here is on assessing how closely each synthetic data resembles the original data, rather than on effectiveness of the models. Tables in Appendix~\ref{appendix:exp:ml} show that the performance of the mixup-driven synthetic datasets is comparable to other ML-based synthetic methods.

\subsection{Image Data}

We synthesize image data using the mixup method, and demonstrate that preserving data structure can prevent model collapse, emphasizing its significance.

\paragraph{Model Collapse with Repeated Synthesis.}

Recent work in Nature discusses \textit{model collapse}, a phenomenon where models trained predominantly on synthetic data begin to forget rare information, leading to significant performance degradation as human-generated data becomes scarce \citep{shumailov2024ai}. They showed that training a language model with texts, followed by training a new model on the synthetic texts generated from the previously learned model over nine iterations, results in reduced performance.

We conduct similar experiments in the image domain to show that using $\operatorname{EpBeta}$ distribution is more effective at preventing model collapse compared to the original mixup which reduces variance. We use the mixup method on the CIFAR-10 dataset \citep{krizhevsky2009cifar} to create synthetic images and repeatedly synthesized from these generated images, using the $\operatorname{EpBeta}$ distribution with $\epsilon_0 = \epsilon_1 = 0.3$ and $\delta = 0.05$, and the $\operatorname{Unif}(0,1)$ distribution, respectively. We then train Resnet-18 on the synthesized images to classify the image labels, and evaluate top-1 accuracy on the original test set.

The baseline model, trained on the original dataset, achieves a top-1 accuracy of 79.55. As demonstrated in Table~\ref{table:resynthesis}, training with synthetic datasets generated using either the $\operatorname{EpBeta}$ or $\operatorname{Unif}$ distributions enhances model performance when the resynthesis process is limited to 10 iterations or fewer. However, beyond 20 iterations of resynthesis, the $\operatorname{EpBeta}$ distribution maintains consistent performance by preserving the original data structure, while the $\operatorname{Unif}(0,1)$ distribution results in substantial performance degradation. These findings suggest that structure-preserving synthetic data generation enables sustained model performance and mitigates the risk of model collapse.

\begin{table}[ht]
\caption{
Top-1 accuracy of image classification models trained on repeatedly synthesized data. Each cell reports the mean and standard deviation of top-1 accuracy across five independently trained models, each using distinct randomly generated synthetic datasets.
}
\label{table:resynthesis}
\begin{center}
\resizebox{0.5\textwidth}{!}{%
\begin{tabular}{c|ccccc}
\hline
Resynthesis & 5 & 10 & 15 & 20 & 25      \\
\hline\hline
$\operatorname{EpBeta}(\delta=0.05)$ & \begin{tabular}[c]{@{}l@{}}85.78\\ (0.13)\end{tabular} & \begin{tabular}[c]{@{}l@{}}85.99\\ (0.21)\end{tabular} & \begin{tabular}[c]{@{}l@{}}86.08\\ (0.28)\end{tabular} & \begin{tabular}[c]{@{}l@{}}86.43\\ (0.14)\end{tabular} & \begin{tabular}[c]{@{}l@{}}85.76\\ (0.26)\end{tabular} \\
$\operatorname{Unif}(0,1)$ & \begin{tabular}[c]{@{}l@{}}84.39\\ (0.29)\end{tabular} & \begin{tabular}[c]{@{}l@{}}83.83\\ (0.13)\end{tabular} & \begin{tabular}[c]{@{}l@{}}74.49\\ (0.53)\end{tabular} & \begin{tabular}[c]{@{}l@{}}21.81\\ (4.21)\end{tabular} & \begin{tabular}[c]{@{}l@{}}12.34\\ (1.07)\end{tabular} \\
\hline
\end{tabular}%
}
\end{center}
\end{table}

\section{CONCLUSION\label{sec:conclusion}}

This paper presents significant theoretical advancements in the context of the mixup method for data synthesis. With a focus on ensuring that synthetic data mirrors the original in all aspects, the primary contribution is the establishment of specific conditions that the mixup weight distribution must meet to preserve the original data’s structure, including its mean, variance, and their conditional counterparts. On the theoretical front, we derive conditions under which the (co)variance for any pair of continuous variables remains intact with mixup. Additionally, we prove that the mean and variance, when conditioned on a categorical variable, can be preserved within a defined error bound. To achieve this, we introduce a class of mixup weight distributions, called $\operatorname{EpBeta}$, a generalized form of the $\operatorname{Beta}$ distribution, which adheres to these theoretical conditions, thereby preserving the structural integrity of the original data. Our numerical experiments confirm that our proposed mixup method maintains essential distributional properties, leading to more accurate statistical inferences. In terms of performance, ours delivers results comparable to other synthetic data generation methods while significantly maintaining performance under repeated synthesis.

While our method primarily focuses on the classical mixup framework, its applicability extends to related data augmentation techniques such as CutMix~\citep{yun2019cutmix}, where mixup weights take binary values (0 or 1), and to its extensions like Gaussian-Mixup~\citep{NEURIPS2022_e6f32e64} which replaces discrete region constraints with continuous distributions. Exploring optimal parameterizations for such distributions remains a research direction that could further enhance the flexibility of the mixup methods.

In addition to these extensions, future research will explore non-equal weight mixup methods that aim to identify representative instances rather than selecting them uniformly at random. Moreover, we may consider more flexible weight distribution classes to further improve the quality of synthetic data while preserving essential statistical properties.

\subsubsection*{Acknowledgements}

This research is partially supported by the Information Technology Research Center (ITRC) support program (IITP-2024-RS-2023-00259004) supervised by the Institute for Information \& Communications Technology Planning \& Evaluation (IITP).
J.H.T. Kim also acknowledges the support from the National Research Foundation of Korea Grant funded by the Korean Government (NRF-2022R1F1A106357511).

\clearpage
\bibliography{__ref}

\clearpage
\section*{Checklist}

 \begin{enumerate}

 \item For all models and algorithms presented, check if you include:
 \begin{enumerate}
   \item A clear description of the mathematical setting, assumptions, algorithm, and/or model. [Yes] See Sec.~\ref{sec:problemsetup}.
   \item An analysis of the properties and complexity (time, space, sample size) of any algorithm. [Not Applicable]
 \end{enumerate}

 \item For any theoretical claim, check if you include:
 \begin{enumerate}
   \item Statements of the full set of assumptions of all theoretical results. [Yes] See Sec.~\ref{sec:problemsetup}.
   \item Complete proofs of all theoretical results. [Yes] See Appendix~\ref{appendix:proofs}.
   \item Clear explanations of any assumptions. [Yes] See Sec.~\ref{sec:problemsetup}.
 \end{enumerate}

 \item For all figures and tables that present empirical results, check if you include:
 \begin{enumerate}
   \item The code, data, and instructions needed to reproduce the main experimental results (either in the supplemental material or as a URL). [Yes] See Sec.~\ref{sec:experiment} and Appendix~\ref{appendix:experiment}. An implementation of Algorithm~\ref{alg} is available at:
\url{https://github.com/leechungpa/structure-preserving-mixup}.
   \item All the training details (e.g., data splits, hyperparameters, how they were chosen). [Yes] See Sec.~\ref{sec:experiment} and Appendix~\ref{appendix:experiment}.
     \item A clear definition of the specific measure or statistics and error bars (e.g., with respect to the random seed after running experiments multiple times). [Yes] See Sec.~\ref{sec:experiment} and Appendix~\ref{appendix:experiment}.
     \item A description of the computing infrastructure used. (e.g., type of GPUs, internal cluster, or cloud provider). [Yes] See Appendix~\ref{appendix:exp:datasyn}.
 \end{enumerate}

 \item If you are using existing assets (e.g., code, data, models) or curating/releasing new assets, check if you include:
 \begin{enumerate}
   \item Citations of the creator If your work uses existing assets. [Yes] See Sec.~\ref{sec:experiment} and Appendix~\ref{appendix:experiment}.
   \item The license information of the assets, if applicable. [Yes] See Table~\ref{table:data}.
   \item New assets either in the supplemental material or as a URL, if applicable. [Not Applicable]
   \item Information about consent from data providers/curators. [Not Applicable]
   \item Discussion of sensible content if applicable, e.g., personally identifiable information or offensive content. [Not Applicable]
 \end{enumerate}

 \item If you used crowdsourcing or conducted research with human subjects, check if you include:
 \begin{enumerate}
   \item The full text of instructions given to participants and screenshots. [Not Applicable]
   \item Descriptions of potential participant risks, with links to Institutional Review Board (IRB) approvals if applicable. [Not Applicable]
   \item The estimated hourly wage paid to participants and the total amount spent on participant compensation. [Not Applicable]
 \end{enumerate}

 \end{enumerate}

\clearpage
\appendix
\onecolumn

\section{PROOFS}
\label{appendix:proofs}

As expectation is preserved, the distribution of synthetic categorical variable is also preserved, by following a simple theorem.

\begin{theorem}[Category-Preserving Mixup]
    \label{thm:app:category:preserve}
    \leavevmode For any synthetic variable $\tilde \categZorC$ generated from $\categZorC$ in \eqref{eq:mixup:z}, the distribution is preserved. That is $\Pr\big\{ \categZorC= \categZorCsmall \big\} = \Pr\big\{\tilde{\categZorC}= \categZorCsmall \big\}$ for any $\categZorCsmall$.
\end{theorem}
\begin{proof}
It can be directly proved by the definition in \eqref{eq:mixup:z}, as follow.
\begin{align*}
     \Pr\big\{\tilde{\categZorC}= z\big\}
     &=
     \Pr\big\{ \categZorC_i= \categZorCsmall\big\}\Pr\big\{ W^\categZorC\geq \cutpoint\big\} 
     + \Pr\big\{ \categZorC_j= \categZorCsmall\big\}\Pr\big\{ W^\categZorC < \cutpoint\big\}
     \\ &=
     \Pr\big\{ \categZorC= \categZorCsmall\big\} \big( \Pr\big\{ W^\categZorC\geq \cutpoint\big\} + \Pr\big\{ W^\categZorC < \cutpoint\big\} \big)
     \\ &=
     \Pr\big\{ \categZorC= \categZorCsmall\big\}
\end{align*}
\end{proof}

\subsection{Regression Coefficients Are Preserved}
\label{appendix:proofs:regcoef}

\begin{theorem}
    \label{thm:app:regression}
    For any synthetic pair  $(\tilde{X}, \tilde{Y})$ generated from $({X}, {Y})$ using the standard mixup scheme, the regression coefficients are preserved. I.e.,
    \begin{equation}
    \argmin_{\beta_0, \beta_1} \E\big\| Y -\beta_0 -\beta_1 X \big\|_2^2 = \argmin_{\beta_0, \beta_1} \E\big\| \Tilde{Y} -\beta_0 -\beta_1 \Tilde{X} \big\|_2^2 \nonumber %
    .
    \end{equation}    
\end{theorem}
\begin{proof}
    It is well known that $\hat{\beta}_0 = \E\big[Y] -\hat{\beta}_1  \E\big[ X \big]$ and $\hat{\beta}_1=\Corr \big[ X,Y \big]$ from the convex optimization:
    \begin{align*}
        0
        &=
        \frac{\partial}{\partial\beta_0}\E\big\| Y -\beta_0 -\beta_1 X\big\|_2^2 \bigg|_{(\beta_0, \beta_1)=(\hat{\beta}_0,\hat{\beta}_1)}
        \\ &=
        2\hat{\beta}_0 - 2 \E\big[Y\big] -\hat{\beta}_1 \E\big[ X \big],
        \\
        0
        &=
        \frac{\partial}{\partial\beta_1}\E\big\| Y -\beta_0 -\beta_1 X \big\|_2^2 \bigg|_{(\beta_0, \beta_1)=(\hat{\beta}_0,\hat{\beta}_1)}
        \\
        &=
        2\E\big[X^2\big]\hat{\beta}_1 -2 \E \big[ X(Y-\hat{\beta}_0) \big]
        \\
        &=
        2\E\big[(X-\E\big[ X \big])^2\big]\hat{\beta}_1 -2 \E \big[ (X-\E\big[X] ) (Y-\E\big[Y] ) \big].
    \end{align*}
    By Corollary~\ref{thm:app:corr}, the correlation is preserved under the standard mixup scheme. Moreover, expectation is always preserved.  Therefore,
    \begin{align*}
        \argmin_{\beta_0, \beta_1} \E\big\| Y -\beta_0 -\beta_1 X \big\|_2^2
        &= \big(\E\big[Y] -\Corr\big[X, Y\big]  \E\big[ X \big], \; \Corr\big[X, Y\big]\big)
        \\
        &= \big(\E\big[\Tilde{X}\big] - \Corr\big[\Tilde{X}, \Tilde{Y}\big]  \E\big[ \Tilde{X} \big], \; \Corr\big[\Tilde{X}, \Tilde{Y}\big]\big)
        \\
        &= \argmin_{\beta_0, \beta_1} \E\big\| \Tilde{Y} -\beta_0 -\beta_1 \Tilde{X} \big\|_2^2.
    \end{align*}
    This can be easily generalized to multiple linear regression. For simplicity, assume that the variables are centered. Define $\mathbf{X}\in\mathbb{R}^{n\times p}$, $\mathbf{Y}\in\mathbb{R}^{n\times 1}$, and  $\beta\in\mathbb{R}^{p\times 1}$ where $\hat{\mathbf{\beta}} =
    \argmin_\mathbf{\beta} \E\big\| \mathbf{Y} - \mathbf{X} \mathbf{\beta} \big\|_2^2$. It is well known that $\hat{\mathbf{\beta}} = (\mathbf{X}^\top \mathbf{X})^{-1} \mathbf{X}^\top \mathbf{Y}$. Then, the coefficients are preserved as below,
    \begin{align}
    \hat{\mathbf{\beta}} \nonumber
    &= \nonumber
    (\mathbf{X}^\top \mathbf{X})^{-1} \mathbf{X}^\top \mathbf{Y}
    \\ 
    &= \nonumber
    \big( (\mathbf{X}-\E[\mathbf{X}])^\top (\mathbf{X}-\E[\mathbf{X}]) \big)^{-1}
    \cdot (\mathbf{X}-\E[\mathbf{X}])^\top (\mathbf{Y}-\E[\mathbf{Y}])
    \\ 
    &=
    \bigg( \big( 1 + 2\E\big[ \big(W^X\big)^2 -W^X \big] \big) (\mathbf{X}-\E[\mathbf{X}])^\top (\mathbf{X}-\E[\mathbf{X}]) \bigg)^{-1}
    \cdot \big( 1 + 2\E\big[ \big(W^X\big)^2 -W^X \big] \big) (\mathbf{X}-\E[\mathbf{X}])^\top (\mathbf{Y}-\E[\mathbf{Y}])
    \label{eq:regression:proof}
    \\ 
    &= \nonumber
    (\Tilde{\mathbf{X}}^\top \Tilde{\mathbf{X}})^{-1} \Tilde{\mathbf{X}}^\top \Tilde{\mathbf{Y}}
    \end{align}
    where the equality in \eqref{eq:regression:proof} holds from Corollary~\ref{thm:app:corr}. These results can be easily extended to the non-centered case when an intercept term is included in the regression, because mixup always preserves the expectation.
\end{proof}

\subsection{Proofs for Sec.~\ref{sec:properties}}
\label{appendix:proof:sec2}

\begin{lemma}[Variance]
\label{thm:app:var}
\leavevmode For any synthetic $\tilde X$ generated 
from $X$ in (\ref{eq:mixup:x}):
\begin{enumerate}
    \item  $\Var\big[\tilde{X}\big] = \Var\big[X\big]$, if and only if the first and second moments of mixup weight are equal. That is
\begin{equation}
    \E\big[\big(W^X\big)^2\big] = \E\big[W^X \big].
\end{equation}

    \item $\Var\big[\tilde{X}\big] <  \Var\big[X\big]$, 
     if and only if $\E\big[\big(W^X\big)^2\big] < \E\big[W^X \big]$.

    \item $\Var\big[\tilde{X}\big] > \Var\big[X\big]$.
    if and only if $\E\big[\big(W^X\big)^2\big] > \E\big[W^X \big]$.
\end{enumerate}
\end{lemma}

\begin{proof}
    Since $\E\big[\tilde{X}\big]=\E\big[X\big]$, the variance of synthetic data $\Var\big[\tilde{X}\big]$ is determined as
\begin{align}
    \Var\big[\tilde{X}\big]
    &= \nonumber
    \E\big[ W^X X_i + \big(1-W^X\big) X_j \big]^2 - \E\big[X\big]^2
    \\ &= \nonumber
    \E\big[\big(W^X\big)^2\big]\E\big[X_i^2\big]
    + \E\big[\big(1-W^X\big)^2\big]\E\big[X_j^2\big]
    + 2\E\big[W^X\big(1-W^X\big)\big]\E\big[X_i\big]\E\big[X_j\big]
    - \E\big[X\big]^2
    \\ &= \nonumber
    \E\big[\big(W^X\big)^2 + \big(1-W^X\big)^2\big](\Var\big[X\big]+\E\big[X\big]^2)
    + 2\E\big[W^X\big(1-W^X\big)\big]\E\big[X\big]^2
    - \E\big[X\big]^2
    \\&= 
    \label{eq:var:condition:full:line:proof}
    \Var\big[X\big]+
    2 \E\big[\big(W^X\big)^2 - W^X \big] \Var\big[X\big].
\end{align}
Therefore, $\Var\big[\tilde{X}\big]=\Var\big[X\big]$ holds if and only if $\E\Big[\big(W^X\big)^2 - W^X \Big]=0$.
The cases of variance reduction and inflation can be trivially shown using a similar approach.
\end{proof}

\begin{theorem}[Covariance]
\label{thm:app:cov}
\leavevmode For any synthetic pair $(\tilde{X}, \tilde{Y})$ generated from $({X}, {Y})$ using the general-weight mixup in \eqref{eq:mixup:x} and \eqref{eq:mixup:y}:
\begin{enumerate}
    \item  $\Cov\big[\tilde{X}, \tilde{Y}\big] = \Cov\big[X, Y\big]$, if and only if
    \begin{equation}
        \E\big[ W^X W^Y \big] = \frac{1}{2} \big(\E\big[ W^X  \big]+ \E\big[W^Y \big]\big).
    \end{equation}

    \item $\Cov\big[\tilde{X}, \tilde{Y}\big] < \Cov\big[X, Y\big]$,
     if and only if $\E\big[ W^X W^Y \big] < \frac{1}{2} \big(\E\big[ W^X  \big]+ \E\big[W^Y \big]\big)$.
    
    \item $\Cov\big[\tilde{X}, \tilde{Y}\big] > \Cov\big[X, Y\big]$, 
 if and only if $\E\big[ W^X W^Y \big] > \frac{1}{2} \big(\E\big[ W^X  \big]+ \E\big[W^Y \big]\big)$.
\end{enumerate}
\end{theorem}
\begin{proof}
    Since $\E\big[\tilde{X}\big]=\E\big[X\big]$ and $\E\big[\tilde{Y}\big]=\E\big[Y\big]$, the covariance of synthetic data $\Cov\big[\tilde{X}, \tilde{Y}\big]$ is determined as
\begin{align}
    \Cov\big[\tilde{X}, \tilde{Y}\big] 
    &= \nonumber
    \E\big[\tilde{X}\tilde{Y}\big] -\E\big[\tilde{X}\big]\E\big[\tilde{Y}\big]
    \\&= \nonumber
    \E\big[
        W^X W^Y X_i Y_i
        + (1-W^X) W^Y X_j Y_i
    \big]
    \E\big[
        W^X (1-W^Y) X_i Y_j
        + (1-W^X) (1-W^Y) X_j Y_j
    \big]
    \\& \quad \nonumber
    - \E\big[X\big]\E\big[Y\big]   
    \\&= \nonumber
    \E\big[
        2W^X W^Y + 1-W^X -W^Y 
    \big] \E\big[ X_i Y_i \big]
    -
    \E\big[
         2W^X W^Y
         + 1
        -W^X -W^Y
    \big] \E\big[X\big]\E\big[Y\big]    
    \\&= \nonumber
     \E\big[
        2W^X W^Y + 1-W^X -W^Y 
    \big]
    \Cov\big[X, Y\big]   
    \\&= \label{eq:cov:condition:full:line}
     \Cov\big[X, Y\big] + \E\big[
        2W^X W^Y -W^X -W^Y 
    \big]
    \Cov\big[X, Y\big].
\end{align}
Therefore, $\Cov\big[\tilde{X}, \tilde{Y}\big] = \Cov\big[X, Y\big]$ if and only if $\E\big[ 2W^X W^Y -W^X -W^Y \big] = 0$, which is equal to Eq.~\eqref{eq:necessary:cov}.
The cases of covariance reduction and inflation can be trivially shown using a similar approach.
\end{proof}

\clearpage
\begin{corollary}[Correlation]
\label{thm:app:corr}
     For any synthetic pair $(\tilde{X}, \tilde{Y})$ generated from $(X,Y)$ using the standard mixup scheme,
     we have
    \begin{equation}
    \nonumber
    \Corr\big[\tilde{X}, \tilde{Y}\big] = \Corr\big[X, Y\big].
    \end{equation}
\end{corollary}

\begin{proof}
From \eqref{eq:cov:condition:full:line} in Theorem~\ref{thm:app:cov} with using the standard mixup scheme, i.e., $W^X_k=W^Y_k$ for all $k\in[m]$, the covariance of synthetic data $\Cov\big[\tilde{X}, \tilde{Y}\big]$ is determined as
\begin{align*}
    \Cov\big[\tilde{X}, \tilde{Y}\big] 
    &=
    \Cov\big[X, Y\big] + \E\big[2W^X W^Y -W^X -W^Y\big]
    \Cov\big[X, Y\big] 
    \\ &=
    \Cov\big[X, Y\big] + 2\E\big[
        \big(W^X\big)^2 -W^X 
    \big]
    \Cov\big[X, Y\big]  
    \\ &=
     \big( 1 + 2\E\big[
        \big(W^X\big)^2 -W^X 
    \big] \big)
    \Cov\big[X, Y\big].
\end{align*}
As a result, using \eqref{eq:var:condition:full:line:proof} in Lemma~\ref{thm:app:var}, the correlation of synthetic data is preserved as 
\begin{align*}
    \Corr\big[\tilde{X}, \tilde{Y}\big]
    &=
    \frac{\Cov\big[\tilde{X}, \tilde{Y}\big]}{\sqrt{\Var\big[\tilde{X}\big] \Var\big[ \tilde{Y}\big]}}
    \\ &=
    \frac{
    \big( 1 + 2\E\big[ \big(W^X\big)^2 -W^X \big] \big)
    \Cov\big[X, Y\big]
    }{
    \sqrt{
    \big( 1 + 2\E\big[ \big(W^X\big)^2 -W^X \big] \big)^2
    \Var\big[X\big] \Var\big[ Y\big]
    }
    }
    \\ &=
    \frac{
    \big( 1 + 2\E\big[ \big(W^X\big)^2 -W^X \big] \big)
    \Cov\big[X, Y\big]
    }{
     \big| 1 + 2\E\big[ \big(W^X\big)^2 -W^X \big] \big|
    \sqrt{   
    \Var\big[X\big] \Var\big[ Y\big]
    }
    }
    \\ &=
     \Corr\big[X, Y\big].
\end{align*}
Note that the last equality comes from the fact that
\begin{align*}
    1 + 2\E\big[ \big(W^X\big)^2 -W^X \big]
    &=
    \E\big[ \big(W^X\big)^2 \big]
    + \E\big[ \big(W^X-1\big)^2 \big]
    \geq 0.
\end{align*}
\end{proof}

\begin{theorem}[Conditional Mean]
\label{thm:app:con:mean} 
    For any synthetic pair  $(\tilde{X}, \tilde{\categZorC})$ generated from $({X}, {\categZorC})$ using the general-weight mixup, where $X$ is continuous and $\categZorC$ is categorical, the synthetic conditional mean $\E\big[\tilde{X} \big| \tilde{\categZorC}=\categZorCsmall \big]$ can be expressed as
    \begin{align}
        \nonumber
        \E\big[\tilde{X} \big| \tilde{\categZorC}=\categZorCsmall \big] 
        &= (1-u\big(W^X, W^\categZorC,\cutpoint\big))\E\big[X \big| \categZorC=\categZorCsmall \big]
        + u\big(W^X, W^\categZorC,\cutpoint\big) \E\big[X \big],
    \end{align}
or, alternatively, 
       \begin{align}
        \E\big[\tilde{X} \big| \tilde{\categZorC}=\categZorCsmall \big] 
        &= \nonumber
        \big(1-u\big(W^X, W^\categZorC,\cutpoint\big) \Pr\{ \categZorC \neq \categZorCsmall\} \big)\E\big[X \big| \categZorC=\categZorCsmall \big] 
        + u\big(W^X, W^\categZorC,\cutpoint\big) \Pr\{ \categZorC \neq \categZorCsmall\} E[X|\categZorC \neq \categZorCsmall].    
    \end{align}
\end{theorem}

\begin{corollary}[Conditional Mean Gap]
\label{thm:app:con:mean:cor}
For any synthetic pair $(\tilde{X}, \tilde{\categZorC})$ generated from $({X}, {\categZorC})$ using the general-weight mixup, where $X$ is continuous and $\categZorC$ is categorical, the difference between the conditional mean  is given by
\begin{align}
    \big|
    \E\big[\tilde{X} \big| \tilde{\categZorC}=\categZorCsmall \big]
    - \E\big[X \big| \categZorC=\categZorCsmall \big]
    \big|
    =
    \big| u\big(W^X, W^\categZorC,\cutpoint\big) \big|
    \cdot \Pr\{ \categZorC \neq \categZorCsmall\}
    \cdot
    \big|
    \E\big[X \big| \categZorC=\categZorCsmall \big]
    - \E\big[X \big| \categZorC \neq \categZorCsmall \big]
    \big| .
\end{align}
\end{corollary}
\begin{proof}[Proof of Theorem~\ref{thm:app:con:mean} and Corollary~\ref{thm:app:con:mean:cor}]
Let us define the sets $A_1 = \{\categZorC_i=\categZorCsmall, W^\categZorC \geq \cutpoint\}$ and $A_2 =\{\categZorC_j=\categZorCsmall, W^\categZorC < \cutpoint\}$, which disjointly divide the set $\big\{\tilde{\categZorC}=\categZorCsmall\big\} = A_1 \dot{\cup} A_2$. Then, $\Pr(A_1)= \Pr\{\categZorC_i=\categZorCsmall\} \Pr\{ W^\categZorC\geq \cutpoint\}$ and $\Pr(A_2)= \Pr\{\categZorC_j=\categZorCsmall\} \Pr\{ W^\categZorC < \cutpoint\}$.

To obtain $\E\big[\tilde{X} \big| \tilde{\categZorC}=\categZorCsmall \big] = \E\big[\tilde{X} \big| A_1 \cup A_2 \big]$, we first determine $\E\big[ \tilde{X}  \big| A_1 \big]$ and $\E\big[ \tilde{X}  \big| A_2 \big]$ as below:
\begin{align*}
    \E\big[ \tilde{X}  \big| A_1 \big]
    &=
    \E\big[ W^X X_i + (1-W^X) X_j \big| \{\categZorC_i=\categZorCsmall, W^\categZorC \geq \cutpoint\} \big]
    \\ &= 
    \E\big[ W^X \big| W^\categZorC \geq \cutpoint \big]
    \E\big[ X_i \big| \categZorC_i=\categZorCsmall \big]
    + \big( 1-\E\big[ W^X \big| W^\categZorC \geq \cutpoint \big] \big)
    \E\big[ X_j \big]
    \\ &= 
    \E\big[ W^X \big| W^\categZorC \geq \cutpoint \big]
    \E\big[ X \big| \categZorC=\categZorCsmall \big]
    + \big( 1-\E\big[ W^X \big| W^\categZorC \geq \cutpoint \big] \big)
    \E\big[ X \big],
    \\
    \E\big[ \tilde{X}  \big| A_2 \big]
    &= \E\big[ W^X \big| W^\categZorC < \cutpoint \big]
    \E\big[ X \big]
    + \big( 1-\E\big[ W^X \big| W^\categZorC < \cutpoint \big] \big)
    \E\big[ X  \big| \categZorC=\categZorCsmall  \big].
\end{align*}
Therefore, the mean of the continuous variable
conditioned by the categorical variable from the synthetic data is
\begin{align*}
    \E\big[\tilde{X} \big| \tilde{\categZorC}=\categZorCsmall \big] 
    &= 
    \E\big[\tilde{X} \big| A_1 \cup A_2 \big]
    \\ &= 
    \E\big[\E\big[ \tilde{X}  \big| A_i \big] \big| A_1 \cup A_2 \big]
    \\ &= 
    \E\big[ \tilde{X}  \big| A_1 \big] \Pr\big[ A_1 \big| A_1 \cup A_2 \big]
    +
    \E\big[ \tilde{X}  \big| A_2 \big] \Pr\big[ A_2 \big| A_1 \cup A_2 \big]
    \\ &= 
    \frac{
    \E\big[ \tilde{X}  \big| A_1 \big] \Pr\big[ A_1 \big]
    +
    \E\big[ \tilde{X}  \big| A_2 \big] \Pr\big[ A_2\big]
    }{Pr\big[ A_1 \cup A_2 \big]}
    \\ &= 
    \frac{
    \E\big[ \tilde{X}  \big| A_1 \big] \Pr\big[ A_1 \big]
    +
    \E\big[ \tilde{X}  \big| A_2 \big] \Pr\big[ A_2\big]
    }{\Pr\big[ A_1 \big] + \Pr\big[ A_2 \big]}
    \\ &= 
    \frac{
    \E\big[ \tilde{X}  \big| A_1 \big] \Pr\{\categZorC_i=\categZorCsmall\} \Pr\{ W^\categZorC\geq \cutpoint\}
    +
    \E\big[ \tilde{X}  \big| A_2 \big]  \Pr\{\categZorC_j=\categZorCsmall\}\Pr\{ W^\categZorC < \cutpoint\}
    }{\Pr\{\categZorC_i=\categZorCsmall\} \Pr\{ W^\categZorC\geq \cutpoint\} + \Pr\{\categZorC_j=\categZorCsmall\} \Pr\{ W^\categZorC< \cutpoint\}}
    \\ &= 
    \frac{
        \E\big[ \tilde{X}  \big| A_1 \big] \Pr\{ W^\categZorC\geq \cutpoint\}
        +
        \E\big[ \tilde{X}  \big| A_2 \big] \Pr\{ W^\categZorC < \cutpoint\}
    }{\Pr\{ W^\categZorC\geq \cutpoint\} + \Pr\{ W^\categZorC< \cutpoint\}}
    \\ &= 
    \E\big[ \tilde{X}  \big| A_1 \big] \Pr\{ W^\categZorC\geq \cutpoint\}
    +
    \E\big[ \tilde{X}  \big| A_2 \big] \Pr\{ W^\categZorC < \cutpoint\}
    \\ &=
    \big( 
    \E\big[ W^X \big| W^\categZorC \geq \cutpoint \big]
    \E\big[ X \big| \categZorC=\categZorCsmall \big]
    + \big( 1-\E\big[ W^X \big| W^\categZorC \geq \cutpoint \big] \big)
    \E\big[ X \big]
    \big)
    \Pr\{ W^\categZorC\geq \cutpoint\}    
    \\ &\quad +
    \big(
    \E\big[ W^X \big| W^\categZorC < \cutpoint \big]
    \E\big[ X \big]
    + \big( 1-\E\big[ W^X \big| W^\categZorC < \cutpoint \big] \big)
    \E\big[ X  \big| \categZorC=\categZorCsmall  \big] 
    \big)
    \big(
    1- \Pr\{ W^\categZorC \geq \cutpoint\}
    \big)
    \\ &=
    (1-u\big(W^X, W^\categZorC,\cutpoint\big))\E\big[X \big| \categZorC=\categZorCsmall \big]
    + u\big(W^X, W^\categZorC,\cutpoint\big) \E\big[X \big],
\end{align*}
where $u\big(W^X, W^\categZorC,\cutpoint\big)$ is defined as
\begin{align*}
    u\big(W^X, W^\categZorC,\cutpoint\big)
    &=
    \Pr\{ W^\categZorC\geq \cutpoint\}
    - \E\big[ W^X \big| W^\categZorC \geq \cutpoint \big]\Pr\{ W^\categZorC\geq \cutpoint\}
    \\ & \quad
    + \E\big[ W^X \big| W^\categZorC < \cutpoint \big]
    - \E\big[ W^X \big| W^\categZorC < \cutpoint \big]\Pr\{ W^\categZorC\geq \cutpoint\}
    \\ &=
    \E\big[ 1-W^X \big| W^\categZorC \geq \cutpoint \big]\Pr\{ W^\categZorC\geq \cutpoint\}
    + \E\big[ W^X \big| W^\categZorC < \cutpoint \big] \Pr\{ W^\categZorC < \cutpoint\}
    \\ &=
    \E\big[ \big(1-W^X\big) \mathbf{I} \{ W^\categZorC\geq \cutpoint\} \big]
    + \E\big[ W^X  \mathbf{I} \{ W^\categZorC < \cutpoint\} \big]
    \\ &=
    \E\big[ \big(1-W^X\big) \mathbf{I} \{ W^\categZorC\geq \cutpoint\} + W^X  \mathbf{I} \{ W^\categZorC < \cutpoint\} \big],
\end{align*}
and  $\mathbf{I}$ is an indicator function.

Moreover, the law of total expectation, 
\begin{align*}
    E[X]
    &=
    E[E[X|Y]]
    =
    E[X|\categZorC =\categZorCsmall ]\Pr\{ \categZorC =\categZorCsmall \} + 
    E[X|\categZorC \neq \categZorCsmall]\Pr\{ \categZorC \neq \categZorCsmall\},
\end{align*}
implies that 
\begin{align*}
    \E\big[\tilde{X} \big| \tilde{\categZorC}=\categZorCsmall \big] 
    &=
    (1-u\big(W^X, W^\categZorC,\cutpoint\big))\E\big[X \big| \categZorC=\categZorCsmall \big]
    + u\big(W^X, W^\categZorC,\cutpoint\big) E[X|\categZorC =\categZorCsmall ]\Pr\{ \categZorC =\categZorCsmall \} 
    \\ 
    &\qquad + 
    u\big(W^X, W^\categZorC,\cutpoint\big) E[X|\categZorC \neq \categZorCsmall] \Pr\{ \categZorC \neq \categZorCsmall\} 
    \\
    &=
    \big(1-u\big(W^X, W^\categZorC,\cutpoint\big) \Pr\{ \categZorC \neq \categZorCsmall\} \big)\E\big[X \big| \categZorC=\categZorCsmall \big]
    + u\big(W^X, W^\categZorC,\cutpoint\big) \Pr\{ \categZorC \neq \categZorCsmall\} E[X|\categZorC \neq \categZorCsmall],
\end{align*}
which is equal to
\begin{align*}
    \big|
    \E\big[\tilde{X} \big| \tilde{\categZorC}=\categZorCsmall \big]
    - \E\big[X \big| \categZorC=\categZorCsmall \big]
    \big|
    =
    \big| u\big(W^X, W^\categZorC,\cutpoint\big) \big|
    \cdot \Pr\{ \categZorC \neq \categZorCsmall\}
    \cdot
    \big|
    \E\big[X \big| \categZorC=\categZorCsmall \big]
    - \E\big[X \big| \categZorC \neq \categZorCsmall \big]
    \big| .
\end{align*}
\end{proof}

\clearpage
\begin{theorem}[Conditional Variance Gap]
\label{thm:app:con:var}
    Assume that $\E\big[\big(W^X\big)^2\big] = \E\big[W^X \big]$. Then, for any synthetic pair  $(\tilde{X}, \tilde{\categZorC})$ generated from $({X}, {\categZorC})$ using the general-weight mixup, where $X$ is continuous and $\categZorC$ is categorical, the difference between the conditional variance  is bounded  as follows:
    \begin{align}
    \nonumber
    \big|\Var\big[ \tilde{X}\big| \tilde{\categZorC}=\categZorCsmall \big]
    -  \Var\big[ X \big| \categZorC=\categZorCsmall \big]\big|
     & \leq
    \big| u\big(W^X, W^\categZorC,\cutpoint\big)\big|
    \cdot \big| \Var\big[ X \big| \categZorC=\categZorCsmall \big] - \Var\big[X\big] \big|
    \\
    \nonumber
    &\qquad +\big| u\big(W^X, W^\categZorC,\cutpoint\big)(1- u\big(W^X, W^\categZorC,\cutpoint\big)) \big|  
    \cdot \big(\E\big[ X \big| \categZorC=\categZorCsmall \big]-\E\big[ X \big]\big)^2.
    \end{align}
\end{theorem}
\begin{proof}
Let us define the sets $A_1 = \{\categZorC_i=\categZorCsmall, W^\categZorC \geq \cutpoint\}$ and $A_2 =\{\categZorC_j=\categZorCsmall, W^\categZorC < \cutpoint\}$, which disjointly divide the set $\big\{\tilde{\categZorC}=\categZorCsmall\big\} = A_1 \dot{\cup} A_2$. Then, $\Pr(A_1)= \Pr\{\categZorC_i=\categZorCsmall\} \Pr\{ W^\categZorC\geq \cutpoint \}$ and $\Pr(A_2)= \Pr\{\categZorC_j=\categZorCsmall\} \Pr\{ W^\categZorC < \cutpoint\}$.

To obtain $\E\big[\tilde{X}^2 \big| \tilde{\categZorC}=\categZorCsmall \big] = \E\big[\tilde{X}^2 \big| A_1 \cup A_2 \big]$, we first determine $\E\big[ \tilde{X}^2  \big| A_1 \big]$ and $\E\big[ \tilde{X}^2  \big| A_2 \big]$ as follows:
\begin{align*}
    \E\big[ \tilde{X}^2  \big| A_1 \big]
    &=
    \E\big[ (W^X)^2 X_i^2 + (1-W^X)^2 X_j^2 + 2 W^X (1-W^X) X_iX_j \big| \{\categZorC_i=\categZorCsmall, W^\categZorC \geq \cutpoint\} \big]
    \\ &= 
    \E\big[ (W^X)^2 \big| W^\categZorC \geq \cutpoint \big]
    \E\big[ X_i^2 \big| \categZorC_i=\categZorCsmall \big]
    +\E\big[ (1-W^X)^2 \big| W^\categZorC \geq \cutpoint \big]
    \E\big[ X_j^2 \big]
    \\&  \quad + 
    2\E\big[ W^X(1-W^X) \big| W^\categZorC \geq \cutpoint \big]
    \E\big[ X_i \big| \categZorC_i=\categZorCsmall \big]\E\big[ X_j \big]
    \\ &= 
    \E\big[ (W^X)^2 \big| W^\categZorC \geq \cutpoint \big]
    \E\big[ X^2 \big| \categZorC=\categZorCsmall \big]
    + \E\big[ (1-W^X)^2 \big| W^\categZorC \geq \cutpoint \big]
    \E\big[ X^2 \big]
    \\& \quad +
    2\E\big[ W^X(1-W^X) \big| W^\categZorC \geq \cutpoint \big]
    \E\big[ X \big| \categZorC=\categZorCsmall \big]\E\big[ X \big],
    \\
    \E\big[ \tilde{X}^2  \big| A_2 \big]
    &=
    \E\big[ (W^X)^2 X_i^2 + (1-W^X)^2 X_j^2 + 2 W^X (1-W^X) X_iX_j \big| \{\categZorC_j=\categZorCsmall, W^\categZorC < \cutpoint\} \big]
    \\ &= 
    \E\big[ (W^X)^2 \big| W^\categZorC < \cutpoint \big]\E\big[ X^2 \big]
    + \E\big[ (1-W^X)^2 \big| W^\categZorC < \cutpoint \big]\E\big[ X^2 \big| \categZorC=\categZorCsmall \big]
    \\& \quad +
    2\E\big[ W^X(1-W^X) \big| W^\categZorC < \cutpoint \big]
    \E\big[ X \big]\E\big[ X \big| \categZorC=\categZorCsmall \big].
\end{align*}
Then,
\begin{align}
    \E\big[\tilde{X}^2 \big| \tilde{\categZorC}=\categZorCsmall \big]
    &= \nonumber 
    \E\big[ \tilde{X}^2  \big| A_1 \big] \Pr\{ W^\categZorC\geq \cutpoint\}
    + \E\big[ \tilde{X}^2  \big| A_2 \big] \Pr\{ W^\categZorC < \cutpoint\}
    \\ &= \nonumber 
    \E\big[ (W^X)^2 \mathbf{I} \{ W^\categZorC\geq \cutpoint\} \big]
    \E\big[ X^2 \big| \categZorC=\categZorCsmall \big]
    + \E\big[ (1-W^X)^2 \mathbf{I} \{ W^\categZorC\geq \cutpoint\} \big]
    \E\big[ X^2 \big]
    \\& \nonumber \quad +
    2\E\big[ W^X(1-W^X) \mathbf{I} \{ W^\categZorC\geq \cutpoint\} \big]
    \E\big[ X \big| \categZorC=\categZorCsmall \big]\E\big[ X \big]
    \\& \nonumber \quad +
    \E\big[ (W^X)^2 \mathbf{I} \{ W^\categZorC< \cutpoint\} \big]\E\big[ X^2 \big]
    + \E\big[ (1-W^X)^2 \mathbf{I} \{ W^\categZorC<\cutpoint\}\big]\E\big[ X^2 \big| \categZorC=\categZorCsmall \big]
    \\& \nonumber \quad +
    2\E\big[ W^X(1-W^X)\mathbf{I} \{ W^\categZorC< \cutpoint\} \big]
    \E\big[ X \big]\E\big[ X \big| \categZorC=\categZorCsmall \big]
    \\ &= \nonumber 
    \E\big[ (W^X)^2 \mathbf{I} \{ W^\categZorC\geq \cutpoint\} +  (1-W^X)^2 \mathbf{I} \{ W^\categZorC< \cutpoint\} \big]
    \E\big[ X^2 \big| \categZorC=\categZorCsmall \big]
    \\& \nonumber \quad +
    \E\big[ (1-W^X)^2 \mathbf{I} \{ W^\categZorC\geq \cutpoint\} + (W^X)^2 \mathbf{I} \{ W^\categZorC< \cutpoint\}  \big]
    \E\big[ X^2 \big]
    \\& \label{eq:proof:condisec}
    \quad +
    2\E\big[ W^X(1-W^X)  \big]
    \E\big[ X \big| \categZorC=\categZorCsmall \big]\E\big[ X \big]. 
\end{align}

Note that $u\big(W^X, W^\categZorC,\cutpoint\big)=\E\big[ \big(1-W^X\big) \mathbf{I} \{ W^\categZorC\geq \cutpoint\} + W^X  \mathbf{I} \{ W^\categZorC < \cutpoint\} \big]$ by the definition in \eqref{eq:def:ufunction}.

Now, we introduce the new random variable $\tilde{W}$ where
\begin{align}    
    \tilde{W}&=\begin{cases}
    1-W^X &\text{ if } W^\categZorC \geq \cutpoint\\
    W^X &\text{ if } W^\categZorC < \cutpoint
    \end{cases}
    \label{eq:def:w:tilde}
    \\
    &= \nonumber (1-W^X) \mathbf{I} \{ W^\categZorC\geq \cutpoint\} + W^X \mathbf{I} \{ W^\categZorC< \cutpoint\},
\end{align}
which implies $\E[\tilde{W}]=u\big(W^X, W^\categZorC,\cutpoint\big)$ and $\E[1-\tilde{W}]=1-u\big(W^X, W^\categZorC,\cutpoint\big)=\E\big[ W^X \mathbf{I} \{ W^\categZorC\geq \cutpoint\} + \big(1-W^X\big) \mathbf{I} \{ W^\categZorC < \cutpoint\} \big]$.

If the assumption $\E\big[\big(W^X\big)^2\big] = \E\big[W^X \big]$, which is the exactly same condition of \eqref{eq:necessary:var} in Lemma~\ref{thm:app:var}, holds, then $\E\big[\tilde{W}^2\big] = \E\big[\tilde{W}\big]$ from
\begin{align*}
    \E\big[ \tilde{W}^2 \big]
    &=
    \E\big[ \big( \big(1-W^X\big)\mathbf{I}\big\{ W^\categZorC \geq \cutpoint \big\} + W^X\mathbf{I}\big\{ W^\categZorC < \cutpoint \big\} \big)^2 \big]
    \\ &=
    \E\big[ \big(1-W^X\big)^2\mathbf{I}\big\{ W^\categZorC \geq \cutpoint \big\} + \big(W^X)^2\mathbf{I}\big\{ W^\categZorC < \cutpoint \big\} + 0 \big]
    \\ &=
    \E\big[ \big(1-W^X\big)\mathbf{I}\big\{ W^\categZorC \geq \cutpoint \big\} + W^X\mathbf{I}\big\{ W^\categZorC < \cutpoint \big\} \big]
    \\ &=
    \E\big[ \tilde{W} \big].
\end{align*}
Therefore, $\Var\big[ \tilde{W}\big] = \E\big[ \tilde{W}^2 \big] - \E\big[ \tilde{W} \big]^2 = \E\big[ 1-\tilde{W} \big]\E\big[ \tilde{W} \big]$.

By using the random variable $\tilde{W}$ defined in \eqref{eq:def:w:tilde}, the conditioned second moment of \eqref{eq:proof:condisec} is simplified as
\begin{align*}
    \E\big[\tilde{X}^2 \big| \tilde{\categZorC}=\categZorCsmall \big]
    &= 
    \E\big[ (W^X)^2 \mathbf{I} \{ W^\categZorC\geq \cutpoint\} +  (1-W^X)^2 \mathbf{I} \{ W^\categZorC< \cutpoint\} \big]
    \E\big[ X^2 \big| \categZorC=\categZorCsmall \big]
    \\& \quad +
    \E\big[ (1-W^X)^2 \mathbf{I} \{ W^\categZorC\geq \cutpoint\} + (W^X)^2 \mathbf{I} \{ W^\categZorC< \cutpoint\}  \big]
    \E\big[ X^2 \big]
    \\& \quad +
    2\E\big[ W^X(1-W^X)  \big]
    \E\big[ X \big| \categZorC=\categZorCsmall \big]\E\big[ X \big]
    \\ &= 
    \E\big[ (1-\tilde{W})^2 \big]
    \E\big[ X^2 \big| \categZorC=\categZorCsmall \big]
     + \E\big[ \tilde{W}^2  \big]
    \E\big[ X^2 \big]
    \\& \quad +
    2\E\big[ \tilde{W}(1-\tilde{W})  \big]
    \E\big[ X \big| \categZorC=\categZorCsmall \big]\E\big[ X \big].
\end{align*}
Similarly, from \eqref{eq:thm:con:mean} in Theorem~\ref{thm:app:con:mean}, the conditional mean is also simplified as
\begin{align*}
    \E\big[\tilde{X} \big| \tilde{\categZorC}=\categZorCsmall \big]^2
    &=  
    (1-u\big(W^X, W^\categZorC,\cutpoint\big))^2\E\big[X \big| \categZorC=\categZorCsmall \big]^2
    + u\big(W^X, W^\categZorC,\cutpoint\big)^2 \E\big[X \big]^2
    \\ & \quad 
    +2(1-u\big(W^X, W^\categZorC,\cutpoint\big))u\big(W^X, W^\categZorC,\cutpoint\big) \E\big[X \big| \categZorC=\categZorCsmall \big] \E\big[X \big]
    \\ &= 
    \E\big[ 1-\tilde{W} \big]^2\E\big[X \big| \categZorC=\categZorCsmall \big]^2
    + \E\big[ \tilde{W} \big]^2 \E\big[X \big]^2
    \\& \quad +
    2\E\big[ 1-\tilde{W} \big] \E\big[ \tilde{W} \big]
    \E\big[ X \big| \categZorC=\categZorCsmall \big]\E\big[ X \big].
\end{align*}

As a result, the conditional variance is determined as
\begin{align*}
    \Var\big[ \tilde{X}\big| \tilde{\categZorC}=\categZorCsmall \big]
    &= 
    \E\big[\tilde{X}^2 \big| \tilde{\categZorC}=\categZorCsmall \big]
    - \E\big[\tilde{X} \big| \tilde{\categZorC}=\categZorCsmall \big]^2
    \\ &= 
    \E\big[ (1-\tilde{W})^2 \big]
    \E\big[ X^2 \big| \categZorC=\categZorCsmall \big]
     + \E\big[ \tilde{W}^2  \big]
    \E\big[ X^2 \big]
    + 2\E\big[ \tilde{W}(1-\tilde{W})  \big]
    \E\big[ X \big| \categZorC=\categZorCsmall \big]\E\big[ X \big]
    \\& \quad 
    - \E\big[ 1-\tilde{W} \big]^2\E\big[X \big| \categZorC=\categZorCsmall \big]^2
    - \E\big[ \tilde{W} \big]^2 \E\big[X \big]^2
    - 2\E\big[ 1-\tilde{W} \big] \E\big[ \tilde{W} \big]
    \E\big[ X \big| \categZorC=\categZorCsmall \big]\E\big[ X \big]
    \\ &= 
    \E\big[ (1-\tilde{W})^2 \big] \Var\big[ X \big| \categZorC=\categZorCsmall \big]
     + \E\big[ \tilde{W}^2  \big] \Var\big[ X \big]
    \\& \quad 
    + \big( \E\big[ (1-\tilde{W})^2 \big] - \E\big[ 1-\tilde{W} \big]^2 \big)\E\big[X \big| \categZorC=\categZorCsmall \big]^2
    + \big( \E\big[ \tilde{W}^2 \big] - \E\big[ \tilde{W} \big]^2 \big) \E\big[X \big]^2
    \\& \quad
    - 2\big(\E\big[ \tilde{W}^2 \big] - \E\big[ \tilde{W} \big]^2 \big)
    \E\big[ X \big| \categZorC=\categZorCsmall \big]\E\big[ X \big]
    \\ &= 
    \E\big[ (1-\tilde{W})^2 \big] \Var\big[ X \big| \categZorC=\categZorCsmall \big]
     + \E\big[ \tilde{W}^2  \big] \Var\big[ X \big]
    + \Var\big[ \tilde{W}\big]
    \big(\E\big[ X \big| \categZorC=\categZorCsmall \big]-\E\big[ X \big]\big)^2
    \\ &= 
    \E\big[ 1-\tilde{W} \big] \Var\big[ X \big| \categZorC=\categZorCsmall \big]
     + \E\big[ \tilde{W} \big] \Var\big[ X \big]
    + \E\big[ 1-\tilde{W} \big]\E\big[ \tilde{W} \big]
    \big(\E\big[ X \big| \categZorC=\categZorCsmall \big]-\E\big[ X \big]\big)^2
    \\ &= 
    \big(1- u\big(W^X, W^\categZorC,\cutpoint\big)\big) \Var\big[ X \big| \categZorC=\categZorCsmall \big]
     + u\big(W^X, W^\categZorC,\cutpoint\big) \Var\big[ X \big]
    \\& \quad 
    + \big(1- u\big(W^X, W^\categZorC,\cutpoint\big)\big) u\big(W^X, W^\categZorC,\cutpoint\big)
    \big(\E\big[ X \big| \categZorC=\categZorCsmall \big]-\E\big[ X \big]\big)^2
    \\ &= 
    \Var\big[ X \big| \categZorC=\categZorCsmall \big]
     + u\big(W^X, W^\categZorC,\cutpoint\big) \big( \Var\big[ X \big] - \Var\big[ X \big| \categZorC=\categZorCsmall \big] \big)
    \\& \quad 
    + \big(1- u\big(W^X, W^\categZorC,\cutpoint\big)\big) u\big(W^X, W^\categZorC,\cutpoint\big)
    \big(\E\big[ X \big| \categZorC=\categZorCsmall \big]-\E\big[ X \big]\big)^2,
\end{align*}
which is equal to
\begin{align*}
    \nonumber
    \big|\Var\big[ \tilde{X}\big| \tilde{\categZorC}=\categZorCsmall \big]
    -  \Var\big[ X \big| \categZorC=\categZorCsmall \big]\big|
     & \leq
    \big| u\big(W^X, W^\categZorC,\cutpoint\big)\big|
    \cdot \big| \Var\big[ X \big| \categZorC=\categZorCsmall \big] - \Var\big[X\big] \big|
    \\
    \nonumber
    &\qquad +\big| u\big(W^X, W^\categZorC,\cutpoint\big)(1- u\big(W^X, W^\categZorC,\cutpoint\big)) \big|  
    \cdot \big(\E\big[ X \big| \categZorC=\categZorCsmall \big]-\E\big[ X \big]\big)^2.
\end{align*}
by the triangular inequality.
\end{proof}

\clearpage
\begin{lemma}
    \label{thm:app:nonnegative:u}
   Under the standard mixup scheme, $u(W,\cutpoint)\in [0,1]$ holds for any $\cutpoint\in\mathbb{R}$ if $\E\big[W^2\big] = \E\big[W \big]$.
\end{lemma}
\begin{proof}
    From the definition of function $u\big(W,\cutpoint\big)$ in \eqref{eq:def:ufunction:abb},
    \begin{align*}
        u(W,\cutpoint)
        &=
        \E\big[ \big(1-W\big) \mathbf{I} \{ W\geq \cutpoint\} + W  \mathbf{I} \{ W < \cutpoint\} \big]
        \\ &=
        \E\big[ \big(1-W\big) \mathbf{I} \{ W\geq \cutpoint\} + W  \big(1 -\mathbf{I} \{ W \geq \cutpoint\} \big) \big]
        \\ &=
        \E\big[\big(W\big)^2\big]
        + \E\big[ \big(1-2W\big) \mathbf{I} \{ W\geq \cutpoint\} \big],
    \end{align*}
    where the last equality holds if  $\E\big[\big(W\big)^2\big] = \E\big[W \big]$.
    Then, it is to trivial to show the non-negativity of $u(W,\cutpoint)$ for all $c\in\mathbb{R}$ as follows:
    \begin{align*}
        u(W,\cutpoint)
        &=
        \E\big[\big(W\big)^2 \mathbf{I} \{ W < \cutpoint\}\big]
        + \E\big[ \big(1-2W+\big(W\big)^2\big) \mathbf{I} \{ W\geq \cutpoint\} \big]
        \\ &=
        \E\big[\big(W\big)^2 \mathbf{I} \{ W < \cutpoint\}\big]
        + \E\big[ \big(1-W\big)^2 \mathbf{I} \{ W\geq \cutpoint\} \big]
        \\ &\geq 0.
    \end{align*}
    In a similar manner we can show the upper bound as
    \begin{align*}
        u(W, \cutpoint)
        &=
        \E\big[ \big(1-W\big) \mathbf{I} \{ W\geq \cutpoint\} + W  \mathbf{I} \{ W < \cutpoint\} \big]
        \\ &=
        \E\big[ \big(1-W\big) \big(1 -\mathbf{I} \{ W < \cutpoint\} \big) + W  \mathbf{I} \{ W < \cutpoint\} \big]
        \\ &=
        1-\E\big[\big(W\big)^2\big]
        + \E\big[ \big(2W -1\big) \mathbf{I} \{ W < \cutpoint\} \big]
        \\ &=
        1-\E\big[\big(W\big)^2 \mathbf{I} \{ W \geq \cutpoint\}\big]
        - \E\big[ \big(1-W\big)^2 \mathbf{I} \{ W < \cutpoint\} \big]
        \\ &\leq 1.
    \end{align*}
\end{proof}

\begin{lemma}[Optimal Cut Point $\cutpoint$]
    \label{thm:app:condimean:cutpoint}
    Under the standard mixup scheme with $\E\big[W^2\big] = \E\big[W \big]$, the optimal cut point $\cutpoint$  is $0.5$. That is
    \begin{equation}
    \nonumber
    0.5 = \argmin_{\cutpoint \in \mathbb{R}} \big| u\big(W,\cutpoint\big) \big| .
    \end{equation}
\end{lemma}
\begin{proof}
    From Lemma~\ref{thm:app:nonnegative:u}, $u(W,\cutpoint)\in [0,1]$ holds for any $\cutpoint\in\mathbb{R}$. Then,
    \begin{align*}
        \big| u\big(W,\cutpoint\big) \big|
        & \geq
         u\big(W,\cutpoint\big) 
        \\ &=
        \E\big[ \big(1-W\big) \mathbf{I} \{ W\geq \cutpoint\} + W  \mathbf{I} \{ W < \cutpoint\} \big]
        \\ & \geq
        \E\big[ \min \big\{ 1-W, W\big\} \mathbf{I} \{ W\geq \cutpoint\} 
        + \min \big\{ 1-W, W\big\} \mathbf{I} \{ W < \cutpoint\} \big]
        \\ &=
        \E\big[ \min \big\{ 1-W, W\big\} \big]
        \\ &=
        \E\big[ \big(1-W\big) \mathbf{I} \{ W\geq 0.5\} + W  \mathbf{I} \{ W < 0.5\} \big]
        \\ &= u(W, 0.5)
    \end{align*}
    for any $W$. Therefore, with the condition $u(W, 0.5)\geq 0$, the optimal cut point $\cutpoint$ for the categorical variable is $0.5$. That is
    \begin{equation*}
        0.5 = \argmin_{\cutpoint \in (0,1)} \big|u\big(W,\cutpoint\big)\big|.
    \end{equation*}
\end{proof}

\clearpage
\subsection{Proofs for Sec.~\ref{sec:mixup:distn}}

\begin{corollary}[Variance-Reduction mixup]
    \label{thm:app:bounded:support:mixup}
    For any synthetic variable $\tilde X$ generated by the mixup from a continuous $X$ in (\ref{eq:mixup:x}), let the support of mixup weight variable $W^X$ be bounded in $[0,1]$. Then 
    \begin{equation}
        \Var\big[\tilde{X}\big] \leq \Var\big[X\big],\nonumber %
    \end{equation}
    where the equality holds when $\Pr\big\{W^X \in \{0, 1\} \big\}=1$.
\end{corollary}
\begin{proof}
    Note that $w^2 - w \leq 0$ holds for any $w\in[0,1]$, which implies $\E\big[\big(W^X\big)^2 - W^X\big]\leq 0$, where the equality condition is $\Pr\big\{W^X \in \{0, 1\} \big\}=1$.
    From Lemma~\ref{thm:app:var}, we have $\Var\big[\tilde{X}\big] \leq \Var\big[X\big]$.
\end{proof}

\begin{example_for_appendix}
Let the mixup weights are generated from the Gaussian distribution $\operatorname{N}(\mu, \sigma^2)$ where $\sigma = \sqrt{\mu-\mu^2}$ for some $\mu\in [0, 1]$, i.e., $W^X=W^Y \sim  \operatorname{N}(\mu, \mu-\mu^2)$. Then, under the standard mixup scheme,
 we have, for any pair $(X,Y)$,  $\Var\big[\tilde{X}\big] = \Var\big[X\big] $, $\Var\big[\tilde{Y}\big] = \Var\big[Y\big]$, and $\Cov\big[\tilde{X},\tilde{Y}\big] = \Cov\big[X,Y\big] $.
\end{example_for_appendix}
\begin{proof}
    By the definition of distribution $\operatorname{N}(\mu, \sigma^2)$ where $\sigma = \sqrt{\mu-\mu^2}$ for some $\mu\in [0, 1]$, $\mu=\E\big[\big(W^X\big)^2\big] =  \E\big[W^X \big]=\E\big[\big(W^Y\big)^2\big] = \E\big[W^Y \big] = \E\big[W^XW^Y\big]$. Then, the conditions of \eqref{eq:necessary:var} in Lemma~\ref{thm:app:var} and \eqref{eq:necessary:cov} in Theorem~\ref{thm:app:cov} hold, which implies that $\Var\big[\tilde{X}\big] = \Var\big[X\big] $, $\Var\big[\tilde{Y}\big] = \Var\big[Y\big]$, and $\Cov\big[\tilde{X},\tilde{Y}\big] = \Cov\big[X,Y\big] $.
\end{proof}

\begin{theorem}
\label{thm:app:epbeta:var}
For given $\epsilon_0, \epsilon_1 \in [0,\infty)$, consider an arbitrary synthetic pair $(\tilde{X}, \tilde{Y})$ generated from $({X}, {Y})$ using the standard mixup scheme with  $W \sim \operatorname{EpBeta}(\alpha, \beta; \epsilon_0, \epsilon_1)$ 
, such that $\alpha, \beta \in (0, \infty)$ satisfy
\begin{equation}
    \nonumber
    (1+\epsilon_1 -\epsilon_0 (\beta/\alpha)) \cdot (1+\epsilon_0-\epsilon_1 (\alpha/\beta)) \cdot (1+\alpha+\beta)
    =
    (1+\epsilon_0+\epsilon_1)^2.
\end{equation}
Then we have  $\Var\big[\tilde{X}\big] = \Var\big[X\big] $ and $\Cov\big[\tilde{X},\tilde{Y}\big] = \Cov\big[X,Y\big] $.
\end{theorem}
\begin{proof}
    Let $W$ follow the $\operatorname{Beta}(\alpha, \beta)$ distribution, which implies $\E[W]=\frac{\alpha}{\alpha+\beta}$ and $\Var[W]=\frac{\alpha\beta}{(\alpha+\beta)^2(\alpha+\beta+1)}$. Then,
    \begin{align*}
        \E\big[(1+\epsilon_0 +\epsilon_1)W-\epsilon_0\big]
        &= \frac{(1+\epsilon_0 +\epsilon_1)\alpha}{\alpha+\beta} - \epsilon_0
        \\
        &= \frac{(1+\epsilon_1)\alpha -\epsilon_0\beta}{\alpha+\beta},
        \\
        \E\big[(1+\epsilon_0 +\epsilon_1)W-\epsilon_0\big]^2
        &= \Var\big[(1+\epsilon_0 +\epsilon_1)W-\epsilon_0\big] + \big(\E\big[(1+\epsilon_0 +\epsilon_1)W-\epsilon_0\big]\big)^2
        \\
        &=\frac{(1+\epsilon_0+\epsilon_1)^2 \alpha\beta}{(\alpha+\beta)^2(\alpha+\beta+1)}
        + \left( \frac{(1+\epsilon_1)\alpha -\epsilon_0 \beta}{\alpha+\beta} \right)^2.
    \end{align*}

By Theorem~\ref{thm:app:cov}, $\Var\big[\tilde{D}\big] = \Var\big[D\big]$ if and only if
\begin{align}
    \label{eq:proof:epbeta}
    \E\big[(1+\epsilon_0 +\epsilon_1)W-\epsilon_0\big]=\E\big[(1+\epsilon_0 +\epsilon_1)W-\epsilon_0\big]^2.
\end{align}

To find the $\alpha, \beta  \in (0, \infty)$ that satisfy \eqref{eq:proof:epbeta}, 
    \begin{align*}
    ((1+\epsilon_1)\alpha -\epsilon_0 \beta)(\alpha+\beta)
    &=
    \frac{(1+\epsilon_0+\epsilon_1)^2 \alpha\beta}{(\alpha+\beta+1)}
    + \big( (1+\epsilon_1)\alpha -\epsilon_0 \beta \big)^2,
    \end{align*}
which is equal to
    \begin{align*}
    (1+\epsilon_1 -\epsilon_0 (\beta/\alpha)) \cdot (1+\epsilon_0-\epsilon_1 (\alpha/\beta)) \cdot (1+\alpha+\beta)
    &=
    (1+\epsilon_0+\epsilon_1)^2.
    \end{align*}

    Note that the above equation can be rewritten as
    \begin{align*}
    \alpha+\beta
    &=
    \frac{(1+\epsilon_0+\epsilon_1)^2}{(1+\epsilon_1 -\epsilon_0 (\beta/\alpha)) \cdot (1+\epsilon_0-\epsilon_1 (\alpha/\beta))}
    -1 ,
    \end{align*}
    which implies that we can find $(\alpha, \beta)$ that satisfies the condition in \eqref{eq:proof:epbeta} for given $\beta/\alpha, \epsilon_0$, and $\epsilon_1$.
\end{proof}

\begin{theorem}
    \label{thm:app:epbeta:condimean}
   Consider an arbitrary synthetic triple $(\tilde{X}, \tilde{Y}, \tilde{\categZorC})$ generated from $({X}, {Y}, \categZorC)$ using the standard mixup scheme with $W \sim \operatorname{EpBeta}(\alpha, \beta; \epsilon_0, \epsilon_1)$ for given $\epsilon_0, \epsilon_1 \in [0,\infty)$ and $\cutpoint=0.5$. Now suppose that, for a given  $\delta \in [0,1]$, $(\alpha, \beta)$ satisfies (\ref{eq:epbeta:constraint}) and the following 

    \begin{equation}
    \nonumber 
    \frac{ 1+\epsilon_0 -\epsilon_1 \alpha /\beta}{1+\alpha/\beta}
    + \frac{2(1+\epsilon_0+\epsilon_1)}{1+\beta/\alpha} \frac{B(\tilde{\epsilon};\alpha+1, \beta)}{B(1;\alpha+1, \beta)}
    - (1+2\epsilon_0) \frac{B\left(\tilde{\epsilon};\alpha, \beta\right)}{B(1;\alpha, \beta)}
    \leq
    \delta,
    \end{equation}
    where $B(x;\alpha, \beta) = \int_{0}^{b} t^{\alpha-1} (1-t)^{\beta-1} \,dt$ is the incomplete beta function and $\tilde{\epsilon}=\frac{0.5+\epsilon_0}{1+\epsilon_0+\epsilon_1}$.
    
 Then, the gap of conditional (on categorical $\categZorC$) mean and variance are bounded as follows:
 \begin{equation*}
     \big|
    \E\big[\tilde{X} \big| \tilde{\categZorC}=\categZorCsmall \big]
    - \E\big[X \big| \categZorC=\categZorCsmall \big]
    \big|
    =
    \delta
    \cdot \Pr\{ \categZorC \neq \categZorCsmall\}
    \cdot
    \big|
    \E\big[X \big| \categZorC=\categZorCsmall \big]
    - \E\big[X \big| \categZorC \neq \categZorCsmall \big]
    \big| 
 \end{equation*} and 
\begin{align*}
    \big|\Var\big[ \tilde{X}\big| \tilde{\categZorC}=\categZorCsmall \big]
    -  \Var\big[ X \big| \categZorC=\categZorCsmall \big]\big|
    & \leq 
    \delta
    \cdot \big| \Var\big[ X \big| \categZorC=\categZorCsmall \big] - \Var\big[X\big] \big|
    +\delta(1-\delta)
    \cdot \big(\E\big[ X \big| \categZorC=\categZorCsmall \big]-\E\big[ X \big]\big)^2.
\end{align*} 
\end{theorem}
\begin{proof}
    Note that the conditional mean and variance gaps are bounded by the functions of $u\big(W^X, W^\categZorC,\cutpoint\big)$ from Corollary~\ref{thm:app:con:mean:cor} and Theorem~\ref{thm:app:con:var}, respectively.
    To determine it, define the random variable $W\sim \operatorname{Beta}(\alpha,\beta)$, which implies $W^X=W^\categZorC=(1+\epsilon_0+\epsilon_1)W-\epsilon_0$. Then, with \eqref{eq:epbeta:conditional:constraint},
\begin{align*}
    u\big(W^X, W^\categZorC,\cutpoint\big) &= u(W^X, 0.5) 
    \\&=
    \E\big[ \big(1-W^X\big) \mathbf{I} \{ W^X\geq 0.5\} + W^X  \mathbf{I} \{ W^X < 0.5\} \big]
    \\&=
    \E\big[ \big(1-W^X\big) (1-\mathbf{I} \{ W^X < 0.5\}) + W^X  \mathbf{I} \{ W^X < 0.5\} \big]
    \\&=
    \E\big[ 1-W^X \big] +
    \E\big[ \big(2W^X-1\big) \mathbf{I} \{ W^X < 0.5\} \big]
    \\&=
    \frac{-\epsilon_1 \alpha + (1+\epsilon_0)\beta}{\alpha+\beta}
    +
    \E\big[ \big(2(1+\epsilon_0+\epsilon_1) W -1- 2\epsilon_0\big)  \mathbf{I} \{ (1+\epsilon_0+\epsilon_1) W - \epsilon_0 < 0.5\} \big]
    \\&=
    \frac{-\epsilon_1 \alpha + (1+\epsilon_0)\beta}{\alpha+\beta}
    +
    \E\bigg[ \bigg(2(1+\epsilon_0+\epsilon_1) W -1 -2\epsilon_0\bigg) \mathbf{I} \bigg\{ W  < \frac{0.5+\epsilon_0}{1+\epsilon_0+\epsilon_1} \bigg\} \bigg]
    \\&=
    \frac{ 1+\epsilon_0 -\epsilon_1 \alpha /\beta}{1+\alpha/\beta}
    + 2(1+\epsilon_0+\epsilon_1) \frac{B(\tilde{\epsilon};\alpha+1, \beta)}{B(1;\alpha, \beta)}
    - (1+2\epsilon_0) \frac{B\left(\tilde{\epsilon};\alpha, \beta\right)}{B(1;\alpha, \beta)}
    \\&=
    \frac{ 1+\epsilon_0 -\epsilon_1 \alpha /\beta}{1+\alpha/\beta}
    + 2(1+\epsilon_0+\epsilon_1) \frac{\alpha}{\alpha+\beta} \frac{B(\tilde{\epsilon};\alpha+1, \beta)}{B(1;\alpha+1, \beta)}
    - (1+2\epsilon_0) \frac{B\left(\tilde{\epsilon};\alpha, \beta\right)}{B(1;\alpha, \beta)}
    \\&=
    \frac{ 1+\epsilon_0 -\epsilon_1 \alpha /\beta}{1+\alpha/\beta}
    + \frac{2(1+\epsilon_0+\epsilon_1)}{1+\beta/\alpha} \frac{B(\tilde{\epsilon};\alpha+1, \beta)}{B(1;\alpha+1, \beta)}
    - (1+2\epsilon_0) \frac{B\left(\tilde{\epsilon};\alpha, \beta\right)}{B(1;\alpha, \beta)}
    \\& \leq \delta,
\end{align*}
where $B(x;\alpha, \beta) = \int_{0}^{b} t^{\alpha-1} (1-t)^{\beta-1} \,dt$ is the incomplete beta function, and $\tilde{\epsilon}=\frac{0.5+\epsilon_0}{1+\epsilon_0+\epsilon_1}$. 

Note that $u(W^X,\cutpoint)$ is bounded in $[0,1]$ under the condition of \eqref{eq:epbeta:constraint} by Lemma~\ref{thm:app:nonnegative:u}, which implies $\big| u\big(W^X, W^\categZorC,\cutpoint\big)\big|\leq \delta$.
From Corollary~\ref{thm:app:con:mean:cor} and Theorem~\ref{thm:app:con:var}, respectively,
    \begin{align}
    \big|
    \E\big[\tilde{X} \big| \tilde{\categZorC}=\categZorCsmall \big]
    - \E\big[X \big| \categZorC=\categZorCsmall \big]
    \big|
    &= \nonumber
    \big| u\big(W^X, W^\categZorC,\cutpoint\big) \big|
    \cdot \Pr\{ \categZorC \neq \categZorCsmall\}
    \cdot
    \big|
    \E\big[X \big| \categZorC=\categZorCsmall \big]
    - \E\big[X \big| \categZorC \neq \categZorCsmall \big]
    \big|
    \\ \nonumber
    &\leq
    \delta
    \cdot \Pr\{ \categZorC \neq \categZorCsmall\}
    \cdot
    \big|
    \E\big[X \big| \categZorC=\categZorCsmall \big]
    - \E\big[X \big| \categZorC \neq \categZorCsmall \big]
    \big|
    ,
    \\
    \big|\Var\big[ \tilde{X}\big| \tilde{\categZorC}=\categZorCsmall \big]
    -  \Var\big[ X \big| \categZorC=\categZorCsmall \big]\big|
    &  \nonumber
    \leq
    \big| u\big(W^X, W^\categZorC,\cutpoint\big) \big|
    \cdot \big| \Var\big[ X \big| \categZorC=\categZorCsmall \big] - \Var\big[X\big] \big|
    \\&  \nonumber
    \quad
    +\big| u\big(W^X, W^\categZorC,\cutpoint\big) \big(1- u\big(W^X, W^\categZorC,\cutpoint\big)\big) \big|
    \cdot \big(\E\big[ X \big| \categZorC=\categZorCsmall \big]-\E\big[ X \big]\big)^2
    \\&  \label{eq:proof:page29}
    \leq \delta \cdot \big| \Var\big[ X \big| \categZorC=\categZorCsmall \big] - \Var\big[X\big] \big|
    + \delta( 1- \delta )
    \cdot \big(\E\big[ X \big| \categZorC=\categZorCsmall \big]-\E\big[ X \big]\big)^2.
    \end{align}
    The last inequality in \eqref{eq:proof:page29} comes from
    \begin{align*}
    \big| u\big(W^X, W^\categZorC,\cutpoint\big) \big(1- u\big(W^X, W^\categZorC,\cutpoint\big)\big) \big|
    & =
    u\big(W^X, W^\categZorC,\cutpoint\big)  - u\big(W^X, W^\categZorC,\cutpoint\big)^2
    \leq
    u\big(W^X, W^\categZorC,\cutpoint\big)
    \leq \delta.
    \end{align*}

 Therefore, the gap of conditional mean and variance are bounded as
 \begin{equation*}
     \big|
    \E\big[\tilde{X} \big| \tilde{\categZorC}=\categZorCsmall \big]
    - \E\big[X \big| \categZorC=\categZorCsmall \big]
    \big|
    =
    \delta
    \cdot \Pr\{ \categZorC \neq \categZorCsmall\}
    \cdot
    \big|
    \E\big[X \big| \categZorC=\categZorCsmall \big]
    - \E\big[X \big| \categZorC \neq \categZorCsmall \big]
    \big|,
 \end{equation*} and 
\begin{align*}
    \big|\Var\big[ \tilde{X}\big| \tilde{\categZorC}=\categZorCsmall \big]
    -  \Var\big[ X \big| \categZorC=\categZorCsmall \big]\big|
    & \leq 
    \delta
    \cdot \big| \Var\big[ X \big| \categZorC=\categZorCsmall \big] - \Var\big[X\big] \big|
    +\delta(1-\delta)
    \cdot \big(\E\big[ X \big| \categZorC=\categZorCsmall \big]-\E\big[ X \big]\big)^2.
\end{align*}

Moreover, note that $\Pr\{ \categZorC \neq \categZorCsmall\}$, $\big| \E\big[X \big| \categZorC=\categZorCsmall \big] - \E\big[X \big| \categZorC \neq \categZorCsmall \big] \big|$, $\big| \Var\big[ X \big| \categZorC=\categZorCsmall \big] - \Var\big[X\big] \big|$, and $\big(\E\big[ X \big| \categZorC=\categZorCsmall \big]-\E\big[ X \big]\big)$ are constants given by the original distribution $\mathfrak{D}$. As result, when $\delta$ goes to $0$,
    \begin{align*}
    \big|  \E\big[\tilde{X} \big| \tilde{\categZorC}=\categZorCsmall \big] - \E\big[X \big| \categZorC=\categZorCsmall \big] \big|
    &\leq
    \delta \cdot \Pr\{ \categZorC \neq \categZorCsmall\}
    \cdot \big| \E\big[X \big| \categZorC=\categZorCsmall \big] - \E\big[X \big| \categZorC \neq \categZorCsmall \big] \big|
    \\ &\to 0,
    \\
    \big|\Var\big[ \tilde{X}\big| \tilde{\categZorC}=\categZorCsmall \big]
    -  \Var\big[ X \big| \categZorC=\categZorCsmall \big]\big|
    & \leq
    \delta \cdot \big| \Var\big[ X \big| \categZorC=\categZorCsmall \big] - \Var\big[X\big] \big|
    + \delta( 1- \delta )
    \cdot \big(\E\big[ X \big| \categZorC=\categZorCsmall \big]-\E\big[ X \big]\big)^2
    \\&\to 0,
    \end{align*}
    which are equal to
    \begin{align*}
        \E\big[\tilde{X} \big| \tilde{\categZorC}=\categZorCsmall \big] &\to \E\big[X \big| \categZorC=\categZorCsmall \big],
        \\
        \Var\big[ \tilde{X}\big| \tilde{\categZorC}=\categZorCsmall \big] &\to  \Var\big[ X \big| \categZorC=\categZorCsmall \big],
    \end{align*}
    as $\delta$ goes to $0$.
\end{proof}

\clearpage

\section{EXPERIMENT DETAILS}
\label{appendix:experiment}

\subsection{Data Descriptions and Synthesis Details}
\label{appendix:exp:datasyn}

In this experiment, we select 6 popular datasets used in \cite{gorishniy2021revisiting, kotelnikov2023tabddpm}. For convenience purposes, the instances with missing values are removed. The preprocessed datasets are summarized in Table~\ref{table:data}.

\begin{table}[h]
\caption{Data description.}
\label{table:data}
\centering
\begin{tabular}{c||cccccc}
\hline
Name               & \# Instance & \# Num & \# Cat & Task           & License & Source                                                             \\ \hline 
Abalone            & 4177   & 8 & 1      & Regress     & CC4.0   & \begin{tabular}[c]{@{}c@{}}UCI ML\\ \citep{misc_abalone_1}\end{tabular}  \\ 
CA Housing & 20433  & 9 & 1      & Regress     & CC0     & \begin{tabular}[c]{@{}c@{}}Kaggle\\ \citep{geron2022hands}\end{tabular}    \\ 
House 16H          & 22784  & 17 & 0      & Regress     & Public  & OpenML                                                             \\ 
Adult              & 48842  & 6 & 9      & Classify & CC4.0   & \begin{tabular}[c]{@{}c@{}}UCI ML\\ \citep{misc_adult_2}\end{tabular}    \\ 
Diabetes           & 768    & 8 & 1      & Classify & Public  & OpenML                                                             \\ 
Wilt               & 4839   & 5 & 1      & Classify & Public  & \begin{tabular}[c]{@{}c@{}}OpenML\\ \citep{johnson2013hybrid}\end{tabular} \\ \hline
\end{tabular}
\end{table}

We synthesize data using the mixup method with various weight distributions, such as $\operatorname{EpBeta}$, $\operatorname{Beta}$, and $\operatorname{Unif}$, and compare them against four baseline methods implemented through an open-source code \citep{qian2023synthcity}; TVAE, CTGAN \citep{xu2019modeling}, TabDDPM \citep{kotelnikov2023tabddpm}, and GReaT \citep{borisov2022language}. Each method is used with its default settings. For example, while TabDDPM allows for the selection of the target variable as an input, we do not utilize this option.

We apply the $\operatorname{EpBeta}$ distribution with four different $\delta$ values: $\delta \in {0.001, 0.005, 0.01, 0.05}$, as well as the $\operatorname{Beta}(0.1, 0.1)$ and $\operatorname{Unif}(0, 1)$ distributions as Mixup weight distributions, resulting in 10 synthetic datasets for each original dataset.

Using two NVIDIA GeForce RTX 3090 GPUs, we report the model training and generation times for producing a single synthetic dataset with the same number of instances as the original. These results are shown in Table~\ref{table:train:time}. Note that for the `House 16H' and `Adult' datasets, the GReaT method is excluded because it fails to converge within the default number of epochs.

\begin{table}[ht]
\caption{Training and generating time (Seconds).}
\label{table:train:time}
\centering
\begin{tabular}{c||ccccc}
\hline
Name       & Mixup & TVAE & CTGAN & TabDDPM  & GReaT \\ \hline
Abalone    & 0.004     & 207  & 213   & 73   & 6001  \\ 
CA Housing & 0.009     & 827  & 702   & 362  & 28852   \\ 
House 16H  & 0.016     & 2364 & 2713  & 239  & -   \\ 
Adult      & 0.046     & 4803 & 4533  & 1042 & -   \\ 
Diabetes   & 0.002     & 39   & 131   & 18   & 1277 \\ 
Wilt       & 0.004     & 261  & 427   & 90   & 6843 \\ \hline
\end{tabular}
\end{table}

\subsection{Relative Bias of Synthetic Data}
\label{appendix:exp:bias}

We compare the relative bias of covariance and expectation of continuous variables from each synthetic dataset, calculated as $\frac{\Cov[\Tilde{X},\Tilde{Y}]-\Cov[X,Y]}{\Cov[X,Y]}$ for covariance and $\frac{\E[\Tilde{X}]-\E[X]}{\E[X]}$ for expectation. In the figures, negative bias is shown in blue, positive bias in red, and grey indicates bias close to zero. Due to space constraints, we present results for only the first synthesized dataset ($m = n$) and the combined results from five subsequently synthesized datasets, equivalent to generating synthetic data with five times the number of original instances ($m = 5n$). Although some small differences in expectation and (co)variance may appear in a single dataset synthesized using the $\operatorname{EpBeta}$ distribution, these differences diminish as the number of synthesized instances increases across all datasets. Additionally, we abbreviate the annotation for `House 16H' due to the large number of variables.

\clearpage
\begin{figure}[h]
    \centering
    \includegraphics[width=\linewidth]{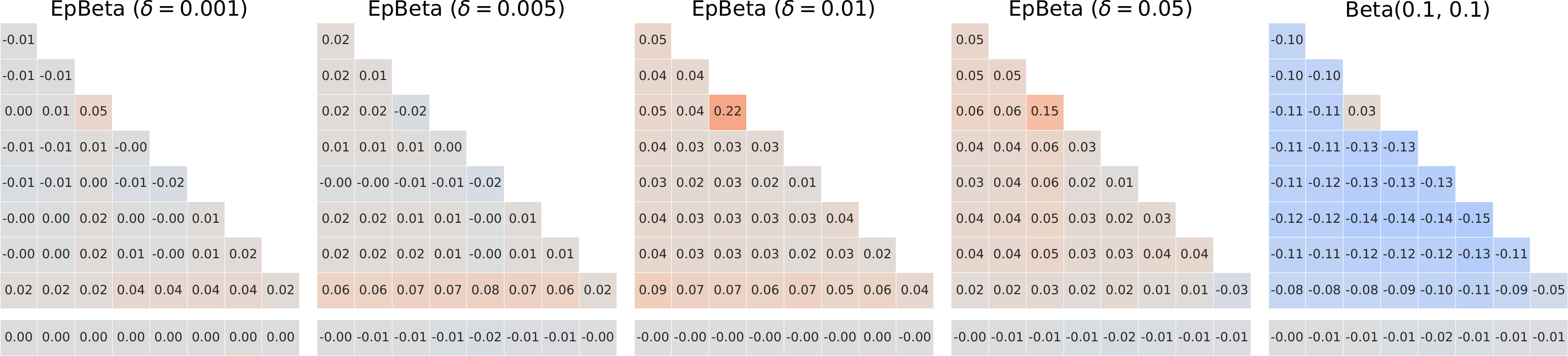}
    \includegraphics[width=\linewidth]{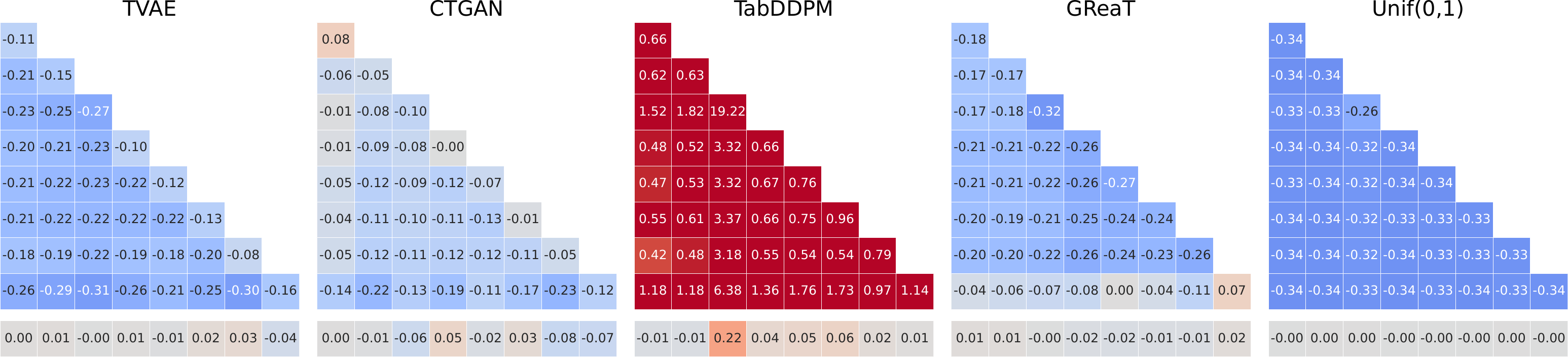}
    \caption{The relative bias of (co)variance (triangle) and expectation (bar) for `Abalone' with $m=n$.}
\end{figure}

\begin{figure}[h]
    \centering
    \includegraphics[width=\linewidth]{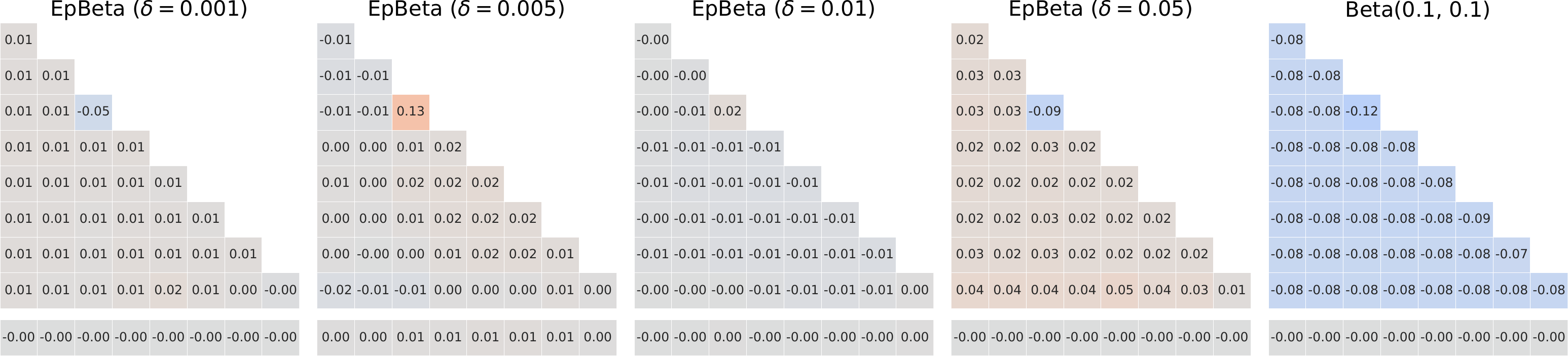}
    \includegraphics[width=\linewidth]{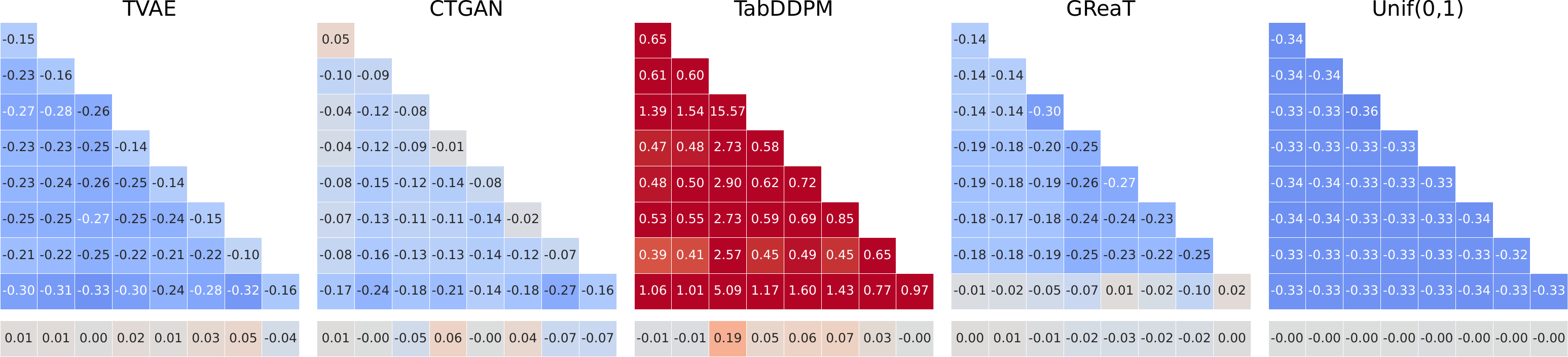}
    \caption{The relative bias of (co)variance (triangle) and expectation (bar) for `Abalone' with $m=5n$.}
    \vspace{-12pt}
\end{figure}

\clearpage
\begin{figure}[h]
    \centering
    \includegraphics[width=\linewidth]{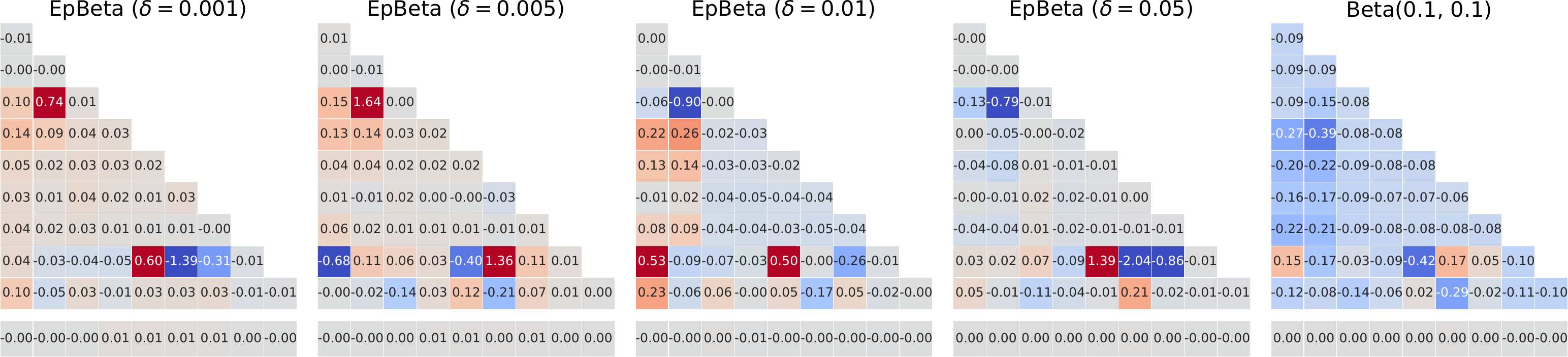}
    \includegraphics[width=\linewidth]{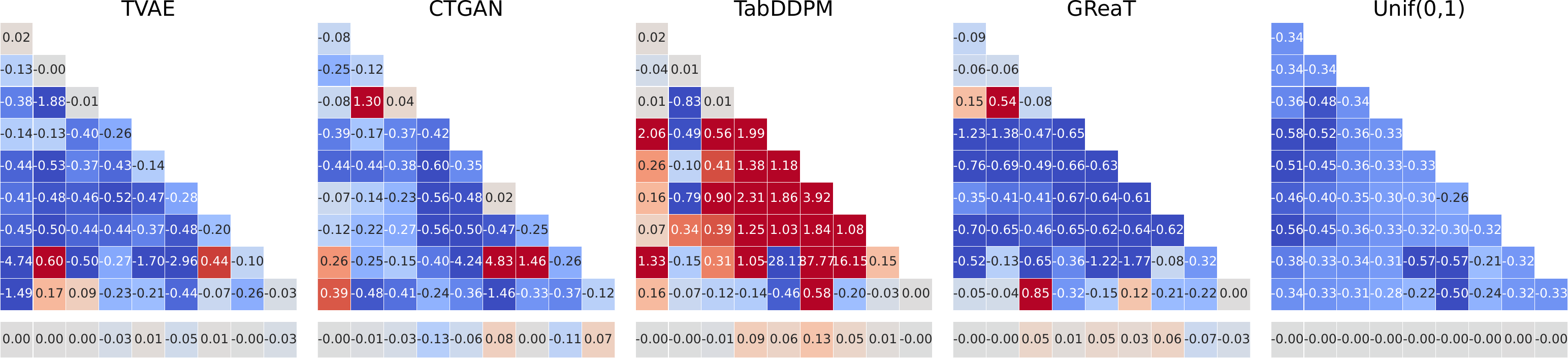}
    \caption{The relative bias of (co)variance (triangle) and expectation (bar) for `CA Housing' with $m=n$.}
\end{figure}

\begin{figure}[h]
    \centering
    \includegraphics[width=\linewidth]{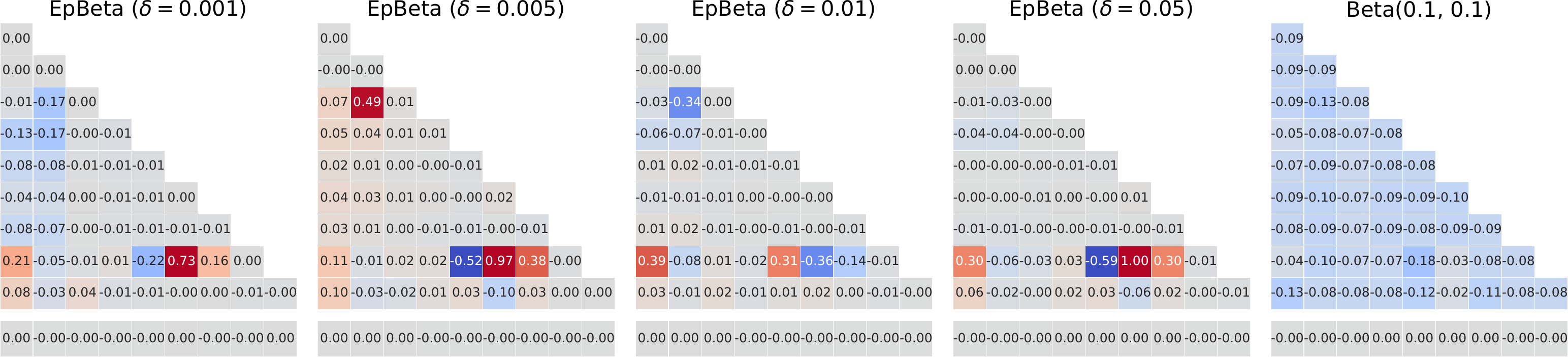}
    \includegraphics[width=\linewidth]{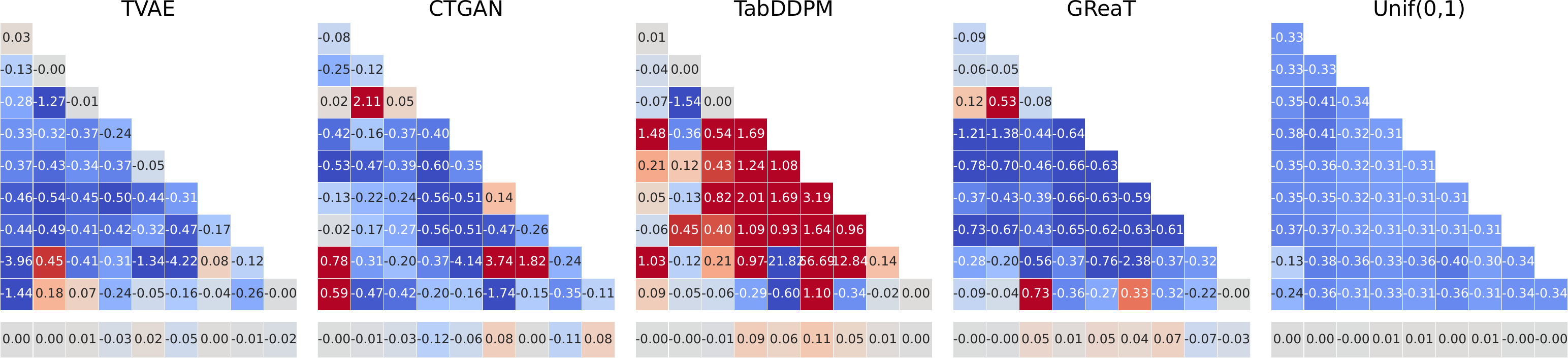}
    \caption{The relative bias of (co)variance (triangle) and expectation (bar) for `CA Housing' with $m=5n$.}
    \vspace{-12pt}
\end{figure}

\clearpage
\begin{figure}[h]
    \centering
    \includegraphics[width=\linewidth]{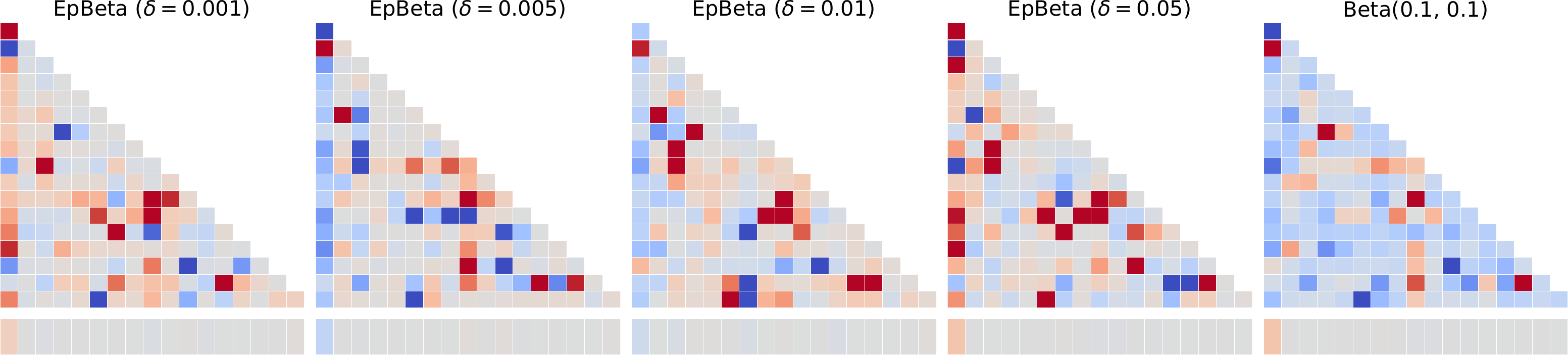}
    \includegraphics[width=\linewidth]{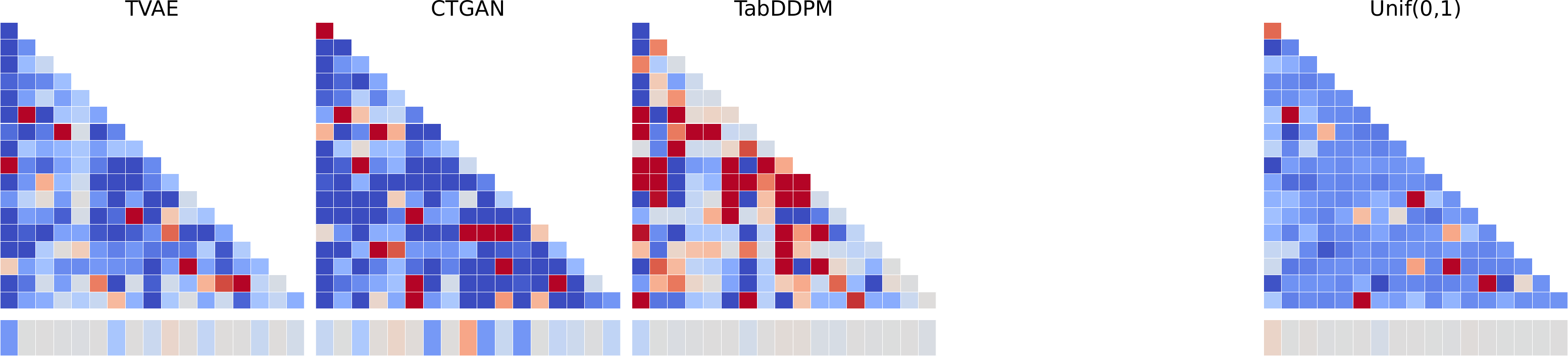}
    \caption{The relative bias of (co)variance (triangle) and expectation (bar) for `House 16H' with $m=n$.}
\end{figure}

\begin{figure}[h]
    \centering
    \includegraphics[width=\linewidth]{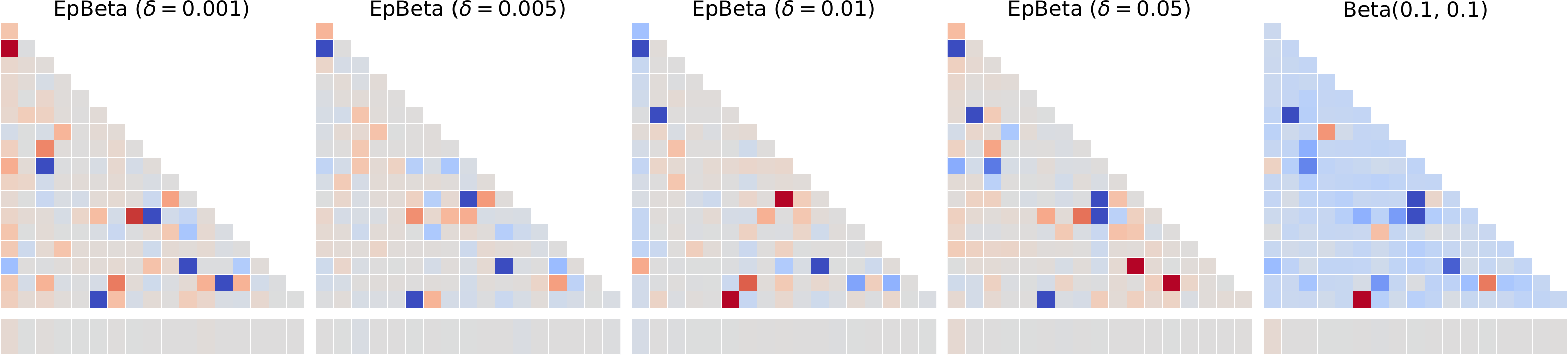}
    \includegraphics[width=\linewidth]{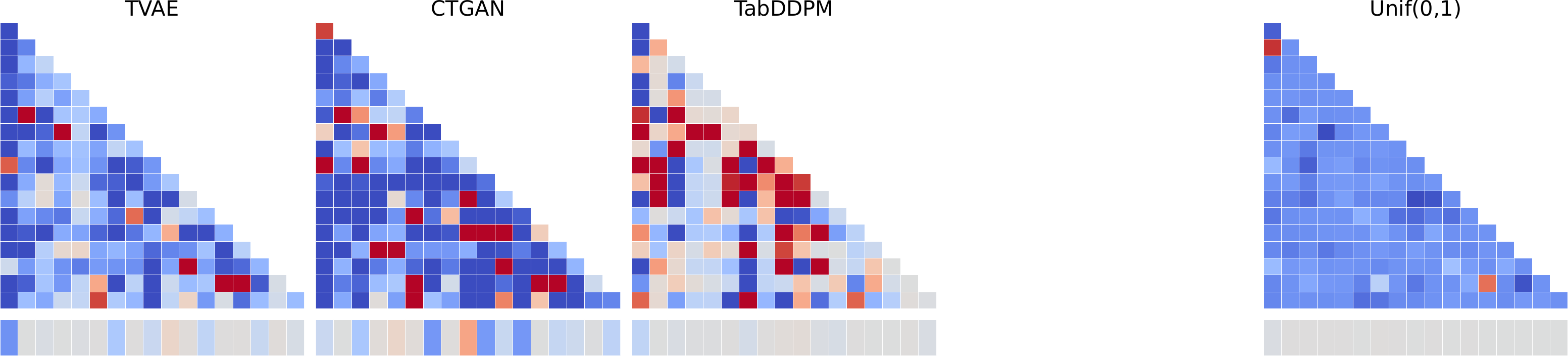}
    \caption{The relative bias of (co)variance (triangle) and expectation (bar) for `House 16H' with $m=5n$.}
    \vspace{-12pt}
\end{figure}

\clearpage
\begin{figure}[h]
    \centering
    \includegraphics[width=\linewidth]{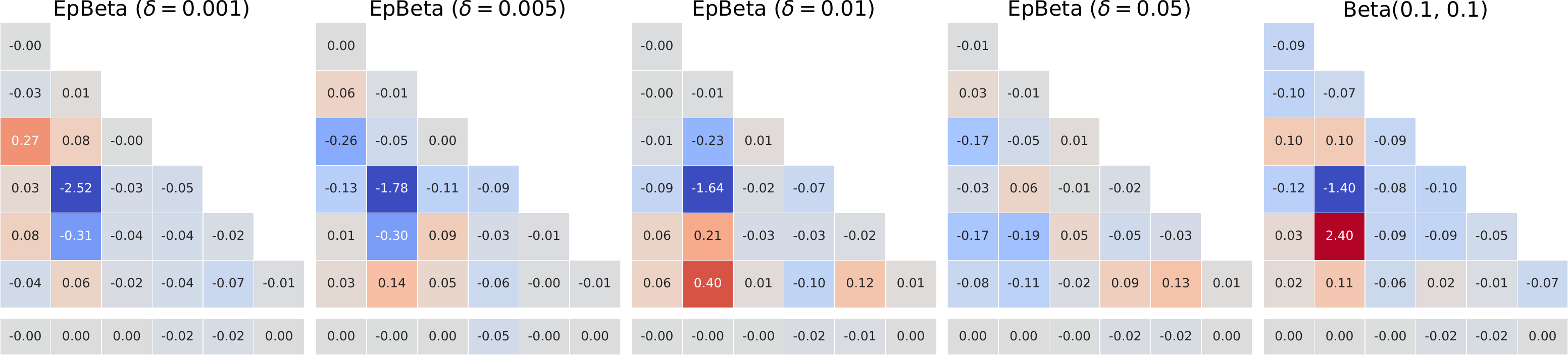}
    \includegraphics[width=\linewidth]{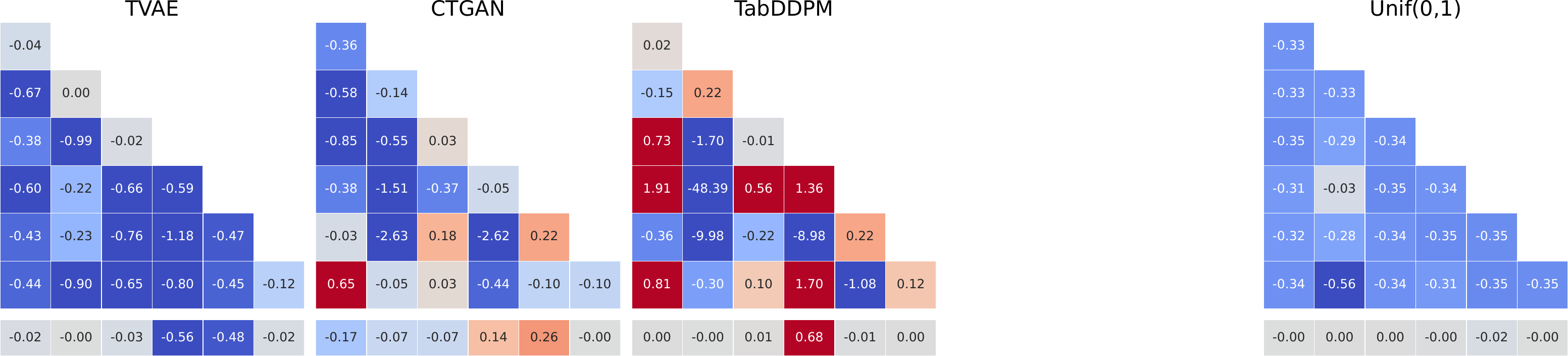}
    \caption{The relative bias of (co)variance (triangle) and expectation (bar) for `Adult' with $m=n$.}
\end{figure}

\begin{figure}[h]
    \centering
    \includegraphics[width=\linewidth]{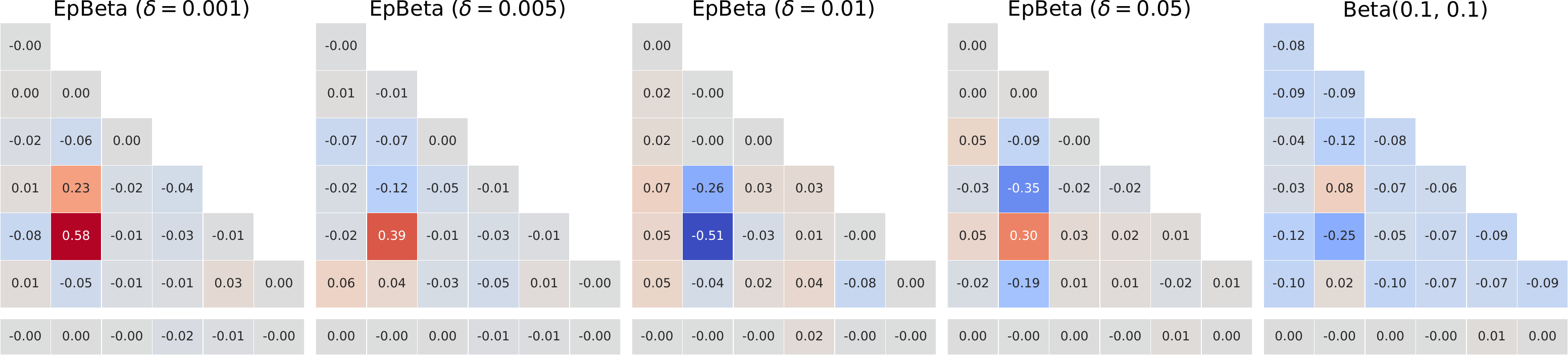}
    \includegraphics[width=\linewidth]{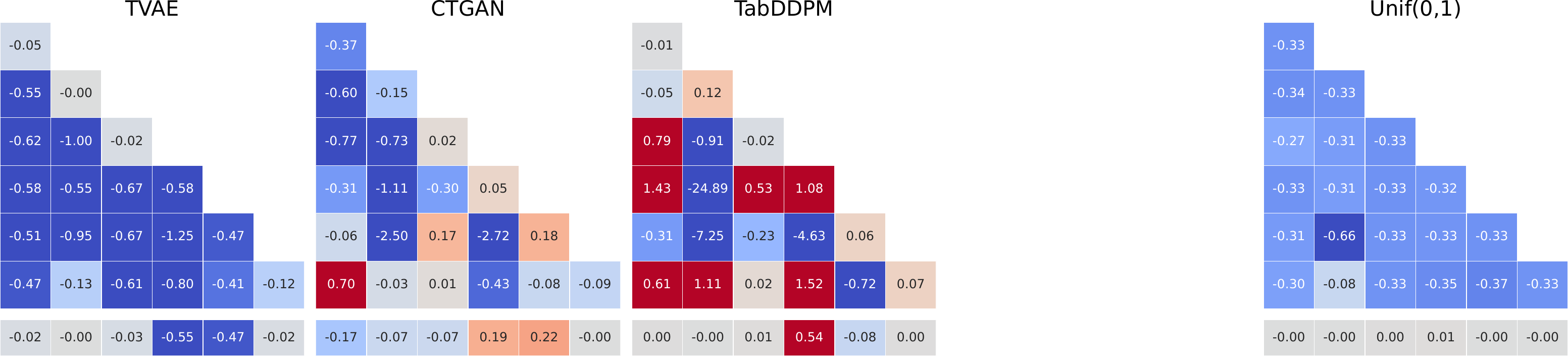}
    \caption{The relative bias of (co)variance (triangle) and expectation (bar) for `Adult' with $m=5n$.}
    \vspace{-12pt}
\end{figure}

\clearpage
\begin{figure}[h]
    \centering
    \includegraphics[width=\linewidth]{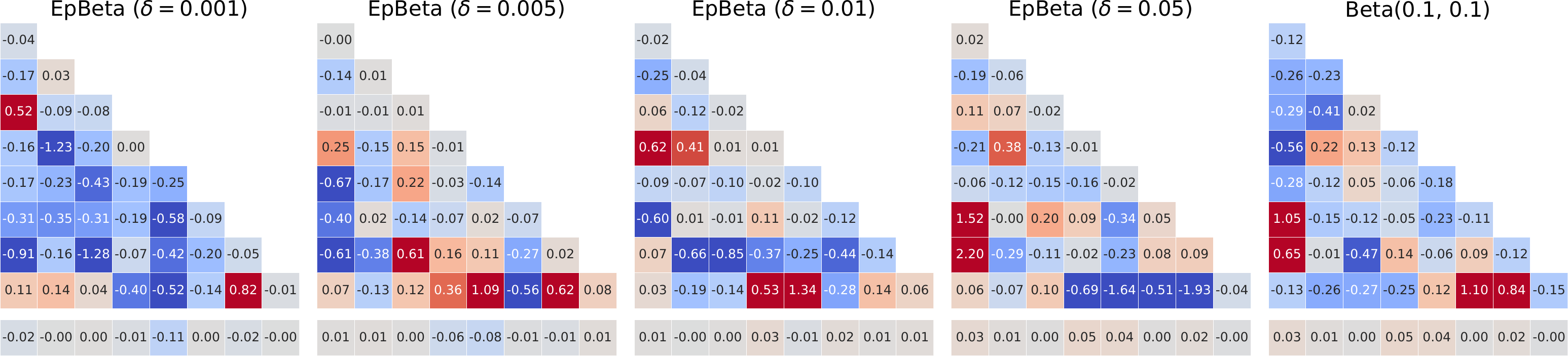}
    \includegraphics[width=\linewidth]{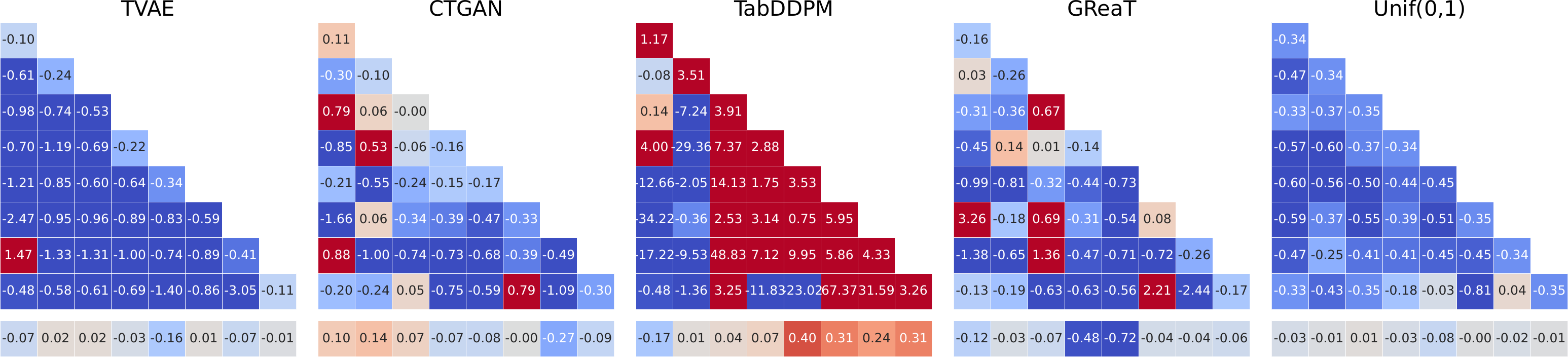}
    \caption{The relative bias of (co)variance (triangle) and expectation (bar) for `Diabetes' with $m=n$.}
\end{figure}

\begin{figure}[h]
    \centering
    \includegraphics[width=\linewidth]{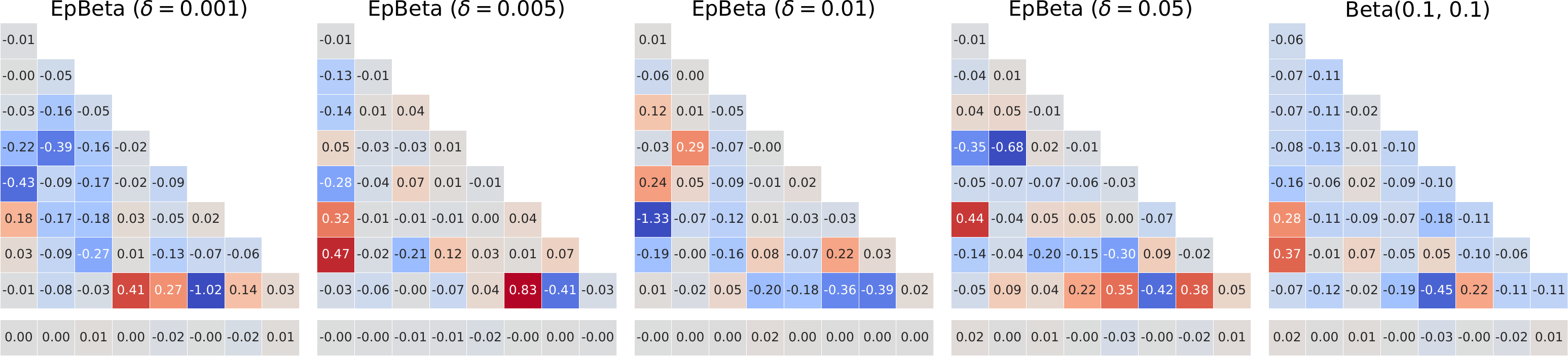}
    \includegraphics[width=\linewidth]{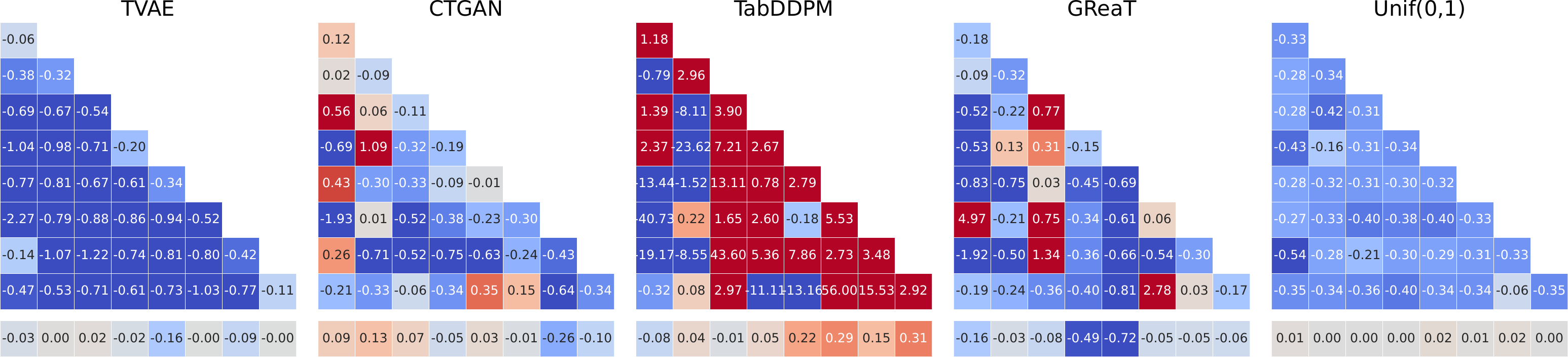}
    \caption{The relative bias of (co)variance (triangle) and expectation (bar) for `Diabetes' with $m=5n$.}
    \vspace{-12pt}
\end{figure}

\clearpage
\begin{figure}[h]
    \centering
    \includegraphics[width=\linewidth]{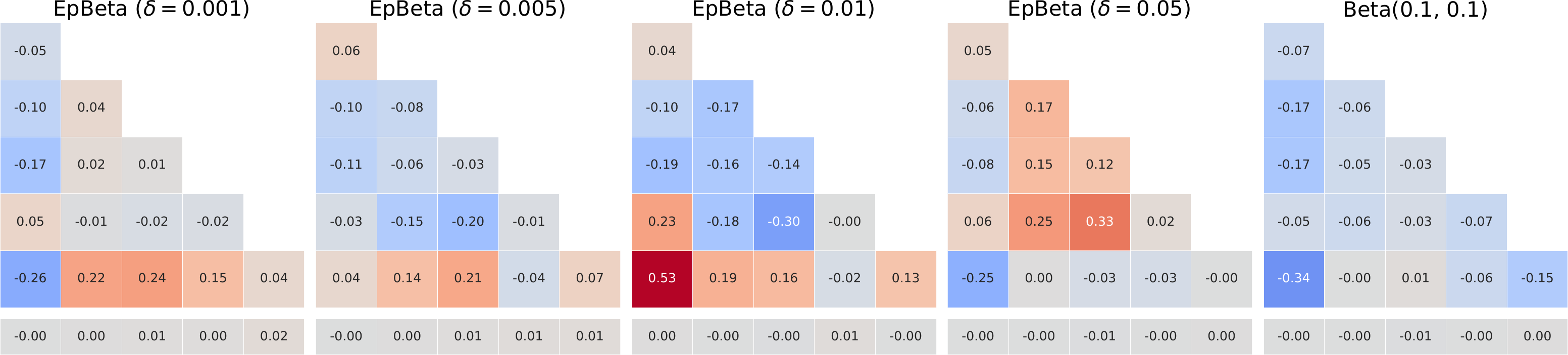}
    \includegraphics[width=\linewidth]{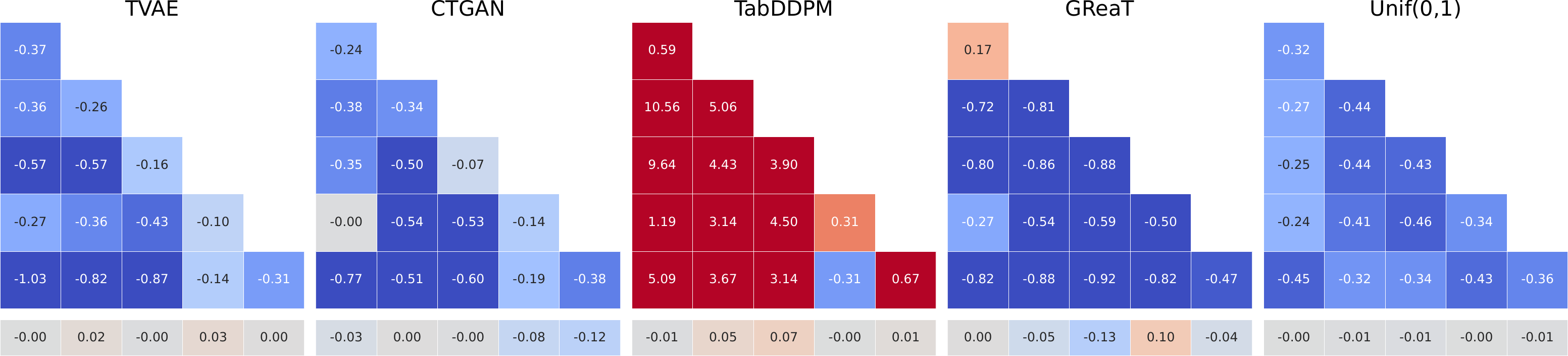}
    \caption{The relative bias of (co)variance (triangle) and expectation (bar) for `Wilt' with $m=n$.}
\end{figure}

\begin{figure}[h]
    \centering
    \includegraphics[width=\linewidth]{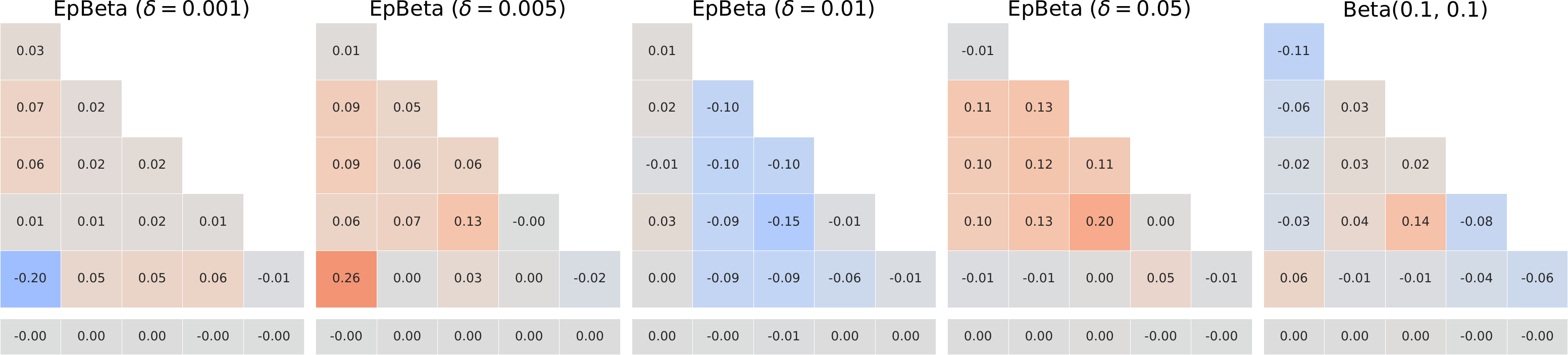}
    \includegraphics[width=\linewidth]{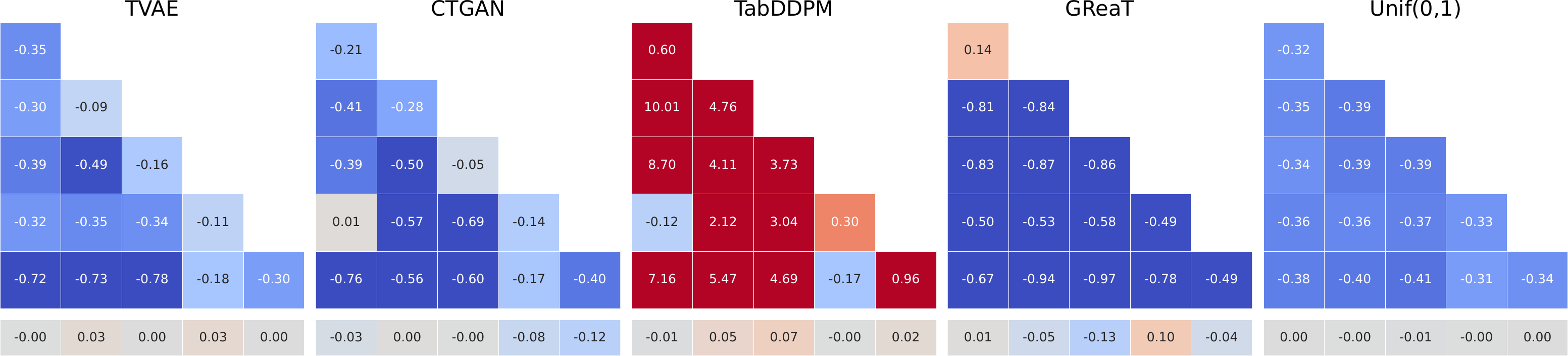}
    \caption{The relative bias of (co)variance (triangle) and expectation (bar) for `Wilt' with $m=5n$.}
    \vspace{-12pt}
\end{figure}

\clearpage

\subsection{Machine Learning Efficiency}
\label{appendix:exp:ml}

We assess machine learning efficiency by training various models on different synthetic datasets, following the experimental protocol of \cite{gorishniy2021revisiting, zhao2021ctab, kotelnikov2023tabddpm}. Our focus is on evaluating how closely each synthetic dataset resembles the original data, rather than on the effectiveness of the trained models using these synthetic datasets.

We select five models for training: a linear model (logistic regression for classification tasks and ridge regression for regression tasks), a decision tree, a random forest, a multilayer perceptron (MLP), and CatBoost \citep{prokhorenkova2018catboost}. We implement four of these models using the Scikit-learn library \citep{scikit-learn}, excluding CatBoost. We employ the same hyperparameters specified in \cite{kotelnikov2023tabddpm}, utilizing a min-max scaler exclusively for training the MLP.

To evaluate model performance, we generate 10 sets of synthetic data for each method. We then individually train five models with different random seeds for each synthetic dataset, resulting in a total of 50 models for each synthetic method. Finally, we evaluate the R-squared value for regression tasks or accuracy for classification tasks against the original data. The table below presents the average and standard deviation of the evaluation metrics for each model. The results indicate that the performance of the mixup-driven synthetic datasets is comparable to that of other machine learning-based synthetic methods.

\begin{table}[h]
\centering
\caption{The performance (R-squared or accuracy) of linear models.}
\begin{tabular}{c|cccccc}
\hline
Name & Abalone & CA Housing & House 16H & Adult & Diabetes & Wilt \\ \hline \hline
Original & \begin{tabular}[c]{@{}c@{}}0.5355\\ (0.0000)\end{tabular} & \begin{tabular}[c]{@{}c@{}}0.6465\\ (0.0000)\end{tabular} & \begin{tabular}[c]{@{}c@{}}0.2527\\ (0.0000)\end{tabular} & \begin{tabular}[c]{@{}c@{}}0.8249\\ (0.0000)\end{tabular} & \begin{tabular}[c]{@{}c@{}}0.7812\\ (0.0000)\end{tabular} & \begin{tabular}[c]{@{}c@{}}0.9682\\ (0.0000)\end{tabular} \\ \hline
\begin{tabular}[c]{@{}c@{}}$\operatorname{EpBeta}$\\ $(\delta=0.001)$\end{tabular} & \begin{tabular}[c]{@{}c@{}}0.5228\\ (0.0023)\end{tabular} & \begin{tabular}[c]{@{}c@{}}0.6360\\ (0.0010)\end{tabular} & \begin{tabular}[c]{@{}c@{}}0.2517\\ (0.0018)\end{tabular} & \begin{tabular}[c]{@{}c@{}}0.8140\\ (0.0205)\end{tabular} & \begin{tabular}[c]{@{}c@{}}0.6503\\ (0.0083)\end{tabular} & \begin{tabular}[c]{@{}c@{}}0.9461\\ (0.0000)\end{tabular} \\
\begin{tabular}[c]{@{}c@{}}$\operatorname{EpBeta}$\\ $(\delta=0.005)$\end{tabular} & \begin{tabular}[c]{@{}c@{}}0.5230\\ (0.0022)\end{tabular} & \begin{tabular}[c]{@{}c@{}}0.6365\\ (0.0004)\end{tabular} & \begin{tabular}[c]{@{}c@{}}0.2495\\ (0.0054)\end{tabular} & \begin{tabular}[c]{@{}c@{}}0.8123\\ (0.0233)\end{tabular} & \begin{tabular}[c]{@{}c@{}}0.6513\\ (0.0084)\end{tabular} & \begin{tabular}[c]{@{}c@{}}0.9461\\ (0.0000)\end{tabular} \\
\begin{tabular}[c]{@{}c@{}}$\operatorname{EpBeta}$\\ $(\delta=0.01)$\end{tabular} & \begin{tabular}[c]{@{}c@{}}0.5237\\ (0.0023)\end{tabular} & \begin{tabular}[c]{@{}c@{}}0.6367\\ (0.0003)\end{tabular} & \begin{tabular}[c]{@{}c@{}}0.2506\\ (0.0026)\end{tabular} & \begin{tabular}[c]{@{}c@{}}0.8106\\ (0.0252)\end{tabular} & \begin{tabular}[c]{@{}c@{}}0.6522\\ (0.0025)\end{tabular} & \begin{tabular}[c]{@{}c@{}}0.9461\\ (0.0000)\end{tabular} \\
\begin{tabular}[c]{@{}c@{}}$\operatorname{EpBeta}$\\ $(\delta=0.05)$\end{tabular} & \begin{tabular}[c]{@{}c@{}}0.5238\\ (0.0023)\end{tabular} & \begin{tabular}[c]{@{}c@{}}0.6364\\ (0.0007)\end{tabular} & \begin{tabular}[c]{@{}c@{}}0.2517\\ (0.0015)\end{tabular} & \begin{tabular}[c]{@{}c@{}}0.8075\\ (0.0243)\end{tabular} & \begin{tabular}[c]{@{}c@{}}0.6561\\ (0.0060)\end{tabular} & \begin{tabular}[c]{@{}c@{}}0.9461\\ (0.0000)\end{tabular} \\ \hline
$\operatorname{Beta}$ & \begin{tabular}[c]{@{}c@{}}0.5247\\ (0.0027)\end{tabular} & \begin{tabular}[c]{@{}c@{}}0.6368\\ (0.0005)\end{tabular} & \begin{tabular}[c]{@{}c@{}}0.2500\\ (0.0012)\end{tabular} & \begin{tabular}[c]{@{}c@{}}0.8022\\ (0.0264)\end{tabular} & \begin{tabular}[c]{@{}c@{}}0.6527\\ (0.0032)\end{tabular} & \begin{tabular}[c]{@{}c@{}}0.9461\\ (0.0000)\end{tabular} \\
$\operatorname{Unif}$ & \begin{tabular}[c]{@{}c@{}}0.5264\\ (0.0014)\end{tabular} & \begin{tabular}[c]{@{}c@{}}0.6380\\ (0.0005)\end{tabular} & \begin{tabular}[c]{@{}c@{}}0.2490\\ (0.0009)\end{tabular} & \begin{tabular}[c]{@{}c@{}}0.8159\\ (0.0143)\end{tabular} & \begin{tabular}[c]{@{}c@{}}0.6978\\ (0.0264)\end{tabular} & \begin{tabular}[c]{@{}c@{}}0.9452\\ (0.0003)\end{tabular} \\ \hline
TVAE & \begin{tabular}[c]{@{}c@{}}0.4006\\ (0.0080)\end{tabular} & \begin{tabular}[c]{@{}c@{}}0.6009\\ (0.0025)\end{tabular} & \begin{tabular}[c]{@{}c@{}}0.1634\\ (0.0173)\end{tabular} & \begin{tabular}[c]{@{}c@{}}0.7963\\ (0.0118)\end{tabular} & \begin{tabular}[c]{@{}c@{}}0.7400\\ (0.0170)\end{tabular} & \begin{tabular}[c]{@{}c@{}}0.9461\\ (0.0000)\end{tabular} \\
CTGAN & \begin{tabular}[c]{@{}c@{}}0.3954\\ (0.0120)\end{tabular} & \begin{tabular}[c]{@{}c@{}}0.5122\\ (0.0027)\end{tabular} & \begin{tabular}[c]{@{}c@{}}0.1338\\ (0.0038)\end{tabular} & \begin{tabular}[c]{@{}c@{}}0.8151\\ (0.0056)\end{tabular} & \begin{tabular}[c]{@{}c@{}}0.7671\\ (0.0117)\end{tabular} & \begin{tabular}[c]{@{}c@{}}0.9461\\ (0.0000)\end{tabular} \\
TabDDPM & \begin{tabular}[c]{@{}c@{}}0.3829\\ (0.0131)\end{tabular} & \begin{tabular}[c]{@{}c@{}}0.6036\\ (0.0040)\end{tabular} & \begin{tabular}[c]{@{}c@{}}0.2216\\ (0.0110)\end{tabular} & \begin{tabular}[c]{@{}c@{}}0.8207\\ (0.0068)\end{tabular} & \begin{tabular}[c]{@{}c@{}}0.7469\\ (0.0087)\end{tabular} & \begin{tabular}[c]{@{}c@{}}0.9482\\ (0.0035)\end{tabular} \\
GReaT & \begin{tabular}[c]{@{}c@{}}0.5156\\ (0.0026)\end{tabular} & \begin{tabular}[c]{@{}c@{}}0.6364\\ (0.0018)\end{tabular} & - & - & \begin{tabular}[c]{@{}c@{}}0.7520\\ (0.0093)\end{tabular} & \begin{tabular}[c]{@{}c@{}}0.9570\\ (0.0066)\end{tabular} \\ \hline
\end{tabular}
\end{table}

\begin{table}
\centering
\caption{The performance (R-squared or accuracy) of tree models.}
\begin{tabular}{c|cccccc} 
\hline
Name & Abalone & CA Housing & House 16H & Adult & Diabetes & Wilt \\ 
\hline\hline
Original                                                                           & \begin{tabular}[c]{@{}c@{}}1.0000\\ (0.0000)\end{tabular}  & \begin{tabular}[c]{@{}c@{}}1.0000\\ (0.0000)\end{tabular}  & \begin{tabular}[c]{@{}c@{}}0.9999\\ (0.0000)\end{tabular}  & \begin{tabular}[c]{@{}c@{}}0.9731\\ (0.0001)\end{tabular} & \begin{tabular}[c]{@{}c@{}}1.0000\\ (0.0000)\end{tabular} & \begin{tabular}[c]{@{}c@{}}1.0000\\ (0.0000)\end{tabular}  \\ 
\hline
\begin{tabular}[c]{@{}c@{}}$\operatorname{EpBeta}$\\ $(\delta=0.001)$\end{tabular} & \begin{tabular}[c]{@{}c@{}}0.4727\\ (0.0226)\end{tabular}  & \begin{tabular}[c]{@{}c@{}}0.7334\\ (0.0081)\end{tabular}  & \begin{tabular}[c]{@{}c@{}}0.5852\\ (0.0106)\end{tabular}  & \begin{tabular}[c]{@{}c@{}}0.8001\\ (0.0016)\end{tabular} & \begin{tabular}[c]{@{}c@{}}0.5719\\ (0.0213)\end{tabular} & \begin{tabular}[c]{@{}c@{}}0.9014\\ (0.0083)\end{tabular}  \\
\begin{tabular}[c]{@{}c@{}}$\operatorname{EpBeta}$\\ $(\delta=0.005)$\end{tabular} & \begin{tabular}[c]{@{}c@{}}0.3721\\ (0.0297)\end{tabular}  & \begin{tabular}[c]{@{}c@{}}0.6642\\ (0.0074)\end{tabular}  & \begin{tabular}[c]{@{}c@{}}0.5196\\ (0.0218)\end{tabular}  & \begin{tabular}[c]{@{}c@{}}0.8010\\ (0.0032)\end{tabular} & \begin{tabular}[c]{@{}c@{}}0.5588\\ (0.0169)\end{tabular} & \begin{tabular}[c]{@{}c@{}}0.8990\\ (0.0070)\end{tabular}  \\
\begin{tabular}[c]{@{}c@{}}$\operatorname{EpBeta}$\\ $(\delta=0.01)$\end{tabular}  & \begin{tabular}[c]{@{}c@{}}0.3496\\ (0.0352)\end{tabular}  & \begin{tabular}[c]{@{}c@{}}0.6327\\ (0.0082)\end{tabular}  & \begin{tabular}[c]{@{}c@{}}0.4903\\ (0.0306)\end{tabular}  & \begin{tabular}[c]{@{}c@{}}0.8000\\ (0.0028)\end{tabular} & \begin{tabular}[c]{@{}c@{}}0.5572\\ (0.0274)\end{tabular} & \begin{tabular}[c]{@{}c@{}}0.8999\\ (0.0059)\end{tabular}  \\
\begin{tabular}[c]{@{}c@{}}$\operatorname{EpBeta}$\\ $(\delta=0.05)$\end{tabular}  & \begin{tabular}[c]{@{}c@{}}0.2367\\ (0.0308)\end{tabular}  & \begin{tabular}[c]{@{}c@{}}0.5315\\ (0.0102)\end{tabular}  & \begin{tabular}[c]{@{}c@{}}0.3722\\ (0.0288)\end{tabular}  & \begin{tabular}[c]{@{}c@{}}0.8052\\ (0.0021)\end{tabular} & \begin{tabular}[c]{@{}c@{}}0.5742\\ (0.0286)\end{tabular} & \begin{tabular}[c]{@{}c@{}}0.9027\\ (0.0052)\end{tabular}  \\ 
\hline
$\operatorname{Beta}$                                                              & \begin{tabular}[c]{@{}c@{}}0.5455\\ (0.0267)\end{tabular}  & \begin{tabular}[c]{@{}c@{}}0.7709\\ (0.0081)\end{tabular}  & \begin{tabular}[c]{@{}c@{}}0.5956\\ (0.0216)\end{tabular}  & \begin{tabular}[c]{@{}c@{}}0.7952\\ (0.0031)\end{tabular} & \begin{tabular}[c]{@{}c@{}}0.5558\\ (0.0215)\end{tabular} & \begin{tabular}[c]{@{}c@{}}0.8987\\ (0.0055)\end{tabular}  \\
$\operatorname{Unif}$                                                              & \begin{tabular}[c]{@{}c@{}}0.2969\\ (0.0367)\end{tabular}  & \begin{tabular}[c]{@{}c@{}}0.5138\\ (0.0099)\end{tabular}  & \begin{tabular}[c]{@{}c@{}}0.2477\\ (0.0388)\end{tabular}  & \begin{tabular}[c]{@{}c@{}}0.7979\\ (0.0036)\end{tabular} & \begin{tabular}[c]{@{}c@{}}0.5796\\ (0.0352)\end{tabular} & \begin{tabular}[c]{@{}c@{}}0.9047\\ (0.0034)\end{tabular}  \\ 
\hline
TVAE                                                                               & \begin{tabular}[c]{@{}c@{}}-0.0660\\ (0.0485)\end{tabular} & \begin{tabular}[c]{@{}c@{}}0.2159\\ (0.0198)\end{tabular}  & \begin{tabular}[c]{@{}c@{}}-0.1702\\ (0.0528)\end{tabular} & \begin{tabular}[c]{@{}c@{}}0.7584\\ (0.0046)\end{tabular} & \begin{tabular}[c]{@{}c@{}}0.6024\\ (0.0329)\end{tabular} & \begin{tabular}[c]{@{}c@{}}0.9395\\ (0.0085)\end{tabular}  \\
CTGAN                                                                              & \begin{tabular}[c]{@{}c@{}}-0.1381\\ (0.0528)\end{tabular} & \begin{tabular}[c]{@{}c@{}}-0.0740\\ (0.0289)\end{tabular} & \begin{tabular}[c]{@{}c@{}}-0.6111\\ (0.1483)\end{tabular} & \begin{tabular}[c]{@{}c@{}}0.7390\\ (0.0067)\end{tabular} & \begin{tabular}[c]{@{}c@{}}0.6295\\ (0.0265)\end{tabular} & \begin{tabular}[c]{@{}c@{}}0.9314\\ (0.0053)\end{tabular}  \\
TabDDPM                                                                            & \begin{tabular}[c]{@{}c@{}}0.1679\\ (0.0369)\end{tabular}  & \begin{tabular}[c]{@{}c@{}}0.4876\\ (0.0096)\end{tabular}  & \begin{tabular}[c]{@{}c@{}}0.1382\\ (0.0318)\end{tabular}  & \begin{tabular}[c]{@{}c@{}}0.7975\\ (0.0020)\end{tabular} & \begin{tabular}[c]{@{}c@{}}0.6910\\ (0.0152)\end{tabular} & \begin{tabular}[c]{@{}c@{}}0.9780\\ (0.0025)\end{tabular}  \\
GReaT                                                                              & \begin{tabular}[c]{@{}c@{}}0.0193\\ (0.0455)\end{tabular}  & \begin{tabular}[c]{@{}c@{}}0.5312\\ (0.0122)\end{tabular}  & -                                                          & -                                                         & \begin{tabular}[c]{@{}c@{}}0.6826\\ (0.0147)\end{tabular} & \begin{tabular}[c]{@{}c@{}}0.9345\\ (0.0164)\end{tabular}  \\
\hline
\end{tabular}
\end{table}

\begin{table}
\centering
\caption{The performance (R-squared or accuracy) of random forest models.}
\begin{tabular}{c|cccccc} 
\hline
Name & Abalone & CA Housing & House 16H & Adult & Diabetes & Wilt \\ 
\hline\hline
Original                                                                           & \begin{tabular}[c]{@{}c@{}}0.9356\\ (0.0012)\end{tabular} & \begin{tabular}[c]{@{}c@{}}0.9758\\ (0.0001)\end{tabular} & \begin{tabular}[c]{@{}c@{}}0.9496\\ (0.0007)\end{tabular} & \begin{tabular}[c]{@{}c@{}}0.9550\\ (0.0008)\end{tabular} & \begin{tabular}[c]{@{}c@{}}1.0000\\ (0.0000)\end{tabular} & \begin{tabular}[c]{@{}c@{}}1.0000\\ (0.0000)\end{tabular}  \\ 
\hline
\begin{tabular}[c]{@{}c@{}}$\operatorname{EpBeta}$\\ $(\delta=0.001)$\end{tabular} & \begin{tabular}[c]{@{}c@{}}0.7091\\ (0.0063)\end{tabular} & \begin{tabular}[c]{@{}c@{}}0.8676\\ (0.0027)\end{tabular} & \begin{tabular}[c]{@{}c@{}}0.7970\\ (0.0079)\end{tabular} & \begin{tabular}[c]{@{}c@{}}0.8404\\ (0.0007)\end{tabular} & \begin{tabular}[c]{@{}c@{}}0.6138\\ (0.0169)\end{tabular} & \begin{tabular}[c]{@{}c@{}}0.9407\\ (0.0014)\end{tabular}  \\
\begin{tabular}[c]{@{}c@{}}$\operatorname{EpBeta}$\\ $(\delta=0.005)$\end{tabular} & \begin{tabular}[c]{@{}c@{}}0.6742\\ (0.0056)\end{tabular} & \begin{tabular}[c]{@{}c@{}}0.8403\\ (0.0017)\end{tabular} & \begin{tabular}[c]{@{}c@{}}0.7680\\ (0.0082)\end{tabular} & \begin{tabular}[c]{@{}c@{}}0.8405\\ (0.0009)\end{tabular} & \begin{tabular}[c]{@{}c@{}}0.6170\\ (0.0162)\end{tabular} & \begin{tabular}[c]{@{}c@{}}0.9429\\ (0.0009)\end{tabular}  \\
\begin{tabular}[c]{@{}c@{}}$\operatorname{EpBeta}$\\ $(\delta=0.01)$\end{tabular}  & \begin{tabular}[c]{@{}c@{}}0.6542\\ (0.0055)\end{tabular} & \begin{tabular}[c]{@{}c@{}}0.8240\\ (0.0028)\end{tabular} & \begin{tabular}[c]{@{}c@{}}0.7509\\ (0.0079)\end{tabular} & \begin{tabular}[c]{@{}c@{}}0.8409\\ (0.0009)\end{tabular} & \begin{tabular}[c]{@{}c@{}}0.6102\\ (0.0177)\end{tabular} & \begin{tabular}[c]{@{}c@{}}0.9438\\ (0.0008)\end{tabular}  \\
\begin{tabular}[c]{@{}c@{}}$\operatorname{EpBeta}$\\ $(\delta=0.05)$\end{tabular}  & \begin{tabular}[c]{@{}c@{}}0.6066\\ (0.0064)\end{tabular} & \begin{tabular}[c]{@{}c@{}}0.7751\\ (0.0028)\end{tabular} & \begin{tabular}[c]{@{}c@{}}0.6932\\ (0.0047)\end{tabular} & \begin{tabular}[c]{@{}c@{}}0.8438\\ (0.0010)\end{tabular} & \begin{tabular}[c]{@{}c@{}}0.6321\\ (0.0279)\end{tabular} & \begin{tabular}[c]{@{}c@{}}0.9451\\ (0.0007)\end{tabular}  \\ 
\hline
$\operatorname{Beta}$                                                              & \begin{tabular}[c]{@{}c@{}}0.7334\\ (0.0103)\end{tabular} & \begin{tabular}[c]{@{}c@{}}0.8758\\ (0.0029)\end{tabular} & \begin{tabular}[c]{@{}c@{}}0.7973\\ (0.0061)\end{tabular} & \begin{tabular}[c]{@{}c@{}}0.8425\\ (0.0011)\end{tabular} & \begin{tabular}[c]{@{}c@{}}0.5883\\ (0.0185)\end{tabular} & \begin{tabular}[c]{@{}c@{}}0.9247\\ (0.0020)\end{tabular}  \\
$\operatorname{Unif}$                                                              & \begin{tabular}[c]{@{}c@{}}0.5866\\ (0.0061)\end{tabular} & \begin{tabular}[c]{@{}c@{}}0.7370\\ (0.0029)\end{tabular} & \begin{tabular}[c]{@{}c@{}}0.6423\\ (0.0088)\end{tabular} & \begin{tabular}[c]{@{}c@{}}0.8481\\ (0.0009)\end{tabular} & \begin{tabular}[c]{@{}c@{}}0.6567\\ (0.0257)\end{tabular} & \begin{tabular}[c]{@{}c@{}}0.9449\\ (0.0008)\end{tabular}  \\ 
\hline
TVAE                                                                               & \begin{tabular}[c]{@{}c@{}}0.4213\\ (0.0047)\end{tabular} & \begin{tabular}[c]{@{}c@{}}0.6834\\ (0.0054)\end{tabular} & \begin{tabular}[c]{@{}c@{}}0.4450\\ (0.0102)\end{tabular} & \begin{tabular}[c]{@{}c@{}}0.8169\\ (0.0023)\end{tabular} & \begin{tabular}[c]{@{}c@{}}0.7075\\ (0.0161)\end{tabular} & \begin{tabular}[c]{@{}c@{}}0.9700\\ (0.0016)\end{tabular}  \\
CTGAN                                                                              & \begin{tabular}[c]{@{}c@{}}0.4263\\ (0.0121)\end{tabular} & \begin{tabular}[c]{@{}c@{}}0.4746\\ (0.0107)\end{tabular} & \begin{tabular}[c]{@{}c@{}}0.2409\\ (0.0136)\end{tabular} & \begin{tabular}[c]{@{}c@{}}0.8336\\ (0.0018)\end{tabular} & \begin{tabular}[c]{@{}c@{}}0.7408\\ (0.0172)\end{tabular} & \begin{tabular}[c]{@{}c@{}}0.9663\\ (0.0022)\end{tabular}  \\
TabDDPM                                                                            & \begin{tabular}[c]{@{}c@{}}0.5351\\ (0.0037)\end{tabular} & \begin{tabular}[c]{@{}c@{}}0.7493\\ (0.0018)\end{tabular} & \begin{tabular}[c]{@{}c@{}}0.6017\\ (0.0058)\end{tabular} & \begin{tabular}[c]{@{}c@{}}0.8484\\ (0.0014)\end{tabular} & \begin{tabular}[c]{@{}c@{}}0.7589\\ (0.0084)\end{tabular} & \begin{tabular}[c]{@{}c@{}}0.9850\\ (0.0013)\end{tabular}  \\
GReaT                                                                              & \begin{tabular}[c]{@{}c@{}}0.5047\\ (0.0120)\end{tabular} & \begin{tabular}[c]{@{}c@{}}0.7809\\ (0.0033)\end{tabular} & -                                                         & -                                                         & \begin{tabular}[c]{@{}c@{}}0.7554\\ (0.0114)\end{tabular} & \begin{tabular}[c]{@{}c@{}}0.9696\\ (0.0064)\end{tabular}  \\
\hline
\end{tabular}
\end{table}

\begin{table}
\centering
\caption{The performance (R-squared or accuracy) of MLP models.}
\begin{tabular}{c|cccccc} 
\hline
Name & Abalone & CA Housing & House 16H & Adult & Diabetes & Wilt \\ 
\hline\hline
Original                                                                           & \begin{tabular}[c]{@{}c@{}}0.5404\\ (0.0064)\end{tabular} & \begin{tabular}[c]{@{}c@{}}0.7324\\ (0.0069)\end{tabular} & \begin{tabular}[c]{@{}c@{}}0.4838\\ (0.0058)\end{tabular} & \begin{tabular}[c]{@{}c@{}}0.8767\\ (0.0006)\end{tabular} & \begin{tabular}[c]{@{}c@{}}0.7792\\ (0.0043)\end{tabular} & \begin{tabular}[c]{@{}c@{}}0.9461\\ (0.0000)\end{tabular}  \\ 
\hline
\begin{tabular}[c]{@{}c@{}}$\operatorname{EpBeta}$\\ $(\delta=0.001)$\end{tabular} & \begin{tabular}[c]{@{}c@{}}0.5155\\ (0.0117)\end{tabular} & \begin{tabular}[c]{@{}c@{}}0.7019\\ (0.0126)\end{tabular} & \begin{tabular}[c]{@{}c@{}}0.4564\\ (0.0104)\end{tabular} & \begin{tabular}[c]{@{}c@{}}0.8403\\ (0.0015)\end{tabular} & \begin{tabular}[c]{@{}c@{}}0.6510\\ (0.0016)\end{tabular} & \begin{tabular}[c]{@{}c@{}}0.9461\\ (0.0000)\end{tabular}  \\
\begin{tabular}[c]{@{}c@{}}$\operatorname{EpBeta}$\\ $(\delta=0.005)$\end{tabular} & \begin{tabular}[c]{@{}c@{}}0.5119\\ (0.0146)\end{tabular} & \begin{tabular}[c]{@{}c@{}}0.6902\\ (0.0098)\end{tabular} & \begin{tabular}[c]{@{}c@{}}0.4363\\ (0.0154)\end{tabular} & \begin{tabular}[c]{@{}c@{}}0.8406\\ (0.0019)\end{tabular} & \begin{tabular}[c]{@{}c@{}}0.6518\\ (0.0029)\end{tabular} & \begin{tabular}[c]{@{}c@{}}0.9461\\ (0.0000)\end{tabular}  \\
\begin{tabular}[c]{@{}c@{}}$\operatorname{EpBeta}$\\ $(\delta=0.01)$\end{tabular}  & \begin{tabular}[c]{@{}c@{}}0.5090\\ (0.0163)\end{tabular} & \begin{tabular}[c]{@{}c@{}}0.6851\\ (0.0093)\end{tabular} & \begin{tabular}[c]{@{}c@{}}0.4298\\ (0.0124)\end{tabular} & \begin{tabular}[c]{@{}c@{}}0.8414\\ (0.0011)\end{tabular} & \begin{tabular}[c]{@{}c@{}}0.6514\\ (0.0016)\end{tabular} & \begin{tabular}[c]{@{}c@{}}0.9461\\ (0.0000)\end{tabular}  \\
\begin{tabular}[c]{@{}c@{}}$\operatorname{EpBeta}$\\ $(\delta=0.05)$\end{tabular}  & \begin{tabular}[c]{@{}c@{}}0.4979\\ (0.0201)\end{tabular} & \begin{tabular}[c]{@{}c@{}}0.6697\\ (0.0090)\end{tabular} & \begin{tabular}[c]{@{}c@{}}0.3963\\ (0.0226)\end{tabular} & \begin{tabular}[c]{@{}c@{}}0.8424\\ (0.0016)\end{tabular} & \begin{tabular}[c]{@{}c@{}}0.6511\\ (0.0008)\end{tabular} & \begin{tabular}[c]{@{}c@{}}0.9461\\ (0.0000)\end{tabular}  \\ 
\hline
$\operatorname{Beta}$                                                              & \begin{tabular}[c]{@{}c@{}}0.5201\\ (0.0121)\end{tabular} & \begin{tabular}[c]{@{}c@{}}0.7051\\ (0.0118)\end{tabular} & \begin{tabular}[c]{@{}c@{}}0.4606\\ (0.0149)\end{tabular} & \begin{tabular}[c]{@{}c@{}}0.8430\\ (0.0018)\end{tabular} & \begin{tabular}[c]{@{}c@{}}0.6513\\ (0.0018)\end{tabular} & \begin{tabular}[c]{@{}c@{}}0.9461\\ (0.0000)\end{tabular}  \\
$\operatorname{Unif}$                                                              & \begin{tabular}[c]{@{}c@{}}0.5113\\ (0.0146)\end{tabular} & \begin{tabular}[c]{@{}c@{}}0.6834\\ (0.0085)\end{tabular} & \begin{tabular}[c]{@{}c@{}}0.4310\\ (0.0115)\end{tabular} & \begin{tabular}[c]{@{}c@{}}0.8500\\ (0.0019)\end{tabular} & \begin{tabular}[c]{@{}c@{}}0.6795\\ (0.0166)\end{tabular} & \begin{tabular}[c]{@{}c@{}}0.9461\\ (0.0000)\end{tabular}  \\ 
\hline
TVAE                                                                               & \begin{tabular}[c]{@{}c@{}}0.4085\\ (0.0195)\end{tabular} & \begin{tabular}[c]{@{}c@{}}0.6341\\ (0.0094)\end{tabular} & \begin{tabular}[c]{@{}c@{}}0.3387\\ (0.0240)\end{tabular} & \begin{tabular}[c]{@{}c@{}}0.8167\\ (0.0059)\end{tabular} & \begin{tabular}[c]{@{}c@{}}0.7124\\ (0.0181)\end{tabular} & \begin{tabular}[c]{@{}c@{}}0.9461\\ (0.0000)\end{tabular}  \\
CTGAN                                                                              & \begin{tabular}[c]{@{}c@{}}0.3971\\ (0.0287)\end{tabular} & \begin{tabular}[c]{@{}c@{}}0.5090\\ (0.0352)\end{tabular} & \begin{tabular}[c]{@{}c@{}}0.2411\\ (0.0145)\end{tabular} & \begin{tabular}[c]{@{}c@{}}0.8125\\ (0.0060)\end{tabular} & \begin{tabular}[c]{@{}c@{}}0.7641\\ (0.0123)\end{tabular} & \begin{tabular}[c]{@{}c@{}}0.9461\\ (0.0000)\end{tabular}  \\
TabDDPM                                                                            & \begin{tabular}[c]{@{}c@{}}0.4742\\ (0.0241)\end{tabular} & \begin{tabular}[c]{@{}c@{}}0.7133\\ (0.0061)\end{tabular} & \begin{tabular}[c]{@{}c@{}}0.4478\\ (0.0120)\end{tabular} & \begin{tabular}[c]{@{}c@{}}0.8404\\ (0.0019)\end{tabular} & \begin{tabular}[c]{@{}c@{}}0.7527\\ (0.0115)\end{tabular} & \begin{tabular}[c]{@{}c@{}}0.9461\\ (0.0000)\end{tabular}  \\
GReaT                                                                              & \begin{tabular}[c]{@{}c@{}}0.5086\\ (0.0152)\end{tabular} & \begin{tabular}[c]{@{}c@{}}0.7117\\ (0.0093)\end{tabular} & -                                                         & -                                                         & \begin{tabular}[c]{@{}c@{}}0.7464\\ (0.0099)\end{tabular} & \begin{tabular}[c]{@{}c@{}}0.9053\\ (0.0203)\end{tabular}  \\
\hline
\end{tabular}
\end{table}

\begin{table}
\centering
\caption{The performance (R-squared or accuracy) of CatBoost models.}
\begin{tabular}{c|cccccc} 
\hline
Name & Abalone & CA Housing & House 16H & Adult & Diabetes & Wilt \\ 
\hline\hline
Original                                                                           & \begin{tabular}[c]{@{}c@{}}0.8188\\ (0.0007)\end{tabular} & \begin{tabular}[c]{@{}c@{}}0.9736\\ (0.0001)\end{tabular} & \begin{tabular}[c]{@{}c@{}}0.9618\\ (0.0001)\end{tabular} & \begin{tabular}[c]{@{}c@{}}0.8990\\ (0.0004)\end{tabular} & \begin{tabular}[c]{@{}c@{}}0.9958\\ (0.0005)\end{tabular} & \begin{tabular}[c]{@{}c@{}}1.0000\\ (0.0000)\end{tabular}  \\ 
\hline
\begin{tabular}[c]{@{}c@{}}$\operatorname{EpBeta}$\\ $(\delta=0.001)$\end{tabular} & \begin{tabular}[c]{@{}c@{}}0.6796\\ (0.0070)\end{tabular} & \begin{tabular}[c]{@{}c@{}}0.8945\\ (0.0019)\end{tabular} & \begin{tabular}[c]{@{}c@{}}0.8411\\ (0.0056)\end{tabular} & \begin{tabular}[c]{@{}c@{}}0.8368\\ (0.0012)\end{tabular} & \begin{tabular}[c]{@{}c@{}}0.6138\\ (0.0224)\end{tabular} & \begin{tabular}[c]{@{}c@{}}0.9288\\ (0.0021)\end{tabular}  \\
\begin{tabular}[c]{@{}c@{}}$\operatorname{EpBeta}$\\ $(\delta=0.005)$\end{tabular} & \begin{tabular}[c]{@{}c@{}}0.6539\\ (0.0071)\end{tabular} & \begin{tabular}[c]{@{}c@{}}0.8713\\ (0.0013)\end{tabular} & \begin{tabular}[c]{@{}c@{}}0.8185\\ (0.0059)\end{tabular} & \begin{tabular}[c]{@{}c@{}}0.8365\\ (0.0015)\end{tabular} & \begin{tabular}[c]{@{}c@{}}0.6158\\ (0.0140)\end{tabular} & \begin{tabular}[c]{@{}c@{}}0.9326\\ (0.0030)\end{tabular}  \\
\begin{tabular}[c]{@{}c@{}}$\operatorname{EpBeta}$\\ $(\delta=0.01)$\end{tabular}  & \begin{tabular}[c]{@{}c@{}}0.6378\\ (0.0055)\end{tabular} & \begin{tabular}[c]{@{}c@{}}0.8590\\ (0.0026)\end{tabular} & \begin{tabular}[c]{@{}c@{}}0.8042\\ (0.0050)\end{tabular} & \begin{tabular}[c]{@{}c@{}}0.8363\\ (0.0019)\end{tabular} & \begin{tabular}[c]{@{}c@{}}0.6088\\ (0.0218)\end{tabular} & \begin{tabular}[c]{@{}c@{}}0.9354\\ (0.0016)\end{tabular}  \\
\begin{tabular}[c]{@{}c@{}}$\operatorname{EpBeta}$\\ $(\delta=0.05)$\end{tabular}  & \begin{tabular}[c]{@{}c@{}}0.6007\\ (0.0067)\end{tabular} & \begin{tabular}[c]{@{}c@{}}0.8216\\ (0.0030)\end{tabular} & \begin{tabular}[c]{@{}c@{}}0.7596\\ (0.0047)\end{tabular} & \begin{tabular}[c]{@{}c@{}}0.8383\\ (0.0016)\end{tabular} & \begin{tabular}[c]{@{}c@{}}0.6267\\ (0.0120)\end{tabular} & \begin{tabular}[c]{@{}c@{}}0.9401\\ (0.0015)\end{tabular}  \\ 
\hline
$\operatorname{Beta}$                                                              & \begin{tabular}[c]{@{}c@{}}0.6934\\ (0.0085)\end{tabular} & \begin{tabular}[c]{@{}c@{}}0.8976\\ (0.0016)\end{tabular} & \begin{tabular}[c]{@{}c@{}}0.8335\\ (0.0053)\end{tabular} & \begin{tabular}[c]{@{}c@{}}0.8341\\ (0.0016)\end{tabular} & \begin{tabular}[c]{@{}c@{}}0.5919\\ (0.0244)\end{tabular} & \begin{tabular}[c]{@{}c@{}}0.9249\\ (0.0027)\end{tabular}  \\
$\operatorname{Unif}$                                                              & \begin{tabular}[c]{@{}c@{}}0.5862\\ (0.0053)\end{tabular} & \begin{tabular}[c]{@{}c@{}}0.7996\\ (0.0030)\end{tabular} & \begin{tabular}[c]{@{}c@{}}0.7189\\ (0.0080)\end{tabular} & \begin{tabular}[c]{@{}c@{}}0.8400\\ (0.0018)\end{tabular} & \begin{tabular}[c]{@{}c@{}}0.6355\\ (0.0209)\end{tabular} & \begin{tabular}[c]{@{}c@{}}0.9410\\ (0.0017)\end{tabular}  \\ 
\hline
TVAE                                                                               & \begin{tabular}[c]{@{}c@{}}0.4185\\ (0.0061)\end{tabular} & \begin{tabular}[c]{@{}c@{}}0.7020\\ (0.0056)\end{tabular} & \begin{tabular}[c]{@{}c@{}}0.4989\\ (0.0061)\end{tabular} & \begin{tabular}[c]{@{}c@{}}0.8110\\ (0.0040)\end{tabular} & \begin{tabular}[c]{@{}c@{}}0.6829\\ (0.0268)\end{tabular} & \begin{tabular}[c]{@{}c@{}}0.9667\\ (0.0039)\end{tabular}  \\
CTGAN                                                                              & \begin{tabular}[c]{@{}c@{}}0.4212\\ (0.0143)\end{tabular} & \begin{tabular}[c]{@{}c@{}}0.5090\\ (0.0127)\end{tabular} & \begin{tabular}[c]{@{}c@{}}0.2997\\ (0.0133)\end{tabular} & \begin{tabular}[c]{@{}c@{}}0.8193\\ (0.0050)\end{tabular} & \begin{tabular}[c]{@{}c@{}}0.7082\\ (0.0175)\end{tabular} & \begin{tabular}[c]{@{}c@{}}0.9638\\ (0.0022)\end{tabular}  \\
TabDDPM                                                                            & \begin{tabular}[c]{@{}c@{}}0.5427\\ (0.0047)\end{tabular} & \begin{tabular}[c]{@{}c@{}}0.7771\\ (0.0025)\end{tabular} & \begin{tabular}[c]{@{}c@{}}0.6557\\ (0.0045)\end{tabular} & \begin{tabular}[c]{@{}c@{}}0.8476\\ (0.0014)\end{tabular} & \begin{tabular}[c]{@{}c@{}}0.7459\\ (0.0102)\end{tabular} & \begin{tabular}[c]{@{}c@{}}0.9862\\ (0.0012)\end{tabular}  \\
GReaT                                                                              & \begin{tabular}[c]{@{}c@{}}0.4966\\ (0.0091)\end{tabular} & \begin{tabular}[c]{@{}c@{}}0.8042\\ (0.0024)\end{tabular} & -                                                         & -                                                         & \begin{tabular}[c]{@{}c@{}}0.7433\\ (0.0098)\end{tabular} & \begin{tabular}[c]{@{}c@{}}0.9697\\ (0.0070)\end{tabular}  \\
\hline
\end{tabular}
\end{table}

\clearpage

\section{STATISTICAL INFERENCE IN A CLASSIFICATION EXAMPLE}
\label{sec:statistical:classification}

Preserving structure is crucial for statistical inference, not only in the regression case mentioned in Sec.~\ref{sec:experiment}, but also in classification. Here, we present a classification example where statistical inference plays a key role.

In this example, we estimate the decision boundary for object classification, a problem known as the support problem in the classification literature. To demonstrate that the $\operatorname{EpBeta}$ method results in a more robust boundary compared to existing mixup methods, we use a dataset with three classes distributed on a two-dimensional x-y plane. Specifically, we generate 500 instances for each class from the following distributions:
$\operatorname{N} \big([0, 0]^\top, \mathbf{I} \big)$ for the first class, $\operatorname{N} \big([2, 0]^\top, \mathbf{I} \big)$ for the second class, and $\operatorname{N} \big([4, 0]^\top, \mathbf{I} \big)$ for the third class, where $\mathbf{I}$ is an identity matrix. 

We then synthesize samples 100 times using mixup with $\operatorname{EpBeta}(\delta=0.05)$ and $\epsilon_0=\epsilon_1 =0.3$, or with $\operatorname{Unif}(0,1)$. For each synthetic sample, we estimate the decision boundary using a support vector machine and calculate the intersection point of the boundary with the x-axis ($y=0$).

\begin{table}[h]
\caption{Classification result.}
\label{table:classification}
\centering
\begin{tabular}{c|cc|c}
\hline
       & bias of boundary (class 1 vs 2) & bias of boundary (class 2 vs 3) & accuracy      \\
\hline\hline
$\operatorname{EpBeta}(\delta=0.05)$ & +0.079 (0.062)            & -0.048 (0.056)            & 0.798 (0.004) \\
$\operatorname{Unif}(0,1)$   & +0.249 (0.057)            & -0.223 (0.049)            & 0.781 (0.004) \\
\hline
\end{tabular}
\end{table}

As shown numerically in Table~\ref{table:classification}, synthetic data generated by the $\operatorname{EpBeta}$ results in a more unbiased decision boundary compared to the $\operatorname{uniform}$. This setting is actually related to the manifold intrusion problem \citep{guo2019mixup}, where the classification accuracy decreases for the second class, which is situated between the other two classes. This distortion affects not only statistical robustness, but also undermines classification accuracy.

\section{ADDITIONAL EXPERIMENT ON IMAGE DATA}

Supervised contrastive learning (SupCL) \citep{khosla2020supervised} is a powerful framework for learning effective representations for a variety of downstream tasks. However, it can lead to class-collapsed representations, where embedding outputs within the same class collapse to a single point, reducing performance \citep{islam2021broad, chen2022perfectly, lee2025supcl}. In other words, decreasing within-class variance of embedding outputs can harm performance.

To show the usefulness of EpBeta, we evaluate the transfer learning performances of SupCL. Specifically, we train the ResNet18 encoder with a 2-layer MLP projector head for 500 epochs, using a batch size of 500 and a temperature parameter of 0.1 in SupCL loss, on CIFAR-10 augmented with either EpBeta ($\delta=0.05$) or Uniform mixup. Following the transfer learning evaluation protocol \citep{kornblith2019better, lee2021improving}, we remove the projector head and train a linear classifier on top of the frozen encoder using 6 downstream datasets: Dogs \citep{khosla2011novel}, DTD \citep{cimpoi2014dtd}, Flowers \citep{nilsback2008data_flowers102}, Food \citep{bossard14}, Pets \citep{parkhi2012pets}, and MIT67 \citep{quattoni2009mit67}. We repeat the entire process five times with different seeds. The table below presents the top-1 linear probing accuracy along with standard deviation. The results show that using EpBeta distribution instead of uniform distribution consistently outperforms, underscoring the benefits of variance preservation with EpBeta.

\begin{table}[ht]
\caption{Top-1 linear probing accuracy.}
\label{table:supcl}
\centering
\begin{tabular}{c|cccccc}
\hline
Name & Dogs  & DTD & Flowers & Food & MIT67 & Pets \\
\hline\hline
$\operatorname{EpBeta}$ & \begin{tabular}[c]{@{}c@{}}0.150\\ (0.004)\end{tabular} & \begin{tabular}[c]{@{}c@{}}0.404\\ (0.007)\end{tabular} & \begin{tabular}[c]{@{}c@{}}0.552\\ (0.007)\end{tabular} & \begin{tabular}[c]{@{}c@{}}0.301\\ (0.002)\end{tabular} & \begin{tabular}[c]{@{}c@{}}0.366\\ (0.007)\end{tabular} & \begin{tabular}[c]{@{}c@{}}0.268\\ (0.002)\end{tabular} \\
$\operatorname{Unif}$ & \begin{tabular}[c]{@{}c@{}}0.140\\ (0.004)\end{tabular} & \begin{tabular}[c]{@{}c@{}}0.390\\ (0.004)\end{tabular} & \begin{tabular}[c]{@{}c@{}}0.540\\ (0.005)\end{tabular} & \begin{tabular}[c]{@{}c@{}}0.284\\ (0.003)\end{tabular} & \begin{tabular}[c]{@{}c@{}}0.355\\ (0.008)\end{tabular} & \begin{tabular}[c]{@{}c@{}}0.246\\ (0.007)\end{tabular} \\ \hline
\end{tabular}
\end{table}

\clearpage
\section{EPBETA PARAMETER EXAMPLES}
\label{appendix:epbeta:parameters}

Each cell in tables enumerates $\alpha$ and $\beta$ in order that satisfy \eqref{eq:epbeta:constraint} in Theorem~\ref{thm:epbeta:var} and equality condition of \eqref{eq:epbeta:conditional:constraint} in  Theorem~\ref{thm:epbeta:condimean} with $\alpha \geq \beta$ for given $\epsilon_0, \epsilon_1 \in \{0.0, 0.1, 0.2, \cdots, 0.9 \}$, and $\delta \in \{0.005, 0.01\}$.

\begin{table}[h]
\centering
\caption{Structure-preserving $\operatorname{EpBeta}$ parameters for $\delta=0.005$.}
\vspace{-8pt}
\begin{tabular}{c|ccccc}
\hline
 & $\epsilon_1=$ 0.0 & 0.1 & 0.2 & 0.3 & 0.4 \\ \hline
$\epsilon_0=$ 0.0 & - & 18.09, 1.91 & 33.16, 6.83 & 45.92, 14.08 & 56.86, 23.14 \\
0.1 & 0.10, 0.01 & 20.17, 1.93 & 37.16, 6.96 & 51.72, 14.41 & 64.35, 23.80 \\
0.2 & 0.20, 0.01 & 22.26, 1.96 & 41.18, 7.06 & 57.57, 14.69 & 71.90, 24.37 \\
0.3 & 0.30, 0.01 & 24.35, 1.97 & 45.21, 7.16 & 63.44, 14.94 & 79.55, 24.88 \\
0.4 & 0.41, 0.01 & 26.44, 1.99 & 49.24, 7.24 & 69.35, 15.16 & 87.23, 25.33 \\
0.5 & 0.51, 0.01 & 28.54, 2.00 & 53.29, 7.31 & 75.28, 15.36 & 94.96, 25.73 \\
0.6 & 0.61, 0.01 & 30.64, 2.02 & 57.36, 7.37 & 81.24, 15.54 & 102.74, 26.09 \\
0.7 & 0.71, 0.01 & 32.73, 2.03 & 61.42, 7.43 & 87.22, 15.69 & 110.57, 26.42 \\
0.8 & 0.81, 0.01 & 34.83, 2.04 & 65.49, 7.48 & 93.19, 15.84 & 118.39, 26.71 \\
0.9 & 0.91, 0.01 & 36.94, 2.05 & 69.57, 7.53 & 99.23, 15.97 & 126.27, 26.99 \\ \hline\hline
 & $\epsilon_1=$ 0.5 & 0.6 & 0.7 & 0.8 & 0.9 \\ \hline
$\epsilon_0=$ 0.0 & 66.34, 33.67 & 74.62, 45.37 & 81.95, 58.06 & 88.44, 71.55 & 94.27, 85.75 \\
0.1 & 75.38, 34.76 & 85.12, 47.03 & 93.79, 60.38 & 101.55, 74.66 & 108.50, 89.67 \\
0.2 & 84.57, 35.74 & 95.82, 48.51 & 105.88, 62.46 & 114.95, 77.43 & 123.13, 93.25 \\
0.3 & 93.86, 36.60 & 106.67, 49.83 & 118.17, 64.33 & 128.60, 79.94 & 138.06, 96.48 \\
0.4 & 103.22, 37.37 & 117.61, 51.01 & 130.66, 66.03 & 142.49, 82.22 & 153.29, 99.45 \\
0.5 & 112.69, 38.07 & 128.72, 52.09 & 143.27, 67.56 & 156.56, 84.30 & 168.81, 102.19 \\
0.6 & 122.20, 38.69 & 139.88, 53.06 & 156.07, 68.98 & 170.87, 86.24 & 184.45, 104.66 \\
0.7 & 131.81, 39.27 & 151.14, 53.95 & 168.97, 70.28 & 185.31, 88.01 & 200.32, 106.96 \\
0.8 & 141.39, 39.78 & 162.52, 54.78 & 181.92, 71.45 & 199.86, 89.63 & 216.45, 109.13 \\
0.9 & 151.09, 40.26 & 173.93, 55.53 & 194.99, 72.54 & 214.49, 91.11 & 232.58, 111.07 \\ \hline
\end{tabular}
\end{table}

\begin{table}[h]
\centering
\caption{Structure-preserving $\operatorname{EpBeta}$ parameters for $\delta=0.01$.}
\vspace{-8pt}
\begin{tabular}{cccccc}
\hline
\multicolumn{1}{c|}{} & $\epsilon_1=$ 0.0 & 0.1 & 0.2 & 0.3 & 0.4 \\ \hline
\multicolumn{1}{c|}{$\epsilon_0=$ 0.0} & - & 8.92, 0.99 & 16.49, 3.50 & 22.84, 7.15 & 28.28, 11.71 \\
\multicolumn{1}{c|}{0.1} & 0.10, 0.04 & 9.98, 1.01 & 18.54, 3.57 & 25.79, 7.34 & 32.08, 12.07 \\
\multicolumn{1}{c|}{0.2} & 0.21, 0.03 & 11.04, 1.02 & 20.59, 3.63 & 28.76, 7.49 & 35.91, 12.37 \\
\multicolumn{1}{c|}{0.3} & 0.31, 0.03 & 12.11, 1.03 & 22.65, 3.69 & 31.76, 7.63 & 39.78, 12.64 \\
\multicolumn{1}{c|}{0.4} & 0.41, 0.03 & 13.17, 1.04 & 24.72, 3.73 & 34.76, 7.75 & 43.68, 12.88 \\
\multicolumn{1}{c|}{0.5} & 0.51, 0.03 & 14.23, 1.05 & 26.79, 3.78 & 37.79, 7.86 & 47.60, 13.10 \\
\multicolumn{1}{c|}{0.6} & 0.62, 0.03 & 15.29, 1.06 & 28.86, 3.81 & 40.82, 7.96 & 51.55, 13.29 \\
\multicolumn{1}{c|}{0.7} & 0.72, 0.03 & 16.35, 1.07 & 30.95, 3.85 & 43.86, 8.04 & 55.52, 13.47 \\
\multicolumn{1}{c|}{0.8} & 0.82, 0.03 & 17.42, 1.07 & 33.03, 3.88 & 46.91, 8.12 & 59.49, 13.63 \\
\multicolumn{1}{c|}{0.9} & 0.92, 0.03 & 18.48, 1.08 & 35.11, 3.90 & 49.96, 8.20 & 63.49, 13.77 \\ \hline \hline
\multicolumn{1}{l|}{} & $\epsilon_1=$ 0.5 & 0.6 & 0.7 & 0.8 & 0.9 \\ \hline
\multicolumn{1}{c|}{$\epsilon_0=$ 0.0} & 33.00, 17.00 & 37.12, 22.87 & 40.76, 29.23 & 44.00, 36.00 & 46.90, 43.11 \\
\multicolumn{1}{c|}{0.1} & 37.57, 17.58 & 42.42, 23.74 & 46.73, 30.44 & 50.59, 37.60 & 54.06, 45.13 \\
\multicolumn{1}{c|}{0.2} & 42.21, 18.09 & 47.82, 24.51 & 52.82, 31.52 & 57.33, 39.03 & 61.41, 46.96 \\
\multicolumn{1}{c|}{0.3} & 46.91, 18.55 & 53.28, 25.20 & 59.03, 32.49 & 64.22, 40.33 & 68.95, 48.64 \\
\multicolumn{1}{c|}{0.4} & 51.65, 18.95 & 58.83, 25.82 & 65.32, 33.37 & 71.23, 41.51 & 76.61, 50.15 \\
\multicolumn{1}{c|}{0.5} & 56.44, 19.32 & 64.42, 26.37 & 71.69, 34.16 & 78.33, 42.58 & 84.41, 51.55 \\
\multicolumn{1}{c|}{0.6} & 61.25, 19.65 & 70.08, 26.89 & 78.14, 34.89 & 85.53, 43.57 & 92.32, 52.84 \\
\multicolumn{1}{c|}{0.7} & 66.11, 19.95 & 75.77, 27.35 & 84.64, 35.56 & 92.80, 44.48 & 100.31, 54.01 \\
\multicolumn{1}{c|}{0.8} & 70.99, 20.23 & 81.51, 27.78 & 91.20, 36.17 & 100.13, 45.31 & 108.43, 55.12 \\
\multicolumn{1}{c|}{0.9} & 75.88, 20.48 & 87.27, 28.17 & 97.81, 36.74 & 107.54, 46.09 & 116.59, 56.13 \\ \hline
\end{tabular}
\vspace{-80pt}
\end{table}

\end{document}